\documentclass[preprint,12pt]{elsarticle}

\usepackage[a4paper, margin=0.7in]{geometry}

\usepackage[utf8]{inputenc}
\usepackage[T1]{fontenc}
\DeclareUnicodeCharacter{2500}{\textemdash}

\usepackage{amsmath, amssymb, amsfonts}

\usepackage{graphicx}
\usepackage{float}

\usepackage{booktabs}
\usepackage{tabularx}
\usepackage{multirow}

\usepackage{microtype}
\usepackage{caption}
\usepackage{color}
\usepackage{soul}

\usepackage{siunitx}

\usepackage[numbers]{natbib}
\usepackage{doi}
\usepackage{url}
\usepackage{hyperref}

\usepackage{zref-totpages}
\usepackage{totcount}
\usepackage{xr}
\usepackage{comment}

\journal{Applied Soft Computing}

\begin{document}

\begin{frontmatter}

\title{Bayesian Optimization for Mixed-Variable Problems in the Natural Sciences}

\author[inst1]{Yuhao Zhang}
\author[inst1]{Ti John}
\author[inst3]{Matthias Stosiek}
\author[inst3,inst4,inst5]{Patrick Rinke}

\affiliation[inst1]{organization={Department of Applied Physics, Aalto University},
            addressline={Puumiehenkuja 2}, 
            city={Espoo},
            postcode={02320}, 
            country={Finland}}

\affiliation[inst2]{organization={Department of Computer Science
Aalto University},
            addressline={Kemistintie 1}, 
            city={Espoo},
            postcode={02150}, 
            country={Finland}}

\affiliation[inst3]{organization={School of Natural Sciences Physics Department, Technical University of Munich},
            addressline={James-Franck-Str. 1}, 
            city={Garching bei München},
            postcode={85748}, 
            country={Germany}}

\affiliation[inst4]{organization={Atomistic Modeling Center, Munich Data Science Institute, Technical University of Munich},
            addressline={Walther-von-Dyck Str. 10}, 
            city={Garching bei München},
            postcode={85748}, 
            country={Germany}}

\affiliation[inst5]{organization={Munich Center for Machine Learning (MCML)},
            addressline={Arcisstr. 21}, 
            city={München},
            postcode={80333}, 
            country={Germany}}

\begin{abstract}
Optimizing expensive black-box objectives over mixed search spaces is a common challenge across the natural sciences. Bayesian optimization (BO) offers sample-efficient strategies through probabilistic surrogate models and acquisition functions. However, its effectiveness diminishes in mixed or high-cardinality discrete spaces, where gradients are unavailable and optimizing the acquisition function becomes computationally demanding. In this work, we generalize the probabilistic reparameterization (PR) approach of Daulton et al. to handle non-equidistant discrete variables, enabling gradient-based optimization in fully mixed-variable settings with Gaussian process (GP) surrogates. With real-world scientific optimization tasks in mind, we conduct systematic benchmarks on synthetic and experimental objectives to obtain an optimized kernel formulations and demonstrate the robustness of our generalized PR method. We additionally show that, when combined with a modified BO workflow, our approach can efficiently optimize highly discontinuous and discretized objective landscapes. This work establishes a practical BO framework for addressing fully mixed optimization problems in the natural sciences, and is particularly well suited to autonomous laboratory settings where noise, discretization, and limited data are inherent.

\end{abstract}


\end{frontmatter}

\section{Introduction}

Optimization problems are common across scientific domains and often involve black-box objectives with high evaluation costs. Common examples include time-consuming laboratory experiments and computationally expensive simulations. In these settings the goal is to minimize the number of experiments required to optimize a target (e.g., material properties) as a function of the input variables (e.g., processing and/synthesis conditions) \cite{SUN20211305,Lofgren2022,BO_opt_advanced,PedersenBO,ZHANG_actuator,SaninBO,DimentBO,MirandaBO}. Such optimization tasks commonly rely on design of experiments (DOE) methodologies or sampling strategies based on experience and intuition ~\cite{DOE_montgomery2017,DOEGreenhill}. For multidimensional design spaces, experience- and intuition-based sampling methods are subject to bias and lack of exploration leading to the risk of missing the optimum solution. Meanwhile, conventional DOE approaches sample the design space in a predetermined fashion which prevents information gained from new measurements to improve the sampling strategy.


Bayesian Optimization (BO) addresses these limitations by providing an adaptive and data-efficient alternative to traditional DOE sampling strategies. BO constructs a probabilistic surrogate model of the objective function and uses it to intelligently guide the selection of new experiments, explicitly balancing exploration of uncertain regions with exploitation of promising candidates. This adaptivity enables BO to operate effectively under stringent data constraints, making it particularly well-suited to laboratory and computational settings where each evaluation is costly. As a result, BO has become a powerful framework for efficient global optimization and has been widely adopted to accelerate materials discovery, consistently demonstrating strong empirical performances \cite{Armi_Perovskite,Walsh_BO,BO_Glopt2,BO_opt_advanced,Jin2022_exp_boss, BO_exp_worflow}. Owing to their nonparametric nature and high flexibility enabled by diverse kernel formulations and the ability to incorporate priors on hyperparameters, GPs have become a widely used surrogate model in BO. This is particularly true when the underlying objective function varies smoothly with respect to its input parameters. GPs naturally encode such smoothness through their kernel covariance functions, enabling accurate modeling of the objective landscape \cite{rasmussen2006gaussian}.

However, the use of BO with GPs is currently limited by its lack of support for fully mixed variables consisting of continuous, integer, discrete and categorical variables which are commonly encountered in experimental or simulation-based optimization tasks. For example, in thin-film deposition, the growth substrate can be encoded as a categorical variable. Deposition and annealing temperatures are continuous, but in practice they are often restricted to a fixed set of integer or discrete values. Although different methods have been proposed to handle non-continuous variables, as will be discussed in Section~\ref{Related work}, they are typically evaluated on benchmarks that frequently contain theoretical functions with numerous sharp local minima (e.g., Ackley and Rastigrin). Moreover, these benchmarks assume a noiseless objective landscape, where repeated evaluations produce identical function values. Consequently, the Gaussian process models are initialized with extremely small noise levels (1e-2 to 1e-6), which are generally unrealistic for real world applications \cite{Daulton2022,COSMOPOLITAN,Eduardo_discrete,2023MCBO}. In real-world experiments, sampled values are often noisy and arise from objective landscapes that contain less challenging features. As a result of these factors, the reported performance may not transfer to experimental optimization tasks. Consequently, the behavior and effectiveness regarding many of these mixed-variable methods in practical experimental settings remain largely untested and unclear.

To address these issues, we implemented a fully mixed variable BO method that employs a GP as surrogate model based on the Probabilistic Reparameterization (PR) approach developed by \citet{Daulton2022}, which we refer to as Generalized PR. We then benchmark and optimize our Generalized PR regarding the AF, kernel and prior specifications for optimal performances on objective landscapes typically encountered in real-world optimization tasks in the natural sciences.

For AF optimization we compare two commonly used acquisition functions: Expected Improvement (EI) and the Upper/Lower Confidence Bound family (UCB/LCB). In the BO community, EI is widely regarded as more exploitative, as it prioritizes regions that are already predicted to yield high improvement \cite{Ryzhov2016,HennigSchuler2012,Frazier2018}. In contrast, UCB/LCB introduces an explicit exploration parameter that, given a suitable hyperparameter choice, encourages sampling in regions of higher model uncertainty, and is therefore considered more exploratory. Because of this difference, community practice often benchmarks EI against UCB/LCB to assess the trade-off between refinement around promising areas and broader global search \cite{DeAthEversonRahatFieldsend2019}. We adopt this convention here to evaluate how exploration–exploitation balance affects convergence in our settings.

Motivated by the aforementioned challenges, this work is guided by the following research objectives.  
First, we evaluate whether the probabilistic reparameterization framework of \citet{Daulton2022}, originally developed for continuous, integer and categorical spaces, can be extended to also effectively support discrete variables in a principled and unified manner. Second, we use a systematic suite of synthetic and real-world benchmark problems to investigate how the BO workflow regarding construction of the GP kernel and the choice of acquisition function should be designed. Our goal is to understand which modeling choices lead to robust performance across the diverse optimization landscapes encountered in natural science applications. Third, we examine the behavior of GP models in fully discrete search spaces and explore possible structural failure modes such as repeated sampling due to modeling artifacts, and if so, how they might be mitigated. Finally, we ask whether a BO workflow built on our extended modeling framework can remain competitive with tree-based surrogate approaches when confronted with highly discontinuous and rugged objective functions, which may arise from discretized spaces where GP-based models are typically expected to struggle.

\section{Methodology}
For mixed variable BO tasks we consider the problem of optimizing a black-box objective function $f:\mathcal{X}\times\mathcal{Q}\rightarrow\mathbb{R}$ over a bounded mixed search space $\mathcal{X}\times\mathcal{Q}$, where $\mathcal{X}=\mathcal{X}^{\left(1\right)}\times\cdot\cdot\cdot\times\mathcal{X}^{\left(c\right)}$, is the domain of $c\geq0$ continuous parameters and $\mathcal{Q}=\mathcal{\mathcal{Q}}^{\left(1\right)}\times\cdot\cdot\cdot\times\mathcal{\mathcal{Q}}^{\left(d\right)}$ is the domain of $d\geq0$ non-continuous variables (integer, discrete and categorical). As a physical example, consider optimizing a multilayer material for a desired property such as strength, conductivity, or transparency. Integer variables may represent the number of layers, categorical variables may correspond to material type, stacking order, or substrate, and discrete variables may take values from a predefined set, such as layer thicknesses or process parameters constrained by fabrication limits. The AF $\alpha: \mathcal{X} \times \mathcal{Q} \to \mathbb{R}$ assigns a real value to each candidate pair $(\boldsymbol{x}, \boldsymbol{q})$ where $\boldsymbol{x}\in\mathcal{X}$ and $\boldsymbol{q}\in\mathcal{Q}$. Maximizing $\alpha$ in this mixed space is therefore necessary but presents a non-trivial challenge. Generally mixed variable BO with GPs perform the following: let $ \boldsymbol{\theta}\in\Theta\subseteq\mathbb{R}^{m}$ be continuous variables, where $m\geq c+d$. We then define a mapping: 
\begin{equation}
    g: \Theta \rightarrow \mathcal{X} \times \mathcal{Q}, \quad \boldsymbol{\theta} \mapsto (\boldsymbol{x}, \boldsymbol{q}),
\label{eq:mapping}
\end{equation}
where $\Theta \subseteq \mathbb{R}^m $ is a continuous search space, and $(\boldsymbol{x}, \boldsymbol{q}) \in \mathcal{X} \times \mathcal{Q}$ denotes the corresponding mixed-variable configuration. Using this mapping, we define a reparameterized AF:

\begin{equation}
\tilde{\alpha}(\boldsymbol{\theta}) = \alpha(g(\boldsymbol{\theta})),
\label{eq:mixed_AF}
\end{equation}
which allows us to optimize the AF entirely in the continuous domain $\Theta$ using a GP with kernel $\kappa$. Let $\boldsymbol{\Theta} = [\boldsymbol{\theta}_1, \dots, \boldsymbol{\theta}_N] \in \Theta^N$ denote the matrix of training inputs,  and $\boldsymbol{\Theta}_* = [\boldsymbol{\theta}_{*1}, \dots, \boldsymbol{\theta}_{*N_*}] \in \Theta^{N_*}$ the test inputs. In this domain the predictive posterior mean and covariance on the test inputs is given by:

\begin{equation}
\mu_{*}=\text{\textbf{k}}_{*}^{T}\left(\text{\textbf{K}}+\sigma_{0}^{2}\text{\textbf{I}}\right)^{-1}\text{\textbf{y}.}
    \label{eq:GP mean}
\end{equation}

\begin{equation}
\varSigma_{*}=k_{**}-\text{\textbf{k}}_{*}^{T}\left(\text{\textbf{K}}+\sigma_{0}^{2}\text{\textbf{I}}\right)\text{\textbf{k}}_{*},
    \label{eq:GP Cov}
\end{equation}
where $\mathbf{k}_* = \kappa(\boldsymbol{\Theta}, \boldsymbol{\Theta}_*) \in \mathbb{R}^{N \times N_*}$ is the covariance matrix between training and test inputs, $k_{**} = \kappa(\boldsymbol{\Theta}_*, \boldsymbol{\Theta}_*) \in \mathbb{R}^{N_* \times N_*}$ is the covariance matrix between test inputs, $\mathbf{K} = \kappa(\boldsymbol{\Theta}, \boldsymbol{\Theta}) \in \mathbb{R}^{N \times N}$ is the covariance matrix between training inputs and $\sigma_{0}^{2}$ is the fitted noise (variance) hyperparameter that models the underlying noise (aleatoric uncertainty) within the data. Designing an effective mapping $g$ is not straightforward, as it should ideally preserve information such as the relative distances and magnitudes between the ordinal (integer and discrete) variables. In addition to non-GP approaches, the next section contains brief overviews of several mapping methods that have been developed to address this challenge.

\subsection{Related Work}\label{Related work}
\textbf{Tree-based Methods:} Mixed-variable optimization can also be approached using tree-based surrogate models such as Random Forests (RFs), which naturally handle discrete and categorical inputs without explicit variable transformation. The ensemble structure of RFs enables a frequentist estimation of predictive uncertainty from the variance of individual tree predictions~\citep{Geurts2006}. Although this entails certain drawbacks relative to Bayesian probabilistic uncertainty in GPs, it provides a practical and direct way of handling problems with discontinuous or step-like objective landscapes without complex variable transformations. Due to these reasons, RFs have emerged as popular surrogate for handling mixed-variable BO tasks within the scientific community \cite{advanced_RF}.   

Recent Bayesian extensions such as Bayesian Additive Regression Trees (BART)~\citep{BARTChipman2008} and the Bayesian Additive Regression Kernel (BARK)~\citep{BARK2025} combine the flexibility of tree ensembles with the Bayesian uncertainty quantification of GPs. These models retain the ability to represent non-smooth objectives while introducing a Bayesian framework for more interpretable uncertainty decomposition, making them promising surrogates for future BO tasks within the natural sciences. However, these newer tree-based models have seen limited use in benchmarks and real-world BO tasks, and their performance therefore remains largely unknown when compared to the more widely used GP-based models.

\textbf{Latent Variable GP Methods}: One of the standard approaches for handling mixed variables in BO while still utilizing a GP as surrogate model involves latent variable/continuous relaxation methods. In these approaches, each non-continuous parameter is mapped onto a continuous latent space $\mathcal{Q}^{\prime}\subset\mathbb{R}^{m}$. This space is often one to three-dimensional depending on the number of different possible values (levels) the integer, discrete or categorical variable can take ~\cite{Zhang2022_LVGP_f_vs_b,CuestaRamirez2022}. For instance, a categorical variable denoting the selection among three lubricants can be embedded in a continuous latent space defined by viscosity and boiling temperature. The AF is optimized within this space, and the optimal point is subsequently mapped back to the original mixed input space corresponding to a specific lubricant. In practice, the continuous space does not have to correspond to a defined physical property but can instead represent an abstract latent space.

\citet{Zhang2022_LVGP_f_vs_b} and \citet{CuestaRamirez2022} have shown this method to generally outperform RFs on objectives landscapes enountered in the natural sciences and smooth synthetic test functions. However, these methods are primarily designed for categorical variables, and when applied to integer or discrete inputs, they do not explicitly preserve relative distance information during the continuous transformation. Additionally, AF values in the continuous latent space do not account for the discretization process that maps latent points back to the original mixed-space. Consequently, AF values in the latent space may overestimate the true value after discretization. In the worst case, this can cause repeated resampling within the mixed-space \cite{Daulton2022}.

\textbf{Kernel Rounding Trick}:\label{subsec_kernel:rounding} This method developed by \citet{Eduardo_discrete}, overcomes the latent variable transformation-related issues when applied to integer variables. In this method the GP surrogate is directly discretized such that each continuous input is rounded to the nearest integer before calculation of the kernel. The resulting posterior mean and variance are piecewise constant in the intervals $[-0.5 +k, k + 0.5]$ for each integer $k$ such that the AF is also appropriately discretized without necessitating further modifications. The discretized kernel is given by
\begin{equation}
\label{eqn:Transformed kernel}
K(k,k') = K_d(x,x') = K\left(T(x),T(x')\right)
\end{equation}
where $K_d(\cdot,\cdot')$ is the discretized form of the continuous kernel $K(\cdot,\cdot')$ with $T(\cdot)$ being the transformation function that rounds continuous inputs $(x,x')\in\mathcal{X}$ to the nearest integer $(k,k')\in\mathcal{Z}$. The impact of this kernel rounding method on model posterior is illustrated in Figure \ref{SI_fig:KR effect} Supporting Information.

However, similar with the aforementioned tree-based methods, the kernel rounding method does not provide analytic gradients for the AF. The lack of analytically accessible gradients prevents efficient gradient-based optimization methods leading to poor scalability in high-dimensional spaces. Moreover, the authors have not extended this method to directly support discrete variables.

Another non-gradient-based GP model that does not rely on latent variable transformations is CASMOPOLITAN by \citet{COSMOPOLITAN}. Similar to the Kernel Rounding (KR) method, CASMOPOLITAN employs a GP to model the mixed-variable objective function. It utilizes local trust regions alongside an interleaved optimization strategy, alternating between exploration of discrete variables and gradient-free optimization of continuous variables. While effective in complex mixed spaces, like latent-variable methods, CASMOPOLITAN does not inherently preserve the ordinal structure of non-continuous variables, which may reduce efficiency when ordinal relationships are important.

\textbf{Parametric models}: Parametric models, such as Bayesian neural networks, which use a Bayesian approach for uncertainty quantification, can address the aforementioned issues with RFs and latent variable GPs~\cite{OLIVIERBNN}. While partially Bayesian neural networks (PBNNs) have demonstrated strong performance with reduced data requirements and lower computational cost~\cite{PBNN}, their application so far has been limited to high-dimensional objectives (with nine features) and datasets containing hundreds of samples. Consequently, their applicability to data-scarce and low-dimensional problems commonly encountered in experimental or computationally expensive simulation tasks remains untested.

\textbf{Probabilistic Reparameterization Method}: A recent study by \citet{Daulton2022}, from Meta proposed an alternative approach to handle mixed variables in BO using a GP as the surrogate model. Their method is called probabilistic reparameterization (PR) where rather than optimizing the AF over a latent continuous space they introduced a discrete probability distribution $p\left(\boldsymbol{Q}\mid\boldsymbol{\theta}\right)$ over a random variable $\boldsymbol{Q}$ with support exclusively over $\mathcal{Q}$. This distribution is parameterized by a vector of continuous parameters $\boldsymbol{\theta}$. Given this reparameterizaion, the probabilistic objective (PO) is defined as:

\begin{equation}
\mathbb{E}_{\boldsymbol{Q}\sim p\left(\boldsymbol{Q}\mid\boldsymbol{\theta}\right)}\left[\alpha\left(\boldsymbol{x},\boldsymbol{Q}\right)\right]=\sum_{\boldsymbol{z}\in\mathcal{Z}}p\left(\boldsymbol{q}\mid\boldsymbol{\theta}\right)\alpha\left(\boldsymbol{x},\boldsymbol{Q}\right),
\label{eq:PO}
\end{equation}
where $\alpha$ is the AF. Since $p\left(\boldsymbol{Q}\mid\boldsymbol{\theta}\right)$ is a discrete probability distribution, its expectation can be expressed as a linear combination where each discrete option is weighted by its probability mass. The authors have formally proven that maximizing this PO for $(\boldsymbol{x},\boldsymbol{\theta})$ is equivalent to maximizing $\alpha$ over the original mixed-variable space $\mathcal{X}\times\mathcal{Q}$. Also, by choosing $p\left(\boldsymbol{Q}\mid\boldsymbol{\theta}\right)$ to be a discrete distribution over $\mathcal{Q}$ means that all sampled instances $\boldsymbol{q}$ are feasible non-continuous values given any $\boldsymbol{\theta}$. Within benchmarks conducted \citet{Daulton2022} PR is shown to outperform the KR and COSMOPOLITAN methods by \citet{Eduardo_discrete} and \citet{COSMOPOLITAN} respectively.

\subsection{Generalized Probabilistic Reparameterization}
Due to its combination of strong benchmark performance, access to gradients and absence of mapping-related issues as present in latent variable methods, we selected the PR implementation by \citet{Daulton2022} as the foundation for implementing our mixed-variable GP model. The original PR implementation already handles binary, integer (termed ordinal in the original paper), and categorical variables. Our generalized PR method directly adopts their formulation without changes and extends it to additionally handle discrete variables. Table \ref{tab:parameter-types} summarizes the full reparameterizations for all four variable types used by the PO, as defined in equation \ref{eq:PO}.

\begin{table}[!htbp]
\centering
\begin{tabular}{ccc}
\toprule
Parameter Type & Random Variable & Continuous Parameter \\
\midrule
Binary & \( Z \sim B\bigl(b(\theta)\bigr) \) & \( \theta \in [0,1] \) \\
\midrule
Integer & \( Z = \lfloor \theta \rfloor + B\bigl(\mathcal{I}(\theta)\bigr) \) & \( \theta \in [0, I-1] \) \\
\midrule
Discrete & \( Z = d_i + (d_{i+1} - d_i) B\bigl(\mathcal{D}(\theta)\bigr) \) & \( \theta\in\left[d_{1},d_{2},...,d_{D}\right] \) \\
\midrule
Categorical & \( Z \sim \text{CAT}\bigl(C(\boldsymbol{\theta})\bigr), \quad \boldsymbol{\theta} = (\theta^{(1)}, \ldots, \theta^{(C)}) \) & \( \theta \in \Delta^{C-1} \) \\
\bottomrule
\end{tabular}
\caption{Summary of probabilistic reparameterizations for different variable types. We denote the $(C-1)$-simplex as $\Delta^{C-1}$.}
\label{tab:parameter-types}
\end{table}

Within Table \ref{tab:parameter-types} $B$ denotes a Bernoulli random variable, $I$ and $D$ represents the number of allowed integer and discrete levels for the respective parameter type. Before sampling the mixed variable parameters, the optimized continuous parameter $\theta$ (or $\boldsymbol{\theta}$ for categorical variables) is passed through the following functions:
$b\left(\theta\right)=\sigma\left(\left(\theta-\frac{1}{2}\right)/\tau\right)$, $\mathcal{I}\left(\theta\right)=\sigma\left(\left(\theta-\left\lfloor \theta\right\rfloor -0.5\right)/\tau\right)$, $\mathcal{D}\left(\theta\right)=\sigma\left(\left(\theta-d_{i}-\frac{d_{i+1}-d_{i}}{2}\right)/\tau\right)$ and $C\left(\boldsymbol{\theta}\right)=\text{softmax}\left(\left(\boldsymbol{\theta}-0.5\right)\tau\right)$ corresponding to the binary, ordinal, discrete, and categorical parameter types, respectively where $\sigma$ is the sigmoid function. $\tau$ is the temperature parameter set to 0.1 for all variable types following the original PR implementation. We empirically demonstrate that the temperature parameter value of 0.1 is also suitable for discrete variables shown in Figures \ref{fig:SI:tau_investigation} and \ref{fig:SI:tau_investigation_Dust2} Supporting Information. For the discrete variables, $d_i$ and $d_{i+1}$ are the allowed discrete levels such that $d_{i}\leq\theta\leq d_{i+1}$, the discrete formulation is therefore analogous to the integer case but takes into account non-equal distances.

We also note that in the original PR implementation, the reparameterizations described above apply only when sampling $Z$ from already optimized $\theta$ values. During backpropagation, a differentiable $\tanh$ function is used in place of the non-differentiable Bernoulli distribution. For consistency, we follow the same procedure. The Adam optimizer was used to perform gradient descent on $\boldsymbol{\theta}$ \cite{Kingma2014AdamAM}.

\subsection{Benchmarks}
In this section, we outline the methodology used to empirically evaluate our generalized PR method on both synthetic and real-world problems. Drawing on the systematic BO benchmarks for continuous objective functions by \citet{LeRiche2021}, we evaluated the generalized PR method and jointly optimized the GP model's kernel construction and AF across varying dimensionalities and discretization levels.

Since our generalized PR method is designed for optimization tasks typically encountered in the natural sciences, the synthetic benchmark function is a modified StyblinskiTang function featuring only a single competing local minimum and introduced asymmetry. Additionally, the function is range-normalized across dimensions to ensure a consistent output y-range regardless of dimensionality. Range-normalization is important as increasing the number of input variables (dimensionality) rarely leads to a proportionally larger output (objective) range in optimization tasks regarding natural sciences due to conservation laws, material limits, or saturation effects. The StyblinskiTang function was chosen as basis for modification due to its generalizability to higher dimensions while preserving its landscape shape. This new modified StyblinskiTang test function is named \textit{Butternut Squash} (\textit{BS}) with its analytical expression and two-dimensional visual plot shown in subsection \ref{SI_subsec:butternutsquash} (Supporting Information). Similar to the StyblinskiTang test function, the number of competing minima and thus complexity of the objective landscape increases with dimensionality, a behavior commonly observed in optimization tasks in the natural sciences.

\begin{figure}[!htbp]
    \centering
    \includegraphics[width=0.5\linewidth]{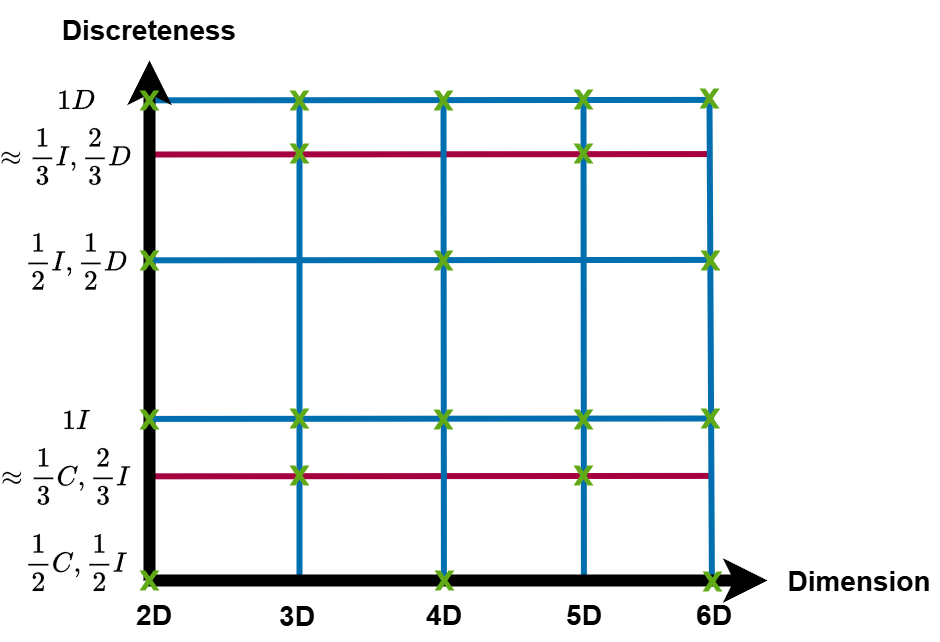}
    \caption{Synthetic \textit{BS} benchmark problems within a function space of dimensionality and variable type where the fractions indicate the proportions of dimensions being continuous (C), integer (I) or discrete (D).}
    \label{fig:function space}
\end{figure}

To obtain more systematic and rigorous benchmark results, the mixed-variable \textit{BS} test function was evaluated across varying dimensionalities and discretization levels. This approach enabled an unbiased evaluation of model performance as a function of both dimensionality and discreteness, since the same underlying function is used across all variations. The different variations of the \textit{BS} function used are illustrated in Figure \ref{fig:function space}.

As shown in Figure \ref{fig:function space}, half of the dimensions are initially continuous within the bounds [-5,5], while the other half are integer-valued in the same range but encoded as positive integers from 0 to 10. For the discrete cases, the allowed values are a subset of the integer values, specifically [0,1,3,4,7,9].\footnote{For the 5D case, the exact fractions for the continuous + integer and integer + dsicrete are $\frac{2}{5},\frac{3}{5}$ hence the approximation within Figure \ref{fig:function space}.} Fully continuous cases were excluded, as they do not make use of the generalized PR method and are thus not relevant to this work. The differing dimensionalities and discretization levels resulted in a total of 20 variants of the \textit{BS} function. The 2D, 3D, 4D, 5D, and 6D variants of the \textit{BS} function were allocated with (5, 35), (10, 80), (20, 100), (40, 160), and (60, 220) pairs of (initial points, iteration budgets), respectively. Each of the 20 variants was further initialized using 10 different Sobol-sequence initialisations to ensure statistical robustness.

The best-performing models or otherwise interesting models from the \textit{BS} benchmarks were then evaluated on two real-world problems. To enable direct comparison and further validate the optimized kernel and our approach, the first problem is a benchmark used in the original PR study by \citet{Daulton2022}. This task involves maximizing the yield of a direct acrylation chemical synthesis by tuning three categorical parameters (solvent, base, and ligand) and two continuous parameters (temperature and concentration). We refer to this problem as \textit{Chemistry}. The solvent, base and ligand consist of 4, 12 and 4 choices respectively while the concentration takes values from 5.7e-2 to 1.53e-1 and temperature from 90-120. Yield values are generated using a pretrained GP surrogate model from \citet{Shields2021} provided in the original PR study. For the \textit{Chemistry} benchmark, a configuration of 20 initial points and a budget of 80 iterations was used. This configuration was repeated across 10 independent runs, each employing a different Sobol-sequence initialisation of the initial points.

The second problem involves maximizing the actuation performance of a thermally activated shape memory polymer, as studied by \citet{ZHANG_actuator}. It requires optimizing three integer-valued processing parameters: ply number (1-9), applied coil stress (1-11), and twist stress (1-11). We refer to this problem as \textit{Actuator}. Similarly to the \textit{Chemistry} function, the actuation values are obtained from the pre-fitted GP surrogate model on the complete dataset. Similarly, the \textit{Actuator} benchmark a configuration of 10 initial points and an iteration budget of 80, was evaluated over 10 Sobol-sequence initialisations.

In the end after acquiring the best overall model from the three benchmarks (\textit{BS}, \textit{Chemistry} and \textit{Actuator}), we tested it on two highly discontinuous functions called Discontinuous Unsmoothed Step-like Test 1 and Discontinuous Unsmoothed Step-like Test 2 (\textit{DUST1} and \textit{DUST2}) which are visualized in Figures \ref{fig:SI:DUST_landscape} and \ref{fig:SI:DUST2_landscape} respectively (Supporting Information). Both DUST functions were designed to model real-world objectives containing large flat regions and step-like features, which can result from phenomena such as phase transitions or discretization of otherwise continuous parameters due to experimental constraints. \textit{DUST1} contains one continuous parameter with bounds [5,25], one integer parameter that is essentially binary taking values [0,1], and one discrete parameter with allowed values [2,4,7,8]. \textit{DUST2} is more complex, with one continuous parameter bounded by [5,50], the same binary parameter, and one discrete parameter taking values [2,3,5,6,9,10,11,12,16,19]. The DUST functions can thus reveal potential weaknesses or limitations of surrogate models when applied within a BO framework to the optimization of highly discontinuous objective landscapes, as often encountered in mixed-variable optimization tasks in the natural sciences.\textit{DUST1} and \textit{DUST2} were allocated with (6, 94), (12, 128), pairs of (initial points, iteration budgets), respectively. Again, each benchmark was run with 10 different Sobol-sequence initialization. 


The number of initial points and maximum iteration budgets were selected heuristically for each banchmark problem. In particular, the iteration budgets were chosen to be sufficiently large to allow for meaningful assessment of model performance. This ensures that the optimization process has adequate opportunity to converge; otherwise, if none of the models reach convergence within the allotted budget, it becomes difficult to draw reliable comparative conclusions. Details regarding the number of initial points and BO iterations for each benchmark function are given in section \ref{SI_subsec:SI:BO_points} Supporting Information.

\subsection{Model Optimization \& Evaluation Metrics}\label{subsec:Models & Metrics}
The GP model of the original PR implementation by \citet{Daulton2022} has not been optimized in terms of kernel design or acquisition function selection. Other mixed-variable GP implementations typically make similar simplifying choices, despite these design decisions being critical to performance. \citet{Daulton2022} employed a generic Mat\'ern-5/2 automatic relevance detection (ARD) kernel without priors within the original PR implementation, as shown in Table \ref{SI_tab:meta kernel} (Supporting Information). After developing and implementing the generalized PR method, by deploying greedy search we optimized the kernel construction using the \textit{BS} functions. In total, 16 different model formulations varying in terms of acquisition function choice (EI or LCB) and kernel construction were benchmarked on the \textit{BS} function using a greedy search. From this initial screening, we selected the best-performing and otherwise interesting models for further benchmarking on the \textit{Actuator} and \textit{Chemistry} Problems. Based on these results, the overall best-performing model was then evaluated on the \textit{DUST1} and \textit{DUST2} benchmarks. In doing so, our generalized PR method is paired with a tested and optimized model ready for real-world simulation or experimental optimization tasks within the natural sciences. 

For all benchmarks, each model was run 10 times using different Sobol sequenced initial points for statistical robustness. The same set of 10 Sobol sequences was used across all models to ensure fair comparisons. Since each model was run 10 times on each benchmark function variant, an exhaustive search of the optimal model settings regarding the kernel formulation and AF choice would be computationally prohibitive. We therefore adopted a greedy search strategy to optimize the kernel and prior choices in a stepwise manner. First, we optimized the kernel type by comparing the radial basis function (RBF) and Mat\'ern-5/2 kernels within a product formulation. This formulation assigns an independent one-dimensional kernel to each input dimension and returns their product. The product kernel structure was initially fixed based on its demonstrated success in both experimental and simulation-based optimization tasks ~\cite{BOSS,ZHANG_actuator,Lofgren2022,Fang2021,Jingrui2024}. For the second step, the best performing kernel type within the product formulation was evaluated against a purely summative one. In the third step, we optimized the choice of prior distribution by comparing the Gamma and LogNormal prior. In the final step, we investigate whether fixing the scale hyperparameter to 1, following \citet{hvafner}, improves performance for standardized inputs. At each stage, we also evaluated the models using both the EI and LCB AFs. The specific greedy search order was chosen based on the assumption that the kernel type and its formulation have the greatest impact on model performance, followed by the choice of prior distribution. In addition to the greedy search kernel formulations, we include the original kernel formulation used within the original  PR implementation.

We additionally evaluate the kernel formulation proposed by \citet{hvafner}, which accounts for the fact that higher dimensional spaces naturally exhibit greater distances between sampling points. Specifically, the expected distance between randomly sampled points in a unit hypercube increases proportionally to \(\sqrt{D}\), where \(D\) is the dimensionality of the space \cite{Chen2009}. Since stationary kernels compute covariances based on these distances, and both diagonal and off-diagonal terms scale with \(\sqrt{D}\), \citet{hvafner} suggest scaling the kernel lengthscales accordingly, i.e., \(l_i \propto \sqrt{D}\), to mitigate the increased complexity in high-dimensional settings. We adopt their proposed kernel configuration in our benchmarks: a Mat\'ern-5/2 ARD kernel with log-normal priors over the lengthscales and fixed scale hyperparameter, as detailed in Table~\ref{SI_tab:hvafner kernel}. 

Although \citet{Daulton2022} have already included the KR method by \citet{Eduardo_discrete} within their benchmarks, they did not apply any special treatment to its kernel formulation or priors. Motivated by the successful use of the KR method (with kernel formulation as detailed in Table~\ref{SI_tab:KR kernel}) by \citet{ZHANG_actuator} for the experimental optimization of thermally-activated polymer actuators in a fully integer search space, we chose to include it in our study. Including the KR method in our systematic benchmarks also allows us to more robustly validate the observed superior performance of PR over KR.

Further details, including the exact mathematical expressions for all kernel constructions used in this study and their prior parameters are provided in Subsection~\ref{SI_subsec:kernels} Supporting Information. The prefixes \texttt{ei} and \texttt{lcb} within the model names indicate whether the EI or LCB AF was used.

We introduce a composite score to conveniently evaluate and compare model performance, as the relatively large number of benchmark functions hinders reliable visual inspection of all the convergence plots. The composite score takes into account both the relative number of converged models out of the 10 different Sobol-point initiated runs and their mean iteration at convergence. Our composite score is defined as:

\begin{equation}
    S_{f}^{(m)}=\frac{\text{C}_{f}^{(m)}}{\text{N}_{f}^{(m)}\cdot\mu_{f}^{(m)}},
    \label{eq:Composite score}
\end{equation}
where $S_{f}^{(m)}$ is the composite score for model $m$ on function $f$, $\text{N}_{f}^{(m)}$ is the total number of model ($m$) runs for function $f$, $\text{C}_{f}^{(m)}$ is the number of converged runs for model ($m$) on function $f$ and $\mu_{f}^{(m)}$ is the mean iteration of these converged models. For the \textit{BS} benchmarks, three tolerance levels (strict, medium, loose) were used to define convergence based on the proximity of the model’s best sample to the true global minimum in the design space. To be considered converged, the sampled integer, discrete, and categorical variables must exactly match those of the global minimum for all criterions. Detailed definitions of each tolerance level for all benchmarked problems are provided in Table \ref{SI_tab:tolerances} (Supporting Information).

To assess overall model performances across the 20 \textit{BS} function instances, we used rank statistics based on composite scores rather than weighted means. Using a weighted mean would disproportionately favor lower-dimensional cases, as they typically have lower convergence values and thus higher composite scores.

The same composite scoring approach was used for evaluating both \textit{Actuator} and \textit{Chemistry}. We employed a single convergence tolerance level for the \textit{Actuator} problem such that exact matches in integer-valued parameters were required. For the \textit{Chemistry}, \textit{DUST1} and \textit{DUST2} problems, three different tolerance levels (strict, medium, and loose) were set with exact matches required under all tolerances for categorical parameters.

All input data were Min-max normalized and the objective values were Z-score standardized for stable hyperparameter fitting of the GP models. The EI AF has no initialized parameters while an exploratory weight value of 2 was used for LCB.

RFs are known to perform better than GPs on discontinuous objective landscapes due to their inherently discrete, tree based structure in regression tasks. In addition, they provide uncertainty quantification and can handle mixed variable spaces without requiring specialized modifications, enabling out-of-the box use within BO frameworks. Furthermore, it has been standard practice to benchmark GP-based mixed variable methods against RFs \cite{Eduardo_discrete,Zhang2022_LVGP_f_vs_b}. For these reasons, we also included a BO workflow using a RF surrogate on both the \textit{DUST1} and \textit{DUST2} problems. The RF model is implemented using the \texttt{scikit-learn} package and initialized with 200 trees and default hyperparameters.


\subsection{Mitigating Resampling of Acquisitions Points}\label{Resampling fix}
\begin{figure}[H]
    \centering
    \includegraphics[width=1\linewidth]{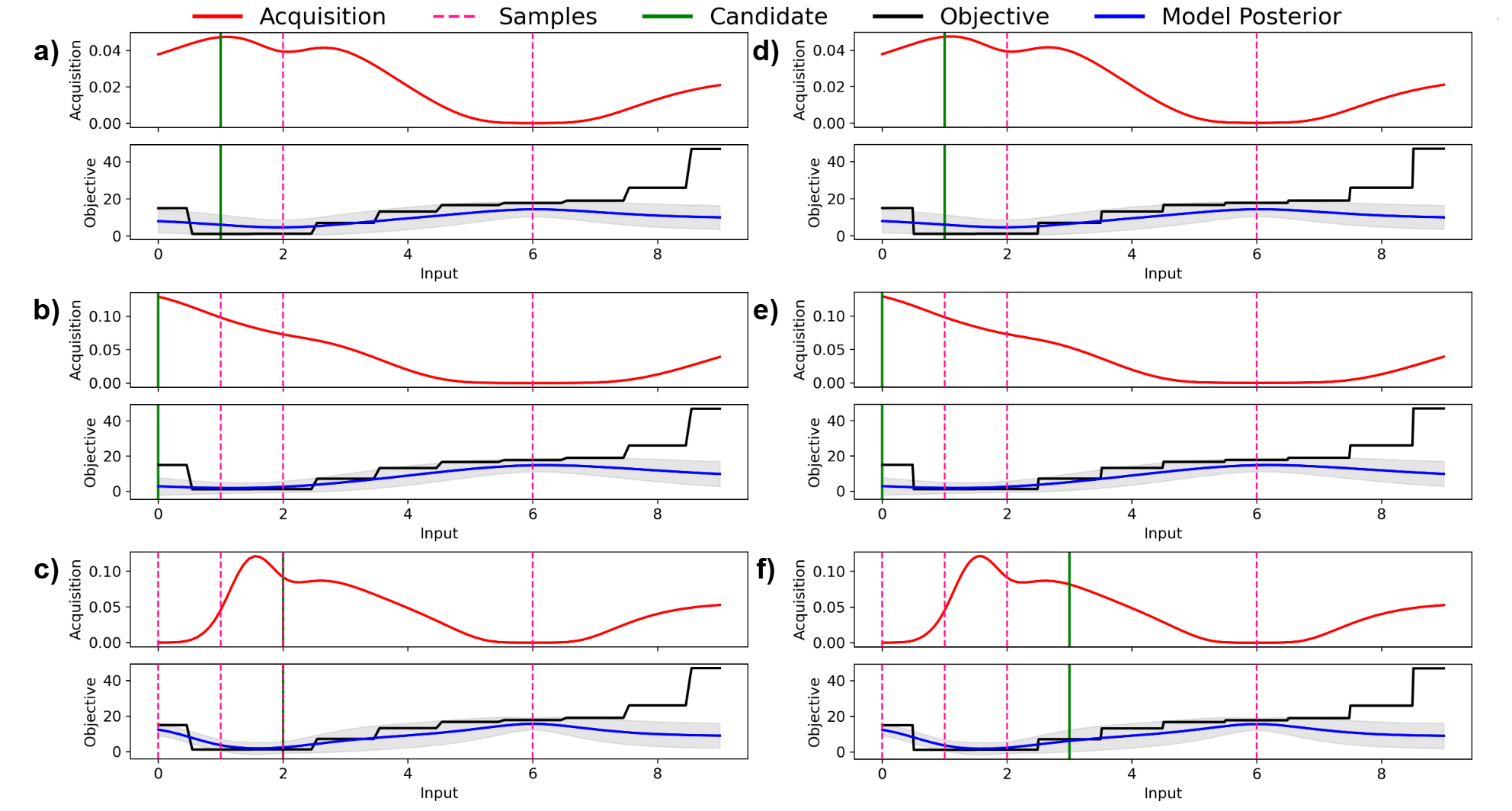}
    \caption{PR using EI with the subtracted fitted noise hyperparameter (left, a–c) shows repeated sampling during the third iteration (c), while PR with the EI penalty term (right, d–f) selects the next maximum (f) when the penalty is applied, avoiding redundant evaluations. Panels a–c and d–f correspond to iterations 1–3 for the subtraction and penalty methods, respectively.}
    \label{fig:acq_repeat}

\end{figure}

While the PR implementation mitigates mapping-related resampling seen in CR/LV methods, we observed that resampling can still occur when the data contain non-negligible noise. We illustrate this resampling phenomenon in Figure \ref{fig:acq_repeat}c), where, during the third BO iteration on the 1D integer \textit{BS} variant, the AF returns an already sampled point when the GP is initialized with a noise level of 0.2. Although the model has converged to the global optimum in this example, in practice, it can become trapped in a local minimum at any time. When this happens, the resampling behavior can persist indefinitely across future iterations unless the model hyperparameters change sufficiently to create a new AF maximum elsewhere in the design space. We suspect this resampling issue was not observed by \citet{Daulton2022} because their benchmarks included at least one continuous parameter and used a small noise hyperparameter of 1e-6.


It is important to clarify that the resampling observed under noisy conditions in the PR implementation arises from a different mechanism than in the CR or LV methods. The behavior is not a limitation of the PR framework itself, but rather a consequence of the GP covariance structure in the presence of observation noise. As shown in Equation~\ref{eq:GP Cov}, the GP covariance includes a fitted noise hyperparameter $\sigma_{0}^{2}$ on the diagonal, which ensures that previously sampled points retain a predictive variance that is lower bounded by the noise level. When the noise is large enough (relative to the scale hyperparameter), this residual variance increases the likelihood of revisiting already evaluated points. When a GP is applied to fully continuous optimization problems, these exploitative revisits usually appear as evaluations at points very close to existing samples. Because the space is continuous, there is always a surrounding neighborhood to probe, which allows the optimizer to exploit locally and update the GP hyperparameters. In contrast, fully discrete domains offer no such neighborhood structure to exploit. As a result, the optimizer may repeatedly select exactly the same point within the design space. This makes resampling far more problematic in discrete spaces, where the search can become effectively stuck with no natural mechanism to escape. The resampling phenomenon is also observed in the actuator optimization study by \citet{ZHANG_actuator} when using the KR method by \citet{Eduardo_discrete} highlighting a distinct failure mode when using a GP as surrogate for BO under noisy and fully discrete-variable optimization tasks.

In experimental optimization settings, resampling due to noisy data is not necessarily undesirable. For example, anomalous datapoints may lead to the homoscedastic noise GP model fitting a larger noise hyperparameter. Resampling in these cases may lower the overall noise, especially when sampled at the anomalous condition assuming that the resampled point is not anomalous. However, there is no guarantee that resampling will eventually end unless the hyperparameters change sufficiently such that some new optimal acquisition point emerges.

To avoid resampling, we initially attempted to subtract the fitted noise hyperparameter from the diagonal of the covariance matrix, retaining only a small jitter term for numerical stability. This approach effectively prevented resampling when using the EI AF but failed for the LCB AF as shown in Figure \ref{SI_fig:0.2_LCB_subtract_repeat}, Supporting Information. To prevent resampling across all AFs we introduce a penalty mechanism that adds a large value (1e6) to the posterior mean of points that have already been sampled. Note that since the AF is minimized, the penalty term is positive. If the AF optimization were instead formulated as a maximization problem, a negative penalty would be applied. Our penalty value was arbitrarily chosen to be much larger than the standardized model posterior mean. The impact of our penalty approach can be seen in Figure \ref{fig:acq_repeat} d)-f) with the penalty method resulting in the AF selecting a new input point instead of resampling.


Even with a small fixed initialization of the model noise hyperparameter of 1e-3, resampling still occurred in the fully discrete \textit{BS} benchmark functions. To mitigate this behavior, the penalty approach was applied across all benchmarks. Since all benchmark functions were noiseless, resampling would not have provided additional information to the model.

\subsection{Mitigating Local Minima Trapping}\label{DUST_trapping}
For the \textit{DUST1} and \textit{DUST2} benchmark problems, we employed a modified BO workflow to mitigate against constant trapping in local minima which we call the modified AF (mAF) approach. Owing to the highly discretized nature of the objective landscapes, the GP surrogate model may otherwise repeatedly sample points that are extremely close in the continuous subspace while keeping the non-continuous variables fixed. Such behavior is undesirable in real-world BO settings, where each acquisition is expensive or time-consuming. Repeatedly querying near-duplicate points in the design space yields uninformative evaluations, motivating the need for this mitigation strategy for our generalized PR method.

For our mAF appraoch we simply imposed a minimum euclidean distance threshold between the suggested candidate point and previously sampled. If the AF generates a candidate point below this threshold, the candidate is kept and a purely exploratory AF will be used instead for the next BO iteration. The exploratory AF simply returns the point in design space with the largest model uncertainty. For \textit{DUST1} this distance threshold was set to 0.1 and 0.05 for \textit{DUST2}. More elaborate decision-making systems are often used in real-world BO, such as HITL approaches that incorporate expert heuristics between each BO iteration \cite{tiihonen2022mHITL}.

\section{Results}


\subsection{Butternut Squash Benchmark Results}

\begin{figure}[H]
    \centering
    \includegraphics[width=1\linewidth]{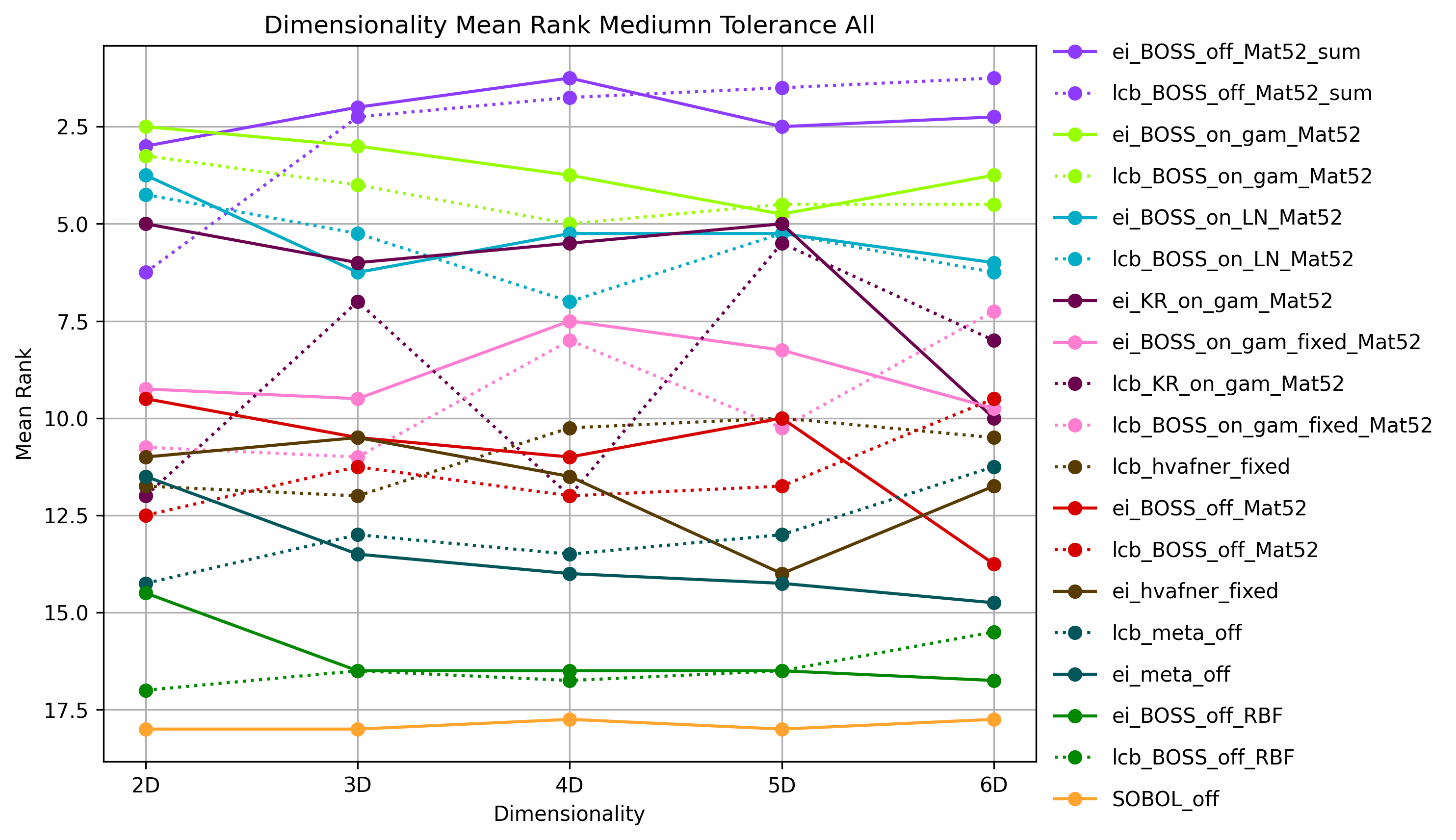}
    \caption{Plot showing the mean ranks in terms of composite score (as defined in Eg.\ref{eq:Composite score}) of all models across each dimension of the \textit{BS} benchmark function.}
    \label{fig:all_rank_Dim_medium}
\end{figure}
\begin{figure}[H]
    \centering
    \includegraphics[width=1\linewidth]{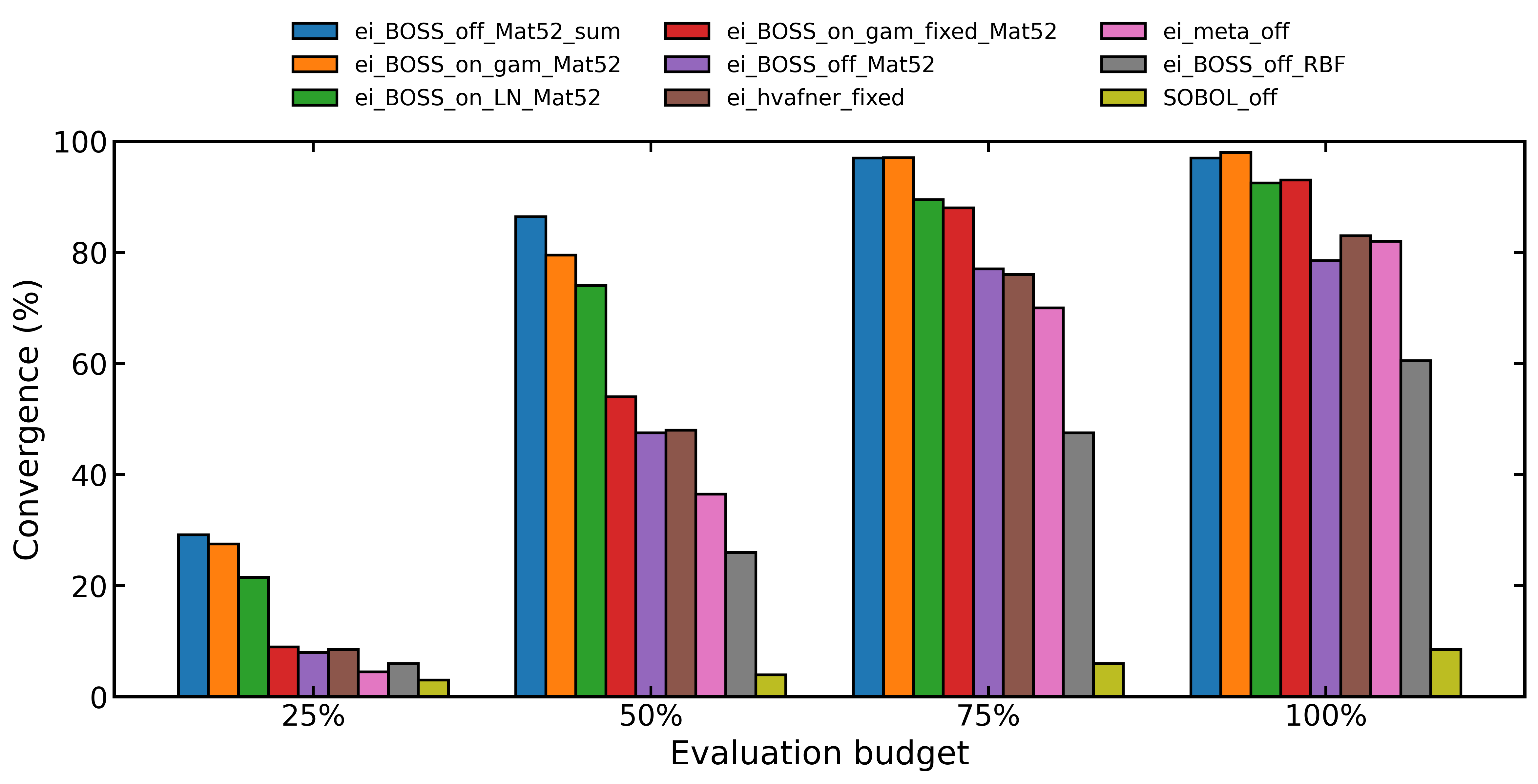}
    \caption{Histogram illustrating absolute convergence performance, measured as the percentage of converged runs aggregated over all 20 \textit{BS} benchmark instances (10 runs per model per instance, each with different Sobol initializations), for all models (excluding KR) using the EI AF. Convergence defined under our medium tolerance criterion counts are pooled across all 20 instances and evaluated under relative budget constraints of 25\%, 50\%, 75\%, and 100\%, where each percentage corresponds to the respective fraction of the maximum iterations for each \textit{BS} variant.}
    \label{fig:all_ei_bins_medium}
\end{figure}

Figure \ref{fig:all_rank_Dim_medium} presents the mean ranks of all models, evaluated in terms of the composite score defined in Eq. \ref{eq:Composite score}, across different dimensionalities of the \textit{BS} benchmark function. This provides a relative comparison of model performance by aggregating the results across each dimension of the \textit{BS} instances. Complete results for all tolerance levels and full converegence plots can be found in subsection~\ref{SI:subsubection_BS_complementary_results} (Supporting Information). These mean rank statistics, computed from composite scores, were used as the basis for the greedy search during kernel optimization. Beyond ranking-based results, in Figure \ref{fig:all_ei_bins_medium} we illustrates absolute model performance in terms of convergence percentage. As the EI-based models outperformed their LCB counterparts in terms of composite score, the figure contains only the EI models. Results for the LCB models are shown in Figure \ref{fig:all_lcb_bins_medium} Supporting Information. Specifically, the figure shows the percentage of converged runs for each EI-based model, aggregated across all 20 \textit{BS} variants, under four relative budget levels (25\%, 50\%, 75\%, and 100\%). Convergence is defined per run and results are pooled over all Sobol initialisations, where each budget corresponds to a fixed fraction of the maximum iterations for the respective variant. The KR model were omitted from Figure \ref{fig:all_ei_bins_medium} as they only were applied to the continuous--integer cases of the BS and would therefore skew their convergence results relative to the other models.

Without any priors, the initial step of our greedy search showed that the product-form Mat\'ern-5/2 kernel (\texttt{BOSS\_off\_Mat52}) outperforms its RBF counterparts (\texttt{BOSS\_off\_RBF}) by achieving higher composite scores for both AFs. These higher composite scores are observed across all tolerance levels and dimensions, and persist when the continuous–integer and integer–discrete cases are considered separately, as shown in subsection~\ref{SI:subsubection_BS_complementary_results} (Supporting Information).


The second greedy search step revealed that the summation formulation of the Mat\'ern-5/2 kernel with no priors (\texttt{ei\_BOSS\_off\_Mat52\_sum} \& \texttt{lcb\_BOSS\_off\_Mat52\_sum}) achieved higher composite scores compared to the product formulation in all cases. However, further benchmarks presented in Section~\ref{Chem_Acq_Sect} indicate that the sum formulation is unreliable and lacks generalizability, as discussed in more detail in Section~\ref{sec_discuss_BS}. For this reason, we proceeded with the more flexible and well-established product formulation of the Mat\'ern-5/2 kernel. Within this setting, the third greedy search step revealed that the Gamma prior provides improved performance compared to the LogNormal prior in most cases, across all tolerance criteria and dimensionalities, for both the continuous--integer and integer--discrete settings.  The final greedy search step revealed that fixing the scale hyperparameter degrades performance, as evidenced by the relatively poor results of the \texttt{BOSS\_on\_gam\_fixed\_Mat52} models.


As the \texttt{ei\_KR\_on\_gam} and \texttt{lcb\_KR\_on\_gam} models are restricted to continuous and integer variables, only the continuous–integer cases of the BS benchmark presented in Section~\ref{SI:subsubection_BS_complementary_results} (Supporting Information) provide a fair basis for comparison with the other models. Within their applicable domain (continuous--integer), the KR models achieve higher composite scores than the \texttt{meta\_off} models, along with faster regret convergence. The KR and \texttt{BOSS\_on\_gam\_Mat52} models share an identical product kernel construction. However, they differ in their treatment of discrete variables. Specifically, KR employs the rounding strategy of Eduardo et al., whereas \texttt{BOSS\_on\_gam\_Mat52} uses the generalized PR formulation. Under this controlled setting, \texttt{BOSS\_on\_gam\_Mat52} consistently achieves higher composite scores than the KR models across all tolerance criteria.



As shown in Figure~\ref{fig:all_ei_bins_medium}, the \texttt{meta\_off} models (i.e., the priorless Mat\'ern-5/2 ARD kernel used by the original PR authors) exhibit lower convergence percentages compared to the other models across all dimensionalities. This trend is consistent across all tolerance levels for both continuous--integer and integer--discrete cases (see Supporting Information, Section~\ref{SI:subsubection_BS_complementary_results}). The relative performance difference is further reflected in Figure~\ref{fig:all_rank_Dim_medium}, where the top-performing models at 50\% budget achieve higher convergence than the \texttt{meta\_off} models at 75\% budget.

Similarly, the Hvarfner-style kernel formulation models (\texttt{hvarfner\_fixed}) show consistently lower performance across all tolerance criteria. Even in the six-dimensional \textit{BS} setting, these models do not attain higher composite scores than the other approaches (see Supporting Information, Section~\ref{SI:subsubection_BS_complementary_results}).

While the performance of most models varies with dimensionality, domain type (i.e, continuous–integer or discrete only), and slightly across the tolerance criteria, the top four models remain consistent. Across all tolerance levels, dimensions, and domain types, the \texttt{BOSS\_on\_gam} and \texttt{meta\_off} models with the EI and LCB acquisition functions consistently rank highest, as shown in the composite score plots and tables in Section~\ref{SI:subsubection_BS_complementary_results} (Supporting Information). Instead of relying on our defined convergence score ranking metrics, one can also visually inspect all the convergence plots provided in section \ref{SI_subsec:butternutsquash} and see that the four aforementioned models do indeed offer the best overall convergence performances. 

Across all model types in our discretized benchmarks, we found that EI generally outperforms LCB in terms of convergence efficiency (see Supporting Information Section~\ref{SI:subsubection_BS_complementary_results}).

\clearpage

\subsection{Chemistry \& Actuator Benchmark Results}\label{Chem_Acq_Sect}
\begin{figure}[H]
  \centering
  \includegraphics[width=1\linewidth]{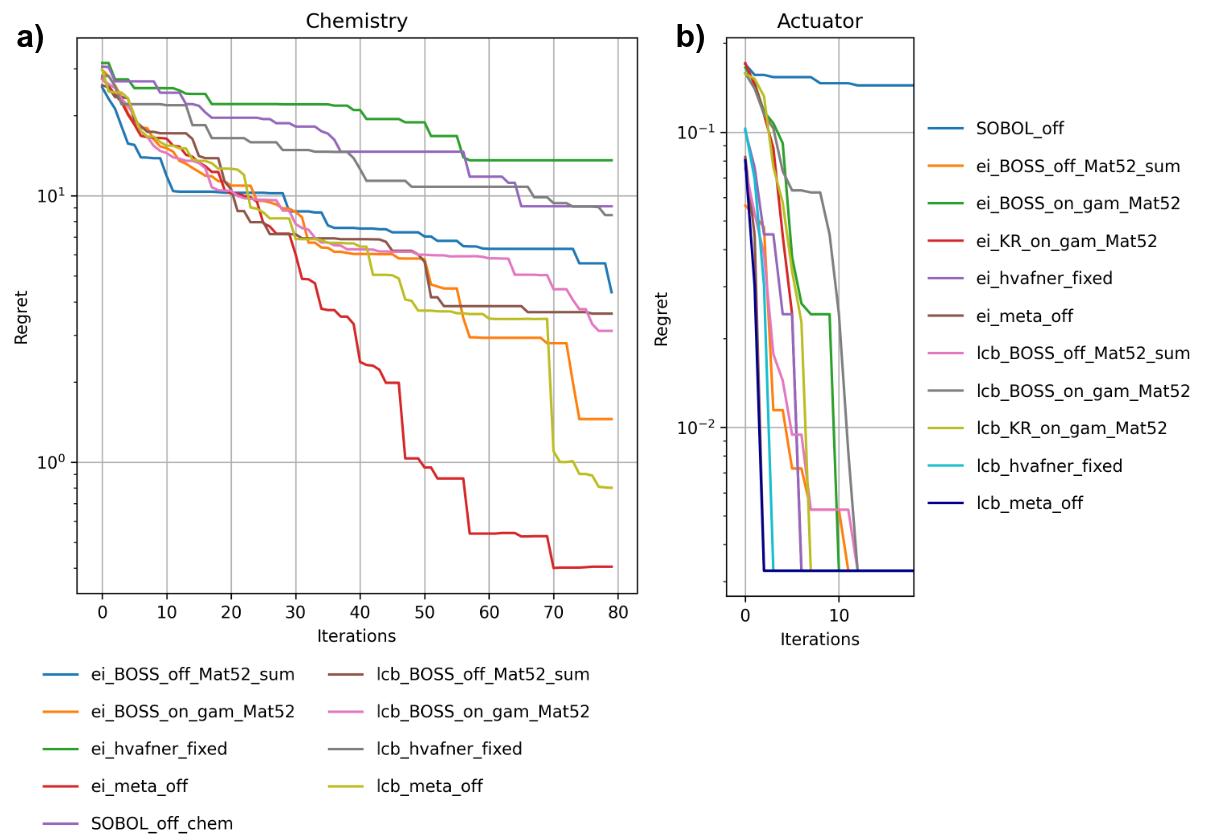}
  \caption{Mean convergence plots of specific chosen models on the \textit{Chemistry} a) and \textit{Actuator} b) benchmark functions.}
  \label{fig:chem_act}
\end{figure}

The mean convergence plots of each model for the \textit{Chemistry} and \textit{Actuator} benchmarks are shown in Figure \ref{fig:chem_act}. All models converged rapidly on the \textit{Actuator} problem.




The \textit{Chemistry} benchmark shows that \texttt{BOSS\_off\_Mat52\_sum} exhibit much poorer composite score performance values compared to the other models as seen in Table~\ref{tab:SI:composite_score_chem_medtol} and Figure~\ref{fig:chem_act} a). While the sum-formulation \texttt{BOSS\_off\_Mat52\_sum} model efficiently converged on the \texttt{BS} landscape, it failed to generalize to the \textit{Chemistry} benchmark. Only under the loose criterion did one out of the 10 Sobol runs converge as shown in Table \ref{tab:SI:composite_score_chem_lowtol} (Supporting Information).

The \textit{Chemistry} benchmark differs from the \textit{BS} benchmarks in that it contains only continuous and categorical variables, thereby increasing the variability of the benchmark set. Although from Figure \ref{fig:chem_act} a) it may seem that the \texttt{ei\_meta\_off} model has by far the fastest convergence, the composite scores reveal that the \texttt{ei\_BOSS\_on\_gam} model has a comparable score under strict tolerance and higher score under the medium and loose tolerances as seen in Tables \ref{tab:SI:composite_score_chem_lowtol}, \ref{tab:SI:composite_score_chem_medtol} and \ref{tab:SI:composite_score_chem_hightol} (Supporting Information). The similar composite scores arise because the mean convergence of the \texttt{ei\_BOSS\_on\_gam}  model is skewed by an outlier (seed 4), while most other runs converge rapidly. In contrast, the \texttt{ei\_meta\_off} model has two runs (seeds 3 and 7) that converge to higher regret values than \texttt{ei\_BOSS\_on\_gam}, as shown in Figures~\ref{fig:SI:meta_chem_runs} and~\ref{fig:SI:mat_chem_runs} (Supporting Information). In addition to eliminating the need for manual inspection of numerous convergence plots, these observations reveal another important advantage of the composite score. By quantifying the number of converged runs under different tolerance criteria, anomalous cases that could bias the mean regret convergence can be identified and accounted for, effects that visual inspection of the mean regret convergence plots alone fails to detect. The \texttt{ei\_meta\_off} kernel was originally benchmarked on this dataset and shown to outperform several state-of-the-art approaches, yet \texttt{ei\_BOSS\_on\_gam} achieves comparable performance when considering both the convergence curves and composite scores as shown in \ref{SI:subsec:chem}. Moreover, unlike the \texttt{ei\_meta\_off} kernel, \texttt{ei\_BOSS\_on\_gam} also performs strongly across the \textit{BS} benchmarks. The Hvarfner style kernel exhibits by far the lowest composite scores, with a pure Sobol sampling sequence offering comparable results. All other models however, offer higher composite score values than Sobol sampling of the design space.

All models converged rapidly on the \textit{Actuator} benchmark problem.

\subsection{DUST1 \& DUST2 Benchmark Results}

Both the \textit{Chemistry} and \textit{BS} benchmarks demonstrate that the \texttt{BOSS\_on\_gam} model kernel formulation performs well on a variety of different mixed-variable problems and objective landscapes compared to that  used by \citet{Daulton2022} in their original PR implementation (\texttt{meta\_off}) and our sum formulation (\texttt{BOSS\_off\_Mat52\_sum}). By considering all benchmark problems, and their tolerance criteriam \texttt{ei\_BOSS\_on\_gam} emerged as the best overall performing model in terms of composite score. We therefore focus exclusively on this model in the following analysis.

\begin{figure}[H]
    \centering \includegraphics[width=1\linewidth]{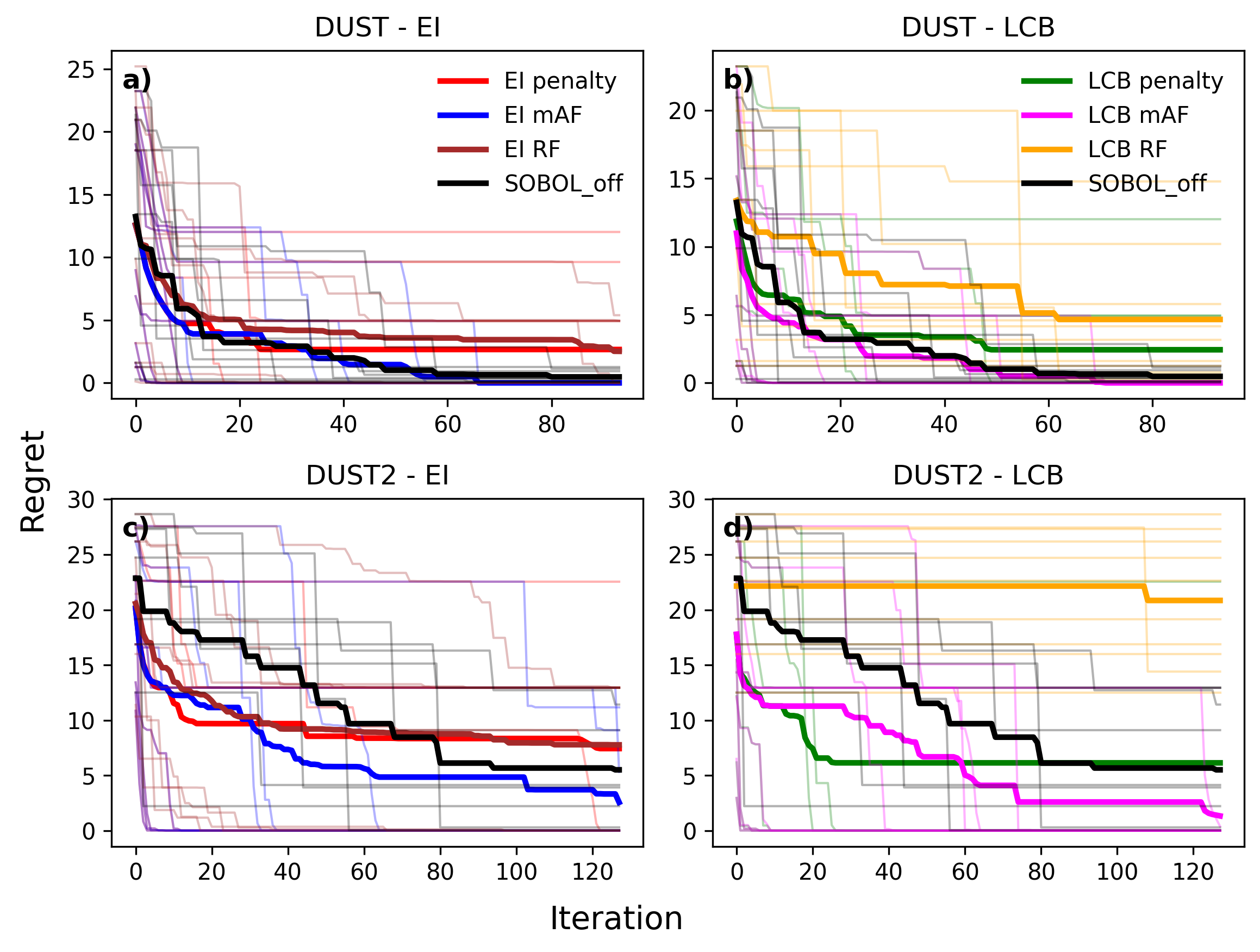}
    \caption{Convergence plots of the \texttt{BOSS\_on\_gam} models on the \textit{DUST1} and \textit{DUST2} benchmark functions using only the penalty method (Red for the EI AF \& Green for the LCB AF) versus the penalty + modified AF (mAF) approach (Blue \& Magenta). Sobol sampling is shown in black, and the RF model in Maroon and Orange. The bold curves are the mean model convergences from their 10 different color-coded Sobol point initiated runs.}
    \label{fig:HITL_explore}
\end{figure}

We tested \texttt{ei\_BOSS\_on\_gam} on the \textit{DUST1} and \textit{DUST2} benchmark functions with convergences shown in Figure \ref{fig:HITL_explore} and composite scores shown in Sections \ref{SI:subsec:DUST1} and \ref{SI:subsec:DUST2} (Supporting Information). We observe that, our penalty-based \texttt{ei\_BOSS\_on\_gam} model gets stuck into local minimas where it repeatedly samples nearby continuous values (see Figure \ref{fig:SI:explore_seed9}, Supporting Information) leading to Sobol sampling eventually converging to lower regret values on both \textit{DUST1} and \textit{DUST2}. The same behaviour also occurs when using the LCB AF as seen in Figure \ref{fig:HITL_explore} b). Overall, the EI and LCB AFs exhibit comparable performance with LCB converging slightly faster on \textit{DUST2}.

Convergence plots on the more complex \textit{DUST2} benchmark seen in Figure \ref{fig:HITL_explore} c) and d) show that both the EI and LCB models converge faster at the initial iterations before being overtaken by the Sobol sampling.  Although this "low-iteration model supremacy" behavior is also observed in the \textit{DUST1} case, it is accentuated in \textit{DUST2}. 

The convergence of our BO workflow using the EI AF with an RF as surrogate on both \textit{DUST1} and \textit{DUST2} is shown in Figures \ref{fig:HITL_explore} a) and c). The RF was initialized using settings as described in Section \ref{subsec:Models & Metrics}. From the convergence plots and composite scores shown in Sections \ref{SI:subsec:DUST1} and \ref{SI:subsec:DUST2} (Supporting Information), our penalty-model despite its local minima trapping, provides comparable performances to the generic RF model using both the EI and LCB AF. Although the RF model is not optimized with respect to its hyperparameters, we consider this benchmark appropriate as it reflects commonly used settings in the natural sciences. In such settings, unless some form of transfer learning is employed, there is generally insufficient data to reliably initialize the RF hyperparameters prior to BO, particularly when the number of available data points is smaller than the number of hyperparameter options in the RF.

To avoid local minima trapping while still utilizing the efficient sampling benefits of the model, we employed the mAF as described in Section \ref{DUST_trapping}. Our simple approach with fixed proximity thresholds that trigger pure exploration serves to illustrate the potential performance gains that can be achieved using our generalized PR method, even under highly discretized and discontinuous objective landscapes. The impact of our mAF approach is apparent from Figure \ref{fig:HITL_explore} with all models using this approach outperforming their purely penalty-based counterparts and the RF models on both \textit{DUST1} and \textit{DUST2} for both EI and LCB AFs.

All models under the mAF approach converged on \textit{DUST1}. On \textit{DUST2} both the purely penalty-based EI and LCB models also exhibit the "low-iteration model supremacy" behavior over their mAF counterparts (Figure \ref{fig:HITL_explore} c) and d)). 


\section{Discussion}

\subsection{Butternut Squash}\label{sec_discuss_BS}
Although the summation kernel formulation exhibits the best performance in terms of overall composite score and absolute convergence (Figures~\ref{fig:all_ei_bins_medium} and \ref{fig:all_rank_Dim_medium}), further investigation, presented in Supporting Information Section~\ref{SI:subsection:sum_invesigation}, indicates that this strong performance is primarily driven by two factors. First, additive (sum) kernels can efficiently model high-dimensional functions that decompose as sums over lower-dimensional subfunctions \citep{Rasmussen2006,Duvenaud2011}. Our \texttt{BS} function, being a modified Styblinski--Tang function, shares this additive structure, as it is defined as the sum of per-dimensional quartic terms (see Eq.~\ref{SI_eq:BS}). Therefore, the superior performance observed here is consistent with the structural alignment between the additive kernel assumption and the underlying objective, despite the added asymmetry. Second, given the same posterior means, the summation kernel produces lower posterior variances across the \textit{BS} design space compared to the product formulation, leading to more exploitative behavior. We believe that the combination of these two factors, the kernel’s ability to accurately model the objective landscape and its more exploitative behaviour, drives the rapid convergence observed on the \textit{BS} problem, although we also expect that the first factor alone would still lead to improved convergence.

However, as shown in the \textit{Chemistry} benchmark results, when the true objective does not conform to this additive structure, the resulting surrogate does not provide guaranteed generalizable performances (see Section~\ref{Chem_Acq_Sect}). From these results and the further investigations, we believe that the strong performance of the summation formulation of the Mat\'ern-5/2 models  (\texttt{BOSS\_off\_Mat52\_sum}) observed in the second greedy search step should be interpreted in the context of the underlying benchmark structure. While the sum kernel achieve higher composite scores in this setting, this behavior reflects a favorable alignment between the kernel’s structural assumptions and the additive nature of the \textit{BS} objective, rather than a generally superior modeling choice. In practice, real-world optimization objectives are treated as black boxes and rarely admit a clear or exact decomposition into independent per-dimensional components. As such, the summation kernel and its associated performance should be regarded as a best-case scenario, corresponding to a setting in which strong prior knowledge about the objective structure is implicitly available and exploited. When this assumption is violated, as demonstrated by the later benchmark results of section \ref{Chem_Acq_Sect}, the additive surrogate can become unreliable and lead to degraded optimization performance. Beyond the product kernels superior robustness to the sum formulation, it has also been successfully deployed in numerous experimental optimization studies, further supporting its suitability for our application~\cite{BOSS,ZHANG_actuator,Lofgren2022,Fang2021,Jingrui2024,Jin2022_exp_boss}. 

Our \textit{BS} benchmark finding, namely that the KR models achieve higher composite scores and faster convergence than the \texttt{meta\_off} models, differs from the results reported by \citet{Daulton2022}. One plausible explanation for this discrepancy lies in differences in the kernel formulations used across the two studies. The \textit{BS} benchmark results support this interpretation. When KR and PR are evaluated under the same kernel formulation, consistent with the experimental setup of \citet{Daulton2022}, PR outperforms KR in mixed variable optimization. This observation brings our findings back in line with those of the original study. Taken together, these results suggest that, regardless of the method used to address mixed variable optimization, kernel construction plays a central role in shaping model behavior and overall BO convergence performance. This perspective contrasts with \citet{Daulton2022}, who argue that a generic kernel formulation is sufficient for their PR implementation. In contrast, our empirical analysis shows that kernel choice can still have a meaningful impact on performance.


Even with increasing dimensionality, the relatively low composite scores observed for the Hvarfner style kernel formulation compared to other model settings may be attributed to two factors. First, the formulation appears to be primarily effective in fully continuous settings, which were the focus of the original study. Second, the dimensionality of the \texttt{BS} benchmark problems may be insufficient for the dimension-aware priors to provide a clear advantage. In the original continuous domain study, such priors only began to yield noticeable performance gains beyond six dimensions, with improvements becoming more pronounced around twelve dimensions when compared to the the same kernel construction using a standard $\mathrm{Gamma}(3,6)$ lengthscale prior. As a result, the \texttt{BS} benchmarks may not benefit from the dimension-aware kernel due to both their mixed-variable nature and relatively low dimensionality. Regardless, very high dimensional problems are often less relevant for practical optimization tasks in the natural sciences due to the curse of dimensionality. As dimensionality increases, the number of required data points grows prohibitively, rendering efficient optimization infeasible in many real-world settings where data acquisition is expensive and/or time-consuming.

The BS results also indicate that even in mixed optimization settings, EI offers better convergence than LCB as generally found in continuous domains. Despite substantial variation in performance across models, all methods consistently outperformed a purely Sobol sampling strategy. This is evidenced by significantly higher convergence percentages across all considered budget levels, as well as superior mean composite ranks as seen in Figures \ref{fig:all_ei_bins_medium} and \ref{fig:all_rank_Dim_medium}. These superior mean composite score ranks hold across all tolerance criteria and dimensionalities, for both continuous--integer and integer--discrete problem settings. Our results demonstrate the applicability of our generalised PR implementation for purely discrete domains and highlights the superior data-efficiency of utilizing a GP-based surrogate over pre-determined sampling strategies. Although our results only apply to the \textit{BS} benchmark, we believe this same behavior holds across different optimization settings typically encountered in the natural sciences.

\subsection{Chemistry \& Actuator}

Although the \texttt{ei\_meta\_off} and \texttt{ei\_BOSS\_on\_gam} kernel formulations achieve very similar composite scores under all three tolerance criteria, \texttt{ei\_BOSS\_on\_gam} also performs strongly across the \textit{BS} benchmarks. In contrast, \texttt{ei\_meta\_off} does not exhibit the same level of robustness, highlighting the former’s ability to generalize across a broader range of optimization settings.

While the sum-formulation posterior mean of the \texttt{BOSS\_off\_Mat52\_sum} model efficiently exploited the \texttt{BS} landscape, it failed to generalize to the \textit{Chemistry} benchmark. If only the \textit{BS} benchmark were used without additional tests and without prior understanding of the effect of kernel construction, the sum kernel formulation model (\texttt{BOSS\_off\_Mat52\_sum}) might have been naively identified as the most appropriate choice. Such an outcome highlights that the effectiveness of mixed-variable optimization approaches strongly depends on the objective landscape. Overall, our \textit{BS} and \textit{Chemistry} results highlight the importance of systematic benchmarking across diverse problem types to obtain reliable and generalizable conclusions. Such practices help prevent kernel constructions from inadvertently exploiting the structure of specific benchmarks, as no single formulation will perform optimally across all cases, an aspect that we believe remains insufficiently addressed in the field.

Since all models converge rapidly on the \textit{Actuator} benchmark with only minor differences, we see no meaningful comparative insights that can be drawn to explain the differences in model performances. This rapid convergence is not surprising, as the objective function is derived from the predictions of an already fitted GP model \citep{ZHANG_actuator}, and is also relatively smooth with no competing local minima. Consequently, all GP based models are able to effectively represent and optimize the resulting posterior landscape. While such benchmarks can serve as useful sanity checks for evaluating GP based optimization methods, they also highlight the potential for unfair advantage and bias when comparing against alternative probabilistic surrogate models such as RFs and BNNs for BO. Therefore, when benchmarking GPs against other surrogate models for BO, we believe it is important to avoid objective functions that are themselves derived from GP models, as these can introduce systematic bias. Such consideration were not explicitly addressed by \citet{Daulton2022} in their use of the also GP-derived \textit{Chemistry} benchmark, which may partly explain why their GP based implementations outperformed non GP surrogate methods in that specific setting.

\subsection{DUST1 \& DUST2}

Improved convergence behaviour arises from the mAF approach by preventing the model from becoming repeatedly trapped in some local minima. Instead of continuously resampling points in the vicinity of the same local minima, the mAF strategy activates the purely exploratory AF that selects the most informative and diverse point for model updates. This promotes effective exploration of the design space and enables the optimization to escape local minimas, ultimately leading to proper convergence. These effects are illustrated in Figure~\ref{fig:SI:explore_seed9} (Supporting Information), where trapping is observed in the absence of mAF, while successful escape from local minima occurs when mAF is employed. 

The faster initial convergence of the EI penalty and EI mAF models on \textit{DUST1} and \textit{DUST2} suggests that the model can navigate the highly discretized landscape more efficiently than Sobol sampling. The faster initial convergences are accentuated in \textit{DUST2} and highlights the increasing efficiency gain of BO over pre-determined sampling methods with increasing dimensionality or configurational space. However, this advantage diminishes once the model enters a local minimum and becomes trapped. The model could be “untrapped” from local minima by increasing exploration, for instance by scaling the model variance to larger values until a more distant acquisition point is sampled. However, determining an appropriate scaling factor is not straightforward. One possible approach is to define it based on a minimum Euclidean distance from previously sampled points, increasing the scaling factor until the next acquisition lies outside this minimum distance. We acknowledge that there may exist many viable ways to mitigate such local-minima trapping, and that the approach proposed in this study of switching to a purely exploratory AF is just one possible strategy.

Although our mAF approach successfully mitigated local-minima trapping and ensured proper convergence of our models on both \textit{DUST1} and \textit{DUST2} benchmarks, the pure penalty-based models still exhibited faster convergence during the initial iterations (low-iteration model supremacy), except for the \textit{DUST1}-LCB case. The observed behaviour suggests that repeated sampling of close points at the initial iterations may be beneficial for convergence over our strict exploration imposing mAF approach. We believe that even better performance could be achieved with more sophisticated approaches, such as allowing more frequent sampling of nearby points at the initial iterations before utilizing the purely explorative AF. Another possible improvement is to employ a penalty method analogous to that used for purely discrete cases as introduced in Section \ref{Resampling fix}, extended to continuous dimensions through a Euclidean cutoff radius. 

In the context of the \textit{DUST1} and \textit{DUST2} benchmark results, we note that the corresponding optimization landscapes represent extreme edge cases and therefore serve as valuable stress tests for our implemented Generalized PR method. Despite the highly challenging and discretized nature of these landscapes, our mAF BO workflow consistently outperforms Sobol sampling, demonstrating the robustness and practical effectiveness of the proposed approach even under severe problem conditions that may be encountered in natural science optimization tasks. Although simple, the automated nature of the proposed mAF approach provides a practical and ready to use solution for deploying BO in real world experiments and especially autonomous laboratory settings, where noisy and discretized conditions would otherwise lead to repeated evaluations.

Based on our DUST benchmark results, we expect the \texttt{ei\_BOSS\_on\_gam} penalty model equipped with some simple modified or HITL BO workflow to perform well across a wide variety of mixed-variable landscapes encountered in the natural sciences. For real-world optimization tasks involving mixed variables, we recommend a more flexible HITL-style BO approach. Such an approach would involve the use of simple decision algorithms or domain expert heuristics, for example, to determine whether duplicates or otherwise closely spaced acquisition points should be accepted at each iteration. Our Genralized PR method is compatible with many existing BO extensions, including batch acquisition, multiobjective, multifidelity, and heteroscedastic noise capabilities. Our promising benchmark results motivate future efforts to combine these capabilities to tackle even more complex optimization tasks where data is scarce.

\subsection{Limitations \& Outlook}

We acknowledge that the performance of different mixed-variable surrogate models within BO, as presented in Section~\ref{Related work}, is highly problem-dependent. Key factors influencing convergence performance include the degree of discreteness in the input space, dimensionality, and the shape of the objective landscape. The design and use of our systematic \textit{BS} benchmark functions attempts to control for some of these variables, specifically by decoupling discretization and dimensionality while fixing the landscape shape. However, our approach still has its limitations, as evident from the performance of the sum-kernel formulation on the \textit{BS} functions, where it is able to exploit the per-dimensional quadratic structure. Therefore, the benchmark suite could be expanded to include a broader range of synthetic functions with varying landscape complexities, as well as a more diverse set of real-world problems. In this regard one might follow the machine vision community, where standardized datasets have successfully driven progress by enabling direct performance comparisons across models \cite{COCO}. However, we argue that this approach is not suitable for the development of BO surrogate models. Unlike vision tasks, BO typically operates in a low-data regime, where only a limited number of objective evaluations are available. In such settings, deep and highly flexible surrogate models that can represent a wide range of landscapes often cannot be trained effectively. Therefore, instead of searching for a single universally best-performing surrogate, we advocate for identifying models that are well suited to specific subclasses of optimization problems. These subclasses may differ in terms of landscape complexity, dimensionality, or the presence of discrete and categorical variables. Surrogate models should thus be selected and deployed according to the characteristics of the problem at hand, rather than assuming that one model can perform optimally across all cases.
Benchmark suites such as \texttt{COCO} \citep{LeRiche2021} and \texttt{EXPObench} \citep{EXPOBench} provide useful test functions for BO, but their coverage remains limited. \texttt{COCO} focuses exclusively on continuous domains, whereas \texttt{EXPObench} introduces some problems with mixed variables. However, neither contains benchmark functions that systematically span key objective landscape characteristics such as dimensionality, discretization, and complexity.

Looking forward, in line with the design of our \textit{BS} benchmark, we envision a structured approach to benchmarking that organizes functions within a higher-dimensional function space. In this framework, each benchmark function could be represented by a feature vector capturing key objective landscape characteristics (dimensionality, discreteness, competing minima etc.), which may be defined through heuristic mathematical equations or learned using machine learning techniques. Each function, represented by its feature vector would therefore be a point within this higher dimensional function space. After establishing the procedure for generating feature vectors, it can be applied to existing benchmarks (e.g., \texttt{COCO}, \texttt{EXPObench}) as well as future ones.

The envisioned comprehensive benchmarking campaign would first evaluate all selected BO models across the full set of benchmark functions within this function space, using a modular framework similar to that proposed by \citet{2023MCBO}. The next step would be to evaluate various methods for generating feature descriptors and identify those that best correlate with model performance trends. For example, well designed descriptors should result in clear performance clustering for each model across the function space. For example, tree-based models are expected to perform better on discontinuous landscapes, while GP-based models are more effective on smooth ones. If no such structure emerges, the descriptors are likely uninformative. In large-scale settings, machine learning could be employed to generate or refine these descriptors.

Although constructing such a high-dimensional benchmark space would be computationally expensive, we believe it would offer substantial value to BO practitioners across all domains. In real-world applications, domain experts often have some knowledge of the objective landscape and know the optimization parameters (e.g., dimensionality, discretization, bounds etc.). With access to a well-characterized function space and associated model performance data, practitioners can approximate their problem as a feature vector using the established method. They can then identify surrogate models that have demonstrated strong performance in the region of the function space corresponding to their approximated feature vector. Such a proposed benchmark campaign, including both performance metrics and computational costs, would thus enable more informed and context-aware surrogate model selection for BO.

\section{Conclusion}
The Probabilistic Reparameterization (PR) implementation by \citet{Daulton2022} is generalized here to support discrete variables. Benchmarks spanning both synthetic and real world experiments across 18 models demonstrate that the generalized PR approach performs well on discrete domains, as evaluated using both convergence metrics and composite scores. We further optimize the kernel formulation, improving upon the generic kernel used in the original implementation. 

While all benchmarked models outperform Sobol our greedy search optimized \texttt{ei\_BOSS\_on\_gam} model achieves the best overall performance and remains competitive across all benchmarks. We also introduce a simple, method agnostic penalty to address repeated sampling in GP-based BO under noisy and non continuous objectives. For highly discontinuous landscapes, we demonstrate that incorporating this penalty into a modified BO workflow leads to substantial performance improvements, as evidenced by the \textit{DUST1} and \textit{DUST2} results.

Overall, this work provides a practical and robust approach for deploying Bayesian optimization in real world settings, particularly in autonomous laboratories where noise, mixed variable spaces, and limited data are inherent challenges, and where repeated unnoticed resampling can lead to significant experimental waste.

\section{Acknowledgements}
The authors gratefully acknowledge CSC– IT Center for Science, Finland, and the Aalto Science-IT project for generous computational resources. The work was funded in part by the Deutsche Forschungsgemeinschaft (DFG, German Research Foundation) under Germany’s Excellence Strategy – EXC 2089/2 – 390776260.

\section{Data Availability}
The data and code used in this study are freely available online: 
\href{https://gitlab.com/yuhzuu/meta_pr/-/tree/Penalty-non-eq-bench?ref_type=heads}{GitLab repository}.

\clearpage
\bibliographystyle{unsrtnat}
\bibliography{references}figures 

@article{LeRiche2021,
  author    = {Rodolphe Le Riche and Victor Picheny},
  title     = {Revisiting Bayesian optimization in the light of the COCO benchmark},
  journal   = {Structural and Multidisciplinary Optimization},
  year      = {2021},
  volume    = {64},
  number    = {5},
  pages     = {3063--3087},
  doi       = {10.1007/s00158-021-02977-1},
  url       = {https://doi.org/10.1007/s00158-021-02977-1},
  issn      = {1615-1488}
}

@inproceedings{Duvenaud2011,
  author       = {Duvenaud, David and Nickisch, Hannes and Rasmussen, Carl Edward},
  title        = {Additive Gaussian Processes},
  booktitle    = {Advances in Neural Information Processing Systems},
  year         = {2011},
  volume       = {24},
  url          = {https://proceedings.neurips.cc/paper/2011/hash/7cce53cf90577442771720a370c3c723-Abstract.html}
}

@book{Rasmussen2006,
  author       = {Rasmussen, Carl Edward and Williams, Christopher K. I.},
  title        = {Gaussian Processes for Machine Learning},
  publisher    = {MIT Press},
  address      = {Cambridge, MA},
  year         = {2006},
  isbn         = {978-0-262-18253-9},
  url          = {http://www.gaussianprocess.org/gpml/}
}

@article{Kingma2014AdamAM,
  title={Adam: A Method for Stochastic Optimization},
  author={Diederik P. Kingma and Jimmy Ba},
  journal={CoRR},
  year={2014},
  volume={abs/1412.6980},
  url={https://api.semanticscholar.org/CorpusID:6628106}
}

@article{advanced_RF,
author = {Lampe, Carola and Kouroudis, Ioannis and Harth, Milan and Martin, Stefan and Gagliardi, Alessio and Urban, Alexander S.},
title = {Rapid Data-Efficient Optimization of Perovskite Nanocrystal Syntheses through Machine Learning Algorithm Fusion},
journal = {Advanced Materials},
volume = {35},
number = {16},
pages = {2208772},
doi = {https://doi.org/10.1002/adma.202208772},
url = {https://advanced.onlinelibrary.wiley.com/doi/abs/10.1002/adma.202208772},
eprint = {https://advanced.onlinelibrary.wiley.com/doi/pdf/10.1002/adma.202208772},
year = {2023}
}

@inproceedings{
tiihonen2022mHITL,
title={More trustworthy Bayesian optimization of materials properties by adding human into the loop},
author={Armi Tiihonen and Louis Filstroff and Petrus Mikkola and Emma Lehto and Samuel Kaski and Milica Todorovi{\'c} and Patrick Rinke},
booktitle={AI for Accelerated Materials Design NeurIPS 2022 Workshop},
year={2022},
url={https://openreview.net/forum?id=JQSzcd_Zc62}
}

@InProceedings{Cat_kernel_Ru,
  title = 	 {{B}ayesian Optimisation over Multiple Continuous and Categorical Inputs},
  author =       {Ru, Binxin and Alvi, Ahsan and Nguyen, Vu and Osborne, Michael A. and Roberts, Stephen},
  booktitle = 	 {Proceedings of the 37th International Conference on Machine Learning},
  pages = 	 {8276--8285},
  year = 	 {2020},
  editor = 	 {III, Hal Daumé and Singh, Aarti},
  volume = 	 {119},
  series = 	 {Proceedings of Machine Learning Research},
  month = 	 {13--18 Jul},
  publisher =    {PMLR},
  url = 	 {https://proceedings.mlr.press/v119/ru20a.html}
}

@article{Cook_gamma,
author = {Cook, John},
year = {2010},
month = {01},
pages = {},
title = {Determining distribution parameters from quantiles},
journal = {UT MD Anderson Cancer Center Department of Biostatistics Working Paper Series}
}

@article{Daulton2022,
  author = {Daulton, Samuel and Wan, Xingchen and Eriksson, David and Balandat, Maximilian and Osborne, Michael A. and Bakshy, Eytan},
  title = {Bayesian Optimization over Discrete and Mixed Spaces via Probabilistic Reparameterization},
  journal = {arXiv preprint arXiv:2210.10199},
  year = {2022},
  url = {https://arxiv.org/abs/2210.10199},
  eprint = {2210.10199},
}

@InProceedings{hvarfner,
  title = 	 {Vanilla {B}ayesian Optimization Performs Great in High Dimensions},
  author =       {Hvarfner, Carl and Hellsten, Erik Orm and Nardi, Luigi},
  booktitle = 	 {Proceedings of the 41st International Conference on Machine Learning},
  pages = 	 {20793--20817},
  year = 	 {2024},
  editor = 	 {Salakhutdinov, Ruslan and Kolter, Zico and Heller, Katherine and Weller, Adrian and Oliver, Nuria and Scarlett, Jonathan and Berkenkamp, Felix},
  volume = 	 {235},
  series = 	 {Proceedings of Machine Learning Research},
  month = 	 {21--27 Jul},
  publisher =    {PMLR},
  url = 	 {https://proceedings.mlr.press/v235/hvarfner24a.html}
}

@article{BO_exp_worflow,
  author    = {Chuan He and Martin Singull and T. Jesper Jacobsson},
  title     = {Bayesian Optimisation for the Experimental Sciences: A Practical Guide to Data-Efficient Optimisation of Laboratory Workflows},
  journal   = {Advanced Intelligent Systems},
  year      = {2026},
  volume    = {n/a},
  number    = {n/a},
  pages     = {e202501149},
  publisher = {John Wiley \& Sons, Ltd},
  doi       = {10.1002/aisy.202501149},
  url       = {https://doi.org/10.1002/aisy.202501149},
  issn      = {2640-4567}
}

@misc{2023MCBO,
      title={Framework and Benchmarks for Combinatorial and Mixed-variable Bayesian Optimization}, 
      author={Kamil Dreczkowski and Antoine Grosnit and Haitham Bou Ammar},
      year={2023},
      eprint={2306.09803},
      archivePrefix={arXiv},
      primaryClass={cs.LG},
      url={https://arxiv.org/abs/2306.09803}, 
}

@article{Eduardo_discrete,
title = {Dealing with categorical and integer-valued variables in Bayesian Optimization with Gaussian processes},
journal = {Neurocomputing},
volume = {380},
pages = {20-35},
year = {2020},
issn = {0925-2312},
doi = {https://doi.org/10.1016/j.neucom.2019.11.004},
url = {https://www.sciencedirect.com/science/article/pii/S0925231219315619},
author = {Eduardo C. Garrido-Merchán and Daniel Hernández-Lobato}
}

@Article{BOSS,
author={Todorovi{\'{c}}, Milica
and Gutmann, Michael U.
and Corander, Jukka
and Rinke, Patrick},
title={Bayesian inference of atomistic structure in functional materials},
journal={npj Computational Materials},
year={2019},
month={3},
day={18},
volume={5},
number={1},
pages={35},
issn={2057-3960},
doi={10.1038/s41524-019-0175-2},
url={https://doi.org/10.1038/s41524-019-0175-2}
}

@article{Shields2021,
  author       = {Benjamin J. Shields and Jason Stevens and Jun Li and Marvin Parasram and Farhan Damani and Jesus I. Martinez Alvarado and Jacob M. Janey and Ryan P. Adams and Abigail G. Doyle},
  title        = {Bayesian reaction optimization as a tool for chemical synthesis},
  journal      = {Nature},
  year         = {2021},
  volume       = {590},
  number       = {7844},
  pages        = {89--96},
  doi          = {10.1038/s41586-021-03213-y},
  url          = {https://doi.org/10.1038/s41586-021-03213-y},
  issn         = {1476-4687}
}

@article{ZHANG_actuator,
title = {Data-efficient optimization of thermally-activated polymer actuators through machine learning},
journal = {Materials \& Design},
volume = {253},
pages = {113908},
year = {2025},
issn = {0264-1275},
doi = {https://doi.org/10.1016/j.matdes.2025.113908},
url = {https://www.sciencedirect.com/science/article/pii/S0264127525003284},
author = {Yuhao Zhang and Maija Vaara and Azin Alesafar and Duc Bach Nguyen and Pedro Silva and Laura Koskelo and Jussi Ristolainen and Matthias Stosiek and Joakim Löfgren and Jaana Vapaavuori and Patrick Rinke}
}

@article{Lofgren2022,
  author    = {Joakim Löfgren and Dmitry Tarasov and Taru Koitto and Patrick Rinke and Mikhail Balakshin and Milica Todorović},
  title     = {Machine Learning Optimization of Lignin Properties in Green Biorefineries},
  journal   = {ACS Sustainable Chemistry \& Engineering},
  year      = {2022},
  volume    = {10},
  number    = {29},
  pages     = {9469--9479},
  publisher = {American Chemical Society},
  doi       = {10.1021/acssuschemeng.2c01895},
  url       = {https://doi.org/10.1021/acssuschemeng.2c01895}
}

@article{Fang2021,
  author    = {Lincan Fang and Esko Makkonen and Milica Todorović and Patrick Rinke and Xi Chen},
  title     = {Efficient Amino Acid Conformer Search with Bayesian Optimization},
  journal   = {Journal of Chemical Theory and Computation},
  year      = {2021},
  volume    = {17},
  number    = {3},
  pages     = {1955--1966},
  publisher = {American Chemical Society},
  doi       = {10.1021/acs.jctc.0c00648},
  url       = {https://doi.org/10.1021/acs.jctc.0c00648},
  issn      = {1549-9618}
}

@article{Jingrui2024,
author = {Li, Jingrui and Pan, Fang and Zhang, Guo-Xu and Liu, Zenghui and Dong, Hua and Wang, Dawei and Jiang, Zhuangde and Ren, Wei and Ye, Zuo-Guang and Todorović, Milica and Rinke, Patrick},
title = {Structural Disorder by Octahedral Tilting in Inorganic Halide Perovskites: New Insight with Bayesian Optimization},
journal = {Small Structures},
volume = {5},
number = {11},
pages = {2400268},
doi = {https://doi.org/10.1002/sstr.202400268},
url = {https://onlinelibrary.wiley.com/doi/abs/10.1002/sstr.202400268},
eprint = {https://onlinelibrary.wiley.com/doi/pdf/10.1002/sstr.202400268},
year = {2024}
}

@article{Geurts2006,
  author    = {Pierre Geurts and Damien Ernst and Louis Wehenkel},
  title     = {Extremely randomized trees},
  journal   = {Machine Learning},
  year      = {2006},
  volume    = {63},
  number    = {1},
  pages     = {3--42},
  doi       = {10.1007/s10994-006-6226-1},
  url       = {https://doi.org/10.1007/s10994-006-6226-1},
  issn      = {1573-0565}
}

@Article{Zhang2022_LVGP_f_vs_b,
author={Zhang, Hengrui
and Chen, Wei (Wayne)
and Iyer, Akshay
and Apley, Daniel W.
and Chen, Wei},
title={Uncertainty-aware mixed-variable machine learning for materials design},
journal={Scientific Reports},
year={2022},
month={Nov},
day={17},
volume={12},
number={1},
pages={19760},
issn={2045-2322},
doi={10.1038/s41598-022-23431-2},
url={https://doi.org/10.1038/s41598-022-23431-2}
}

@article{EXPOBench,
title = {Benchmarking surrogate-based optimisation algorithms on expensive black-box functions},
journal = {Applied Soft Computing},
volume = {147},
pages = {110744},
year = {2023},
issn = {1568-4946},
doi = {https://doi.org/10.1016/j.asoc.2023.110744},
url = {https://www.sciencedirect.com/science/article/pii/S1568494623007627},
author = {Laurens Bliek and Arthur Guijt and Rickard Karlsson and Sicco Verwer and Mathijs {de Weerdt}}
}

@article{CuestaRamirez2022,
  author = {Cuesta Ramirez, Jhouben and Le Riche, Rodolphe and Roustant, Olivier and Perrin, Guillaume and Durantin, Cédric and Glière, Alain},
  title = {A comparison of mixed-variables Bayesian optimization approaches},
  journal = {Advanced Modeling and Simulation in Engineering Sciences},
  year = {2022},
  volume = {9},
  number = {1},
  pages = {6},
  doi = {10.1186/s40323-022-00218-8},
  url = {https://doi.org/10.1186/s40323-022-00218-8},
  sn = {2213-7467},
  doi = {10.1186/s40323-022-00218-8}
}

@misc{COSMOPOLITAN,
      title={Think Global and Act Local: Bayesian Optimisation over High-Dimensional Categorical and Mixed Search Spaces}, 
      author={Xingchen Wan and Vu Nguyen and Huong Ha and Binxin Ru and Cong Lu and Michael A. Osborne},
      year={2021},
      eprint={2102.07188},
      archivePrefix={arXiv},
      primaryClass={stat.ML},
      url={https://arxiv.org/abs/2102.07188}, 
}

@book{DOE_montgomery2017,
  title={Design and Analysis of Experiments},
  author={Montgomery, D.C.},
  isbn={9781119113478},
  lccn={2017002997},
  url={https://books.google.fi/books?id=Py7bDgAAQBAJ},
  year={2017},
  publisher={John Wiley \& Sons, Incorporated}
}

@Article{PBNN,
author ="Allec, Sarah I. and Ziatdinov, Maxim",
title  ="Active and transfer learning with partially Bayesian neural networks for materials and chemicals",
journal  ="Digital Discovery",
year  ="2025",
volume  ="4",
issue  ="5",
pages  ="1284-1297",
publisher  ="RSC",
doi  ="10.1039/D5DD00027K",
url  ="http://dx.doi.org/10.1039/D5DD00027K"}

@article{OLIVIERBNN,
title = {Bayesian neural networks for uncertainty quantification in data-driven materials modeling},
journal = {Computer Methods in Applied Mechanics and Engineering},
volume = {386},
pages = {114079},
year = {2021},
issn = {0045-7825},
doi = {https://doi.org/10.1016/j.cma.2021.114079},
url = {https://www.sciencedirect.com/science/article/pii/S0045782521004102},
author = {Audrey Olivier and Michael D. Shields and Lori Graham-Brady}
}

@Inbook{Chen2009,
author="Chen, Lei",
editor="LIU, LING
and {\"O}ZSU, M. TAMER",
title="Curse of Dimensionality",
bookTitle="Encyclopedia of Database Systems",
year="2009",
publisher="Springer US",
address="Boston, MA",
pages="545--546",
isbn="978-0-387-39940-9",
doi="10.1007/978-0-387-39940-9_133",
url="https://doi.org/10.1007/978-0-387-39940-9_133"
}

@InProceedings{hvafner,
  title = 	 {Vanilla {B}ayesian Optimization Performs Great in High Dimensions},
  author =       {Hvarfner, Carl and Hellsten, Erik Orm and Nardi, Luigi},
  booktitle = 	 {Proceedings of the 41st International Conference on Machine Learning},
  pages = 	 {20793--20817},
  year = 	 {2024},
  editor = 	 {Salakhutdinov, Ruslan and Kolter, Zico and Heller, Katherine and Weller, Adrian and Oliver, Nuria and Scarlett, Jonathan and Berkenkamp, Felix},
  volume = 	 {235},
  series = 	 {Proceedings of Machine Learning Research},
  month = 	 {21--27 Jul},
  publisher =    {PMLR},
  url = 	 {https://proceedings.mlr.press/v235/hvarfner24a.html}
}

@article{Armi_Perovskite,
title = {A data fusion approach to optimize compositional stability of halide perovskites},
journal = {Matter},
volume = {4},
number = {4},
pages = {1305-1322},
year = {2021},
issn = {2590-2385},
doi = {https://doi.org/10.1016/j.matt.2021.01.008},
url = {https://www.sciencedirect.com/science/article/pii/S2590238521000084},
author = {Shijing Sun and Armi Tiihonen and Felipe Oviedo and Zhe Liu and Janak Thapa and Yicheng Zhao and Noor Titan P. Hartono and Anuj Goyal and Thomas Heumueller and Clio Batali and Alex Encinas and Jason J. Yoo and Ruipeng Li and Zekun Ren and I. Marius Peters and Christoph J. Brabec and Moungi G. Bawendi and Vladan Stevanovic and John Fisher and Tonio Buonassisi},
}

@Article{Walsh_BO,
  author    = {Yifan Wu and Aron Walsh and Alex M. Ganose},
  title     = {Race to the bottom: Bayesian optimisation for chemical problems},
  journal   = {Digital Discovery},
  year      = {2024},
  volume    = {3},
  number    = {6}, 
  pages     = {1086--1100},
  publisher = {RSC},
  doi       = {10.1039/D3DD00234A},
  url       = {http://dx.doi.org/10.1039/D3DD00234A},
}

@article{BO_opt_advanced,
author = {Lampe, Carola and Kouroudis, Ioannis and Harth, Milan and Martin, Stefan and Gagliardi, Alessio and Urban, Alexander S.},
title = {Rapid Data-Efficient Optimization of Perovskite Nanocrystal Syntheses through Machine Learning Algorithm Fusion},
journal = {Advanced Materials},
volume = {35},
number = {16},
pages = {2208772},
doi = {https://doi.org/10.1002/adma.202208772},
url = {https://onlinelibrary.wiley.com/doi/abs/10.1002/adma.202208772},
eprint = {https://onlinelibrary.wiley.com/doi/pdf/10.1002/adma.202208772},
year = {2023}
}

@article{BO_Glopt2,
  title = {Efficient Global Structure Optimization with a Machine-Learned Surrogate Model},
  author = {Bisbo, Malthe K. and Hammer, Bj\o{}rk},
  journal = {Phys. Rev. Lett.},
  volume = {124},
  issue = {8},
  pages = {086102},
  numpages = {6},
  year = {2020},
  month = {02},
  publisher = {American Physical Society},
  doi = {10.1103/PhysRevLett.124.086102},
  url = {https://link.aps.org/doi/10.1103/PhysRevLett.124.086102}
}

@book{rasmussen2006gaussian,
  title={Gaussian Processes for Machine Learning},
  author={Rasmussen, Carl Edward and Williams, Christopher K. I.},
  year={2006},
  publisher={MIT Press},
  address={Cambridge, MA},
  isbn={026218253X},
  url={http://www.GaussianProcess.org/gpml}
}

@misc{COCO,
      title={Microsoft COCO: Common Objects in Context}, 
      author={Tsung-Yi Lin and Michael Maire and Serge Belongie and Lubomir Bourdev and Ross Girshick and James Hays and Pietro Perona and Deva Ramanan and C. Lawrence Zitnick and Piotr Dollár},
      year={2015},
      eprint={1405.0312},
      archivePrefix={arXiv},
      primaryClass={cs.CV},
      url={https://arxiv.org/abs/1405.0312}, 
}

@article{Ryzhov2016,
  author  = {Ryzhov, Ilya O.},
  title   = {On the Convergence Rates of Expected Improvement Methods},
  journal = {Operations Research},
  year    = {2016},
  volume  = {64},
  number  = {6},
  pages   = {1515--1528},
  doi     = {10.1287/opre.2016.1537}
}

@article{HennigSchuler2012,
  author  = {Hennig, Philipp and Schuler, Christian J.},
  title   = {Entropy Search for Information-Efficient Global Optimization},
  journal = {Journal of Machine Learning Research},
  year    = {2012},
  volume  = {13},
  pages   = {1809--1837},
  url     = {https://www.jmlr.org/papers/volume13/hennig12a/hennig12a.pdf}
}

@incollection{Frazier2018,
  author    = {Frazier, Peter I.},
  title     = {Bayesian Optimization},
  booktitle = {INFORMS Tutorials in Operations Research},
  year      = {2018},
  publisher = {INFORMS},
  pages     = {255--278},
  url       = {https://people.orie.cornell.edu/pfrazier/bo_tutorial.pdf}
}

@article{DeAthEversonRahatFieldsend2019,
  author    = {George De Ath and Richard M. Everson and Alma A. M. Rahat and Jonathan E. Fieldsend},
  title     = {Greed is Good: Exploration and Exploitation Trade‐offs in Bayesian Optimisation},
  journal   = {arXiv preprint},
  year      = {2019},
  volume    = {abs/1911.12809},
  url       = {https://arxiv.org/abs/1911.12809}
}

@article{BARK2025,
  author    = {Toby Boyne and Jose Pablo Folch and Robert M. Lee and Behrang Shafei and Ruth Misener},
  title     = {BARK: A Fully Bayesian Tree Kernel for Black-box Optimization},
  journal   = {arXiv preprint arXiv:2503.05574},
  year      = {2025},
  url       = {https://arxiv.org/abs/2503.05574},
  doi       = {10.48550/arXiv.2503.05574},
  note      = {Preprint available at \url{https://arxiv.org/abs/2503.05574}},
}

@article{BARTChipman2008,
  author    = {Hugh A. Chipman and Edward I. George and Robert E. McCulloch},
  title     = {BART: Bayesian additive regression trees},
  journal   = {Annals of Applied Statistics},
  year      = {2010},
  volume    = {4},
  number    = {1},
  pages     = {266--298},
  doi       = {10.1214/09-AOAS285},
  url       = {https://doi.org/10.1214/09-AOAS285},
  note      = {Preprint available at \url{https://arxiv.org/abs/0806.3286}}
}

@article{DOEGreenhill,
author = {Greenhill, Stewart and Rana, Santu and Gupta, Sunil and Vellanki, Pratibha and Venkatesh, Svetha},
year = {2020},
month = {01},
pages = {1-1},
title = {Bayesian Optimization for Adaptive Experimental Design: A Review},
volume = {PP},
journal = {IEEE Access},
doi = {10.1109/ACCESS.2020.2966228}
}

@article{SUN20211305,
title = {A data fusion approach to optimize compositional stability of halide perovskites},
journal = {Matter},
volume = {4},
number = {4},
pages = {1305-1322},
year = {2021},
issn = {2590-2385},
doi = {https://doi.org/10.1016/j.matt.2021.01.008},
url = {https://www.sciencedirect.com/science/article/pii/S2590238521000084},
author = {Shijing Sun and Armi Tiihonen and Felipe Oviedo and Zhe Liu and Janak Thapa and Yicheng Zhao and Noor Titan P. Hartono and Anuj Goyal and Thomas Heumueller and Clio Batali and Alex Encinas and Jason J. Yoo and Ruipeng Li and Zekun Ren and I. Marius Peters and Christoph J. Brabec and Moungi G. Bawendi and Vladan Stevanovic and John Fisher and Tonio Buonassisi}
}

@article{PedersenBO,
author = {Pedersen, Jack K. and Clausen, Christian M. and Krysiak, Olga A. and Xiao, Bin and Batchelor, Thomas A. A. and Löffler, Tobias and Mints, Vladislav A. and Banko, Lars and Arenz, Matthias and Savan, Alan and Schuhmann, Wolfgang and Ludwig, Alfred and Rossmeisl, Jan},
title = {Bayesian Optimization of High-Entropy Alloy Compositions for Electrocatalytic Oxygen Reduction},
journal = {Angewandte Chemie International Edition},
volume = {60},
number = {45},
pages = {24144-24152},
doi = {https://doi.org/10.1002/anie.202108116},
url = {https://onlinelibrary.wiley.com/doi/abs/10.1002/anie.202108116},
eprint = {https://onlinelibrary.wiley.com/doi/pdf/10.1002/anie.202108116},
year = {2021}
}

@article{SaninBO,
author = {Sanin, Alexey and Flowers, Jackson K. and Piotrowiak, Tobias H. and Felsen, Frederic and Merker, Leon and Ludwig, Alfred and Bresser, Dominic and Stein, Helge Sören},
title = {Integrating Automated Electrochemistry and High-Throughput Characterization with Machine Learning to Explore Si─Ge─Sn Thin-Film Lithium Battery Anodes},
journal = {Advanced Energy Materials},
volume = {15},
number = {11},
pages = {2404961},
doi = {https://doi.org/10.1002/aenm.202404961},
url = {https://advanced.onlinelibrary.wiley.com/doi/abs/10.1002/aenm.202404961},
eprint = {https://advanced.onlinelibrary.wiley.com/doi/pdf/10.1002/aenm.202404961},
year = {2025}
}

@article{Jin2022_exp_boss,
  author    = {Soo-Ah Jin and Tero K{\"a}m{\"a}r{\"a}inen and Patrick Rinke and Orlando J. Rojas and Milica Todorovi{\'c}},
  title     = {Machine learning as a tool to engineer microstructures: Morphological prediction of tannin-based colloids using Bayesian surrogate models},
  journal   = {MRS Bulletin},
  year      = {2022},
  volume    = {47},
  number    = {1},
  pages     = {29--37},
  doi       = {10.1557/s43577-021-00183-4},
  url       = {https://doi.org/10.1557/s43577-021-00183-4},
  issn      = {1938-1425}
}

@article{DimentBO,
author = {Diment, Daryna and Löfgren, Joakim and Alopaeus, Marie and Stosiek, Matthias and Cho, MiJung and Xu, Chunlin and Hummel, Michael and Rigo, Davide and Rinke, Patrick and Balakshin, Mikhail},
title = {Enhancing Lignin-Carbohydrate Complexes Production and Properties With Machine Learning},
journal = {ChemSusChem},
volume = {18},
number = {8},
pages = {e202401711},
doi = {https://doi.org/10.1002/cssc.202401711},
url = {https://chemistry-europe.onlinelibrary.wiley.com/doi/abs/10.1002/cssc.202401711},
eprint = {https://chemistry-europe.onlinelibrary.wiley.com/doi/pdf/10.1002/cssc.202401711},
year = {2025}
}

@article{MirandaBO,
    author = {Miranda-Valdez, Isaac Y. and Mäkinen, Tero and Koivisto, Juha and Alava, Mikko J.},
    title = {Bayesian optimization to infer parameters in viscoelasticity},
    journal = {Journal of Rheology},
    volume = {69},
    number = {6},
    pages = {1059-1066},
    year = {2025},
    month = {10},
    issn = {0148-6055},
    doi = {10.1122/8.0001068},
    url = {https://doi.org/10.1122/8.0001068},
    eprint = {https://pubs.aip.org/sor/jor/article-pdf/69/6/1059/20782370/1059_1_8.0001068.pdf},
}

\clearpage

\setcounter{page}{1}
\setcounter{section}{0}
\setcounter{table}{0}
\setcounter{figure}{0}
\setcounter{equation}{0}

\renewcommand{\thepage}{S\arabic{page}}
\renewcommand{\thesection}{S\arabic{section}}  
\renewcommand{\thetable}{S\arabic{table}}  
\renewcommand{\thefigure}{S\arabic{figure}}
\renewcommand{\theequation}{S\arabic{equation}}

\newpage

\begin{center}
    {\LARGE \textbf{Supporting Information}}\\[1em]
    {\large for}\\[0.5em]
    {\Large \textbf{Bayesian Optimization with Generalized Probabilistic Reparameterization for Non-Uniform Mixed Spaces}}\\[1.5em]
    {\normalsize
    Yuhao Zhang$^{1}$, Ti John$^{2}$, Matthias Stosiek$^{3}$, Patrick Rinke$^{3}$\\[0.5em]
    $^{1}$Department of Applied Physics, Aalto University, Espoo, Finland\\
    $^{2}$Department of Computer Science, Aalto University, Espoo, Finland\\
    $^{3}$School of Natural Sciences, Physics Department, Technical University of Munich, Munich, Germany
    }
\end{center}

\vspace{1em}

\section{Complementary Related Work}

\begin{figure}[H]
    \centering
    \includegraphics[width=1\linewidth]{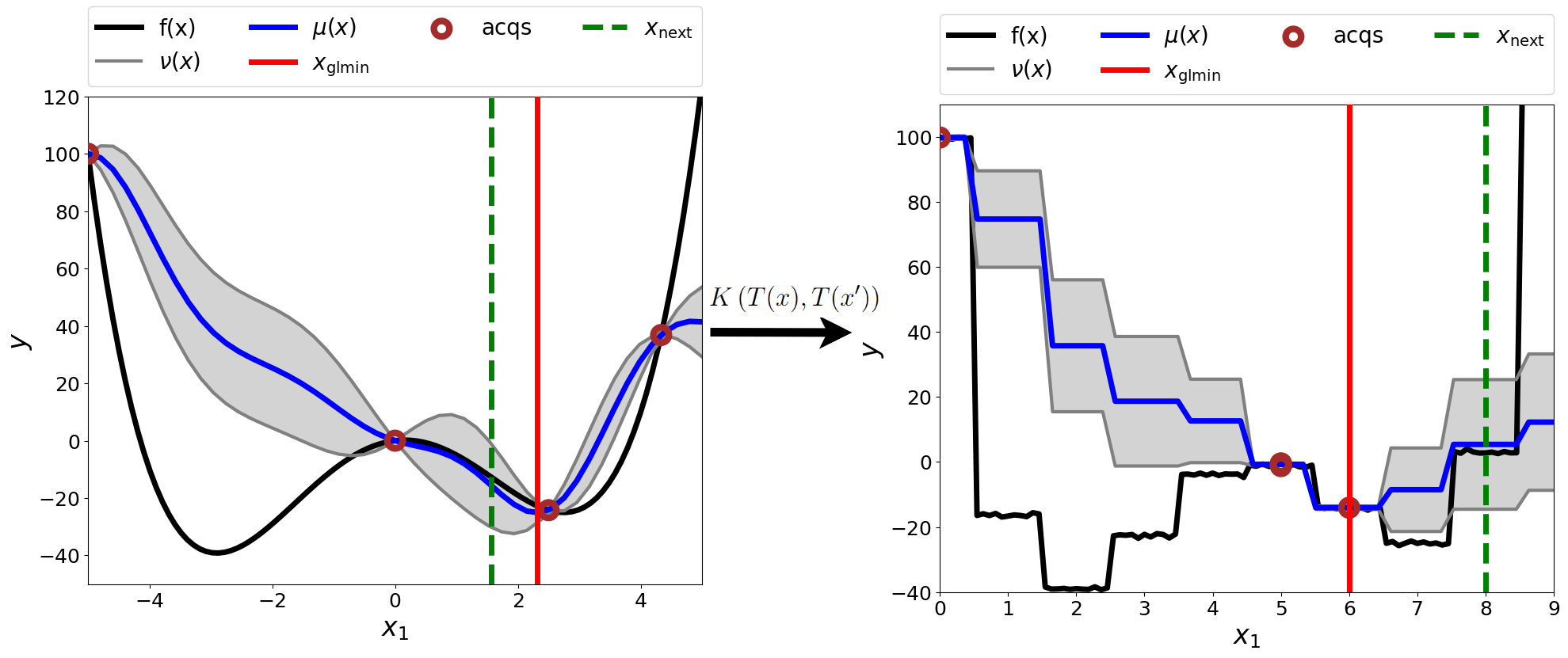}
    \caption{Effect of kernel rounding trick method by Eduardo et al.\cite{Eduardo_discrete} on otherwise continuous GP posterior}
    \label{SI_fig:KR effect}
\end{figure}

\section{Complementary Experimental Background}

\subsection{Butternut Squash Function}\label{SI_subsec:butternutsquash}

The Butternut Squash function for an input vector $\text{\textbf{x}}_{i}=\left(x_{i1},x_{i2},x_{i3},...,x_{id}\right)$ is given by:
\begin{equation}
    f\left(\text{\textbf{x}}_{i}\right)=\frac{1}{2d}\left(\sum_{j=i}^{d}\left(0.15x_{ij}^{4}-3x_{ij}^{2}+3x_{ij}\right)+\left(\sum_{k=0}^{\left\lfloor (d-1)/2\right\rfloor }0.5\left(x_{i,2k+1}+3.38763191\right)^{2}\right)\right)+12.4180436,
    \label{SI_eq:BS}
\end{equation}
where $d$ is the dimensionality.
\begin{figure}[H]
    \centering
    \includegraphics[width=1\linewidth]{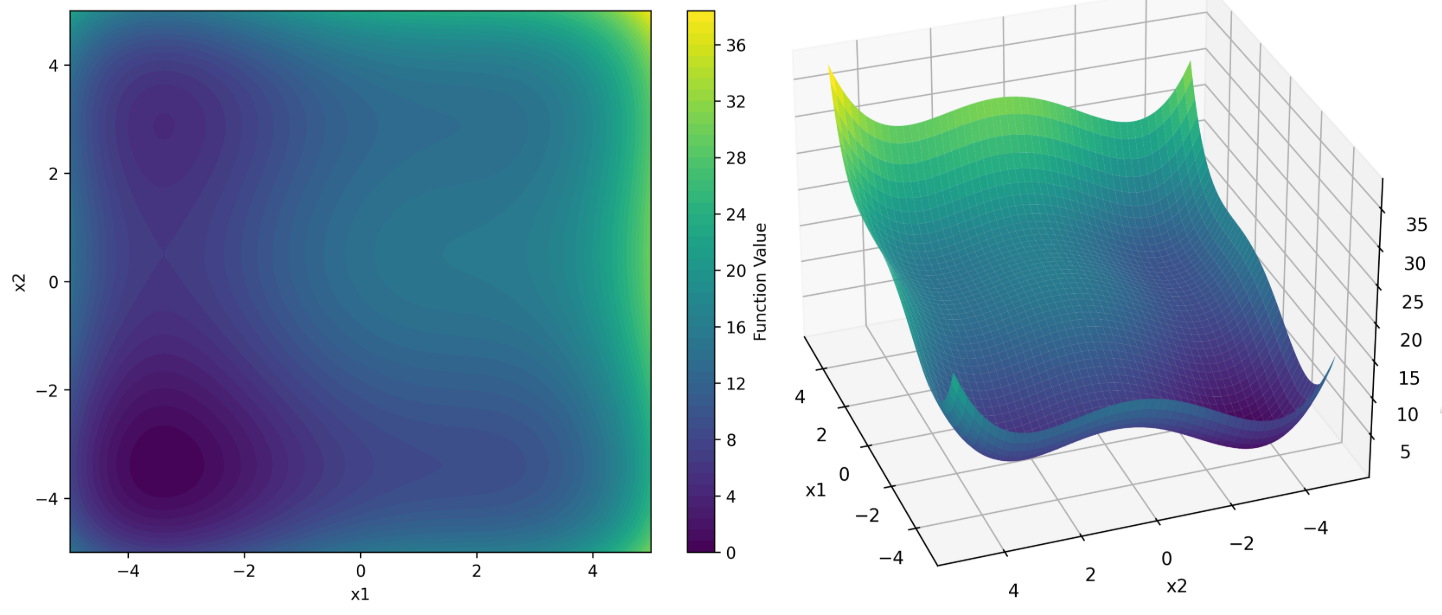}
    \caption{Two dimensional Butternut Squash function with both axis continuous as shown as contour plot (Left) and surface plot (Right).}
    \label{SI_fig:bs2d3d}
\end{figure}

\subsection{Benchmark Initialisation Information}\label{SI_subsec:SI:BO_points}

\begin{table}[H]
\centering
\begin{tabular}{lrrr}
\toprule
\textbf{Function} & \textbf{Init. Points} & \textbf{Iter. Points} & \textbf{Total Points} \\
\midrule
2D Butternut Squash & 5  & 35  & 40  \\
3D Butternut Squash & 10 & 80  & 90  \\
4D Butternut Squash & 20 & 100 & 120 \\
5D Butternut Squash & 40 & 160 & 200 \\
6D Butternut Squash & 60 & 220 & 280 \\
Actuator            & 10 & 80  & 90  \\
Chemistry           & 20 & 80  & 100 \\
DUST                & 6  & 94  & 100 \\
DUST2               & 12 & 128 & 140 \\
\bottomrule
\end{tabular}
\caption{Table showing the number of initial sobol points, iterative points, and total evaluations for each benchmark function. Each model was initialised 10 times on 10 different Sobol sequenced initial points.}
\label{tab:SI:function_BO_points}
\end{table}

Table here rows are all function types, columns are function, initpts, iterpts, sobol for all

\section{Complementary Results}

\subsection{Kernel Formulations}\label{SI_subsec:kernels}
\begin{table}[H]
\centering
\begin{tabularx}{\textwidth}{l X}
\toprule
Authors & Daulton et al.\cite{Daulton2022} \\ 
\midrule
Kernel name & meta\_off \\
\midrule
Base kernel & Mat\'ern-5/2 \\
\midrule
Formulation & 
$k_{\text{scale}}\left(k_{\text{cat}}^{\text{ARD}} \times k_{\text{cont,int}}^{\text{ARD}} \times k_{\text{bin}}^{\text{ARD}}\right) + 
k_{\text{scale}}\left(k_{\text{cat}}^{\text{ARD}} + k_{\text{cont,int}}^{\text{ARD}} + k_{\text{bin}}^{\text{ARD}}\right)$ \\
\midrule
Scale prior & None \\
\midrule
Lengthscale prior & None \\
\midrule
Comments & 
In this study with discrete variable problems, $k_{\text{cont,int}}^{\text{ARD}}$ becomes $k_{\text{cont,int,discr}}^{\text{ARD}}$. \\
\bottomrule
\end{tabularx}
\caption{Kernel formulation used within the original PR implementation by Daulton et al.}
\label{SI_tab:meta kernel}
\end{table}

\begin{table}[H]
\centering
\begin{tabularx}{\textwidth}{l X}
\toprule
Authors & Hvafner et al.\cite{hvarfner} \\
\midrule
Kernel name & hvafner\_fixed \\
\midrule
Base kernel & Mat\'ern-5/2 \\
\midrule
Formulation & 
$k_{\text{scale}}\left(k_{\text{cont,int,discr}}^{\text{ARD}}\right)$ \\
\midrule
Scale prior & 
$\sigma^{2} = 1$ \\
\midrule
Lengthscale prior & 
$\text{LN}\left(\sqrt{2} + \log\sqrt{D}, \sqrt{3}\right)$ \\
\midrule
Comments & 
Scale hyperparameter fixed to 1 (when y data is standardized). $D$ is the dimensionality of the input space. Originally designed for purely continuous cases, within this study it is generalized to mix variable problems through the generalized PR method. \\
\bottomrule
\end{tabularx}
\caption{Default Botorch kernel design using the dimensional aware priors by Hvafner et al.}
\label{SI_tab:hvafner kernel}
\end{table}

\begin{table}[H]
\centering
\begin{tabularx}{\textwidth}{l X}
\toprule
Authors & Eduardo et al.\cite{Eduardo_discrete} \\
\midrule
Kernel name & KR\_on\_gam\_Mat52 \\
\midrule
Base kernel & Mat\'ern-5/2 \\
\midrule
Formulation & 
$k_{\text{scale}}\left(k_{\text{cont}}^{\text{Prod}} \times k_{\text{int}}^{\text{Prod}}\right)$ \\
\midrule
Scale prior & 
$\text{Gam}\left(2,\left(\frac{1}{2(\max(\boldsymbol{Y}) - \min(\boldsymbol{Y}))}\right)^2\right)$ \\
\midrule
Lengthscale prior & 
$\text{Gam}(\alpha, \lambda)$ \\
\midrule
Comments & 
 This approach rounds integer-valued variables to the nearest integer within the kernel, making it incompatible with discrete variables. Additionally, the variable transformation involved in this method prevents the use of gradient-based optimization for the acquisition function. \\
\bottomrule
\end{tabularx}
\caption{BOSS-style kernel design and priors\cite{BOSS} utilizing the kernel rounding method by Eduardo et al. for handling continuous+Integer variable problems.}
\label{SI_tab:KR kernel}
\end{table}

\begin{table}[H]
\centering
\begin{tabularx}{\textwidth}{l X}
\toprule
Kernel Name & BOSS\_off\_RBF\tabularnewline
\midrule 
Base kernel & RBF\tabularnewline
\midrule 
Formulation & $k_{scale}\left(k_{cont}^{Prod}\times k_{int}^{Prod}\times k_{discr}^{Prod}\times k_{cat}^{ARD}\right)$\tabularnewline
\midrule 
Scale prior & -\tabularnewline
\midrule 
Lengthscale prior & -\tabularnewline
\midrule 
Comments & -\tabularnewline
\bottomrule
\end{tabularx}
\caption{Radial basis function kernel with product formulation and no priors}
\label{SI_tab:BOSS off RBF kernel}
\end{table}

\begin{table}[H]
\centering
\begin{tabularx}{\textwidth}{l X}
\toprule
Kernel Name & BOSS\_off\_Mat52 \\
\midrule
Base kernel & Mat\'ern-5/2 \\
\midrule
Formulation & $k_{scale}\left(k_{cont}^{Prod} \times k_{int}^{Prod} \times k_{discr}^{Prod} \times k_{cat}^{ARD}\right)$ \\
\midrule
Scale prior & - \\
\midrule
Lengthscale prior & - \\
\midrule
Comments & - \\
\bottomrule
\end{tabularx}
\caption{Mat\'ern-5/2 kernel with product formulation and no priors}
\label{tab:boss_off_mat52}
\end{table}

\begin{table}[H]
\centering
\begin{tabularx}{\textwidth}{l X}
\toprule
Kernel Name & BOSS\_off\_Mat52\_sum \\
\midrule
Base kernel & Mat\'ern-5/2 \\
\midrule
Formulation & $k_{scale}\left(k_{cont}^{sum} + k_{int}^{sum} + k_{discr}^{sum} + k_{cat}^{ARD}\right)$ \\
\midrule
Scale prior & - \\
\midrule
Lengthscale prior & - \\
\midrule
Comments & - \\
\bottomrule
\end{tabularx}
\caption{Mat\'ern-5/2 kernel with sum formulation and no priors}
\label{tab:boss_off_mat52_sum}
\end{table}

\begin{table}[H]
\centering
\begin{tabularx}{\textwidth}{l X}
\toprule
Kernel Name & BOSS\_on\_gam\_Mat52 \\
\midrule
Base kernel & Mat\'ern-5/2 \\
\midrule
Formulation & $k_{scale}\left(k_{cont}^{Prod} \times k_{int}^{Prod} \times k_{discr}^{Prod} \times k_{cat}^{ARD}\right)$ \\
\midrule
Scale prior & $\text{Gam}\left(2,\left(\frac{1}{2\left(\max(\boldsymbol{Y}) - \min(\boldsymbol{Y})\right)}\right)^{2}\right)$ \\
\midrule
Lengthscale prior & $\text{Gam}(\alpha, \lambda)$ \\
\midrule
Comments & - \\
\bottomrule
\end{tabularx}
\caption{Mat\'ern-5/2 kernel with product formulation and Gamma priors}
\label{tab:boss_on_gam_mat52}
\end{table}

\begin{table}[H]
\centering
\begin{tabularx}{\textwidth}{l X}
\toprule
Kernel Name & BOSS\_on\_LN\_Mat52 \\
\midrule
Base kernel & Mat\'ern-5/2 \\
\midrule
Formulation & $k_{scale}\left(k_{cont}^{Prod} \times k_{int}^{Prod} \times k_{discr}^{Prod} \times k_{cat}^{ARD}\right)$ \\
\midrule
Scale prior & $\text{Gam}\left(2,\left(\frac{1}{2\left(\max(\boldsymbol{Y}) - \min(\boldsymbol{Y})\right)}\right)^{2}\right)$ \\
\midrule
Lengthscale prior & $\text{LN}\left(\sqrt{2}+\log\sqrt{D}, \sqrt{3}\right)$ \\
\midrule
Comments & - \\
\bottomrule
\end{tabularx}
\caption{Mat\'ern-5/2 kernel with log-normal prior on lengthscale and gamma prior on scale hyperparmeters}
\label{tab:boss_on_ln_mat52}
\end{table}

\begin{table}[H]
\centering
\begin{tabularx}{\textwidth}{l X}
\toprule
Kernel Name & BOSS\_on\_gam\_fixed\_Mat52 \\
\midrule
Base kernel & Mat\'ern-5/2 \\
\midrule
Formulation & $k_{scale}\left(k_{cont}^{Prod} \times k_{int}^{Prod} \times k_{discr}^{Prod} \times k_{cat}^{ARD}\right)$ \\
\midrule
Scale prior & $\sigma^{2} = 1$ \\
\midrule
Lengthscale prior & $\text{Gam}(\alpha, \lambda)$ \\
\midrule
Comments & - \\
\bottomrule
\end{tabularx}
\caption{Mat\'ern-5/2 kernel with fixed output variance and Gamma prior on lengthscale}
\label{tab:boss_on_gam_fixed_mat52}
\end{table}

$k_{\text{scale}}$ denotes a scaling kernel applied to the underlying kernel. The subscripts $cat$, $cont$, $int$, and $discr$ refer to kernels defined over categorical, continuous, integer, and discrete variables, respectively. All kernels take the form of the base kernel except for the categorical kernels. The categorical kernel ($k_{\text{cat}}$) is based on the Hamming distance, following the formulation by Ru et al.~\cite{Cat_kernel_Ru}, as employed by \citet{Daulton2022}. in the original PR formulation\cite{Daulton2022}. The superscript $ARD$ indicates that the kernel employs automatic relevance determination for the corresponding variable types, assigning a separate lengthscale to each dimension. The superscript $Prod$ denotes a product kernel, where the overall kernel is constructed as the product of individual one-dimensional kernels for each dimension of the specified variable type. The shape and rate parameters, $\alpha$ and $\lambda$, are determined by specifying lower and upper quantiles using the method proposed by Cook et al.~\cite{Cook_gamma}, with $p_1 = 0.05$ and $p_2 = 0.5$ set relative to the input bounds for the given kernel.

\subsection{Tolerance Criterions}

\begin{table}[H]
\begin{center}
    \begin{tabular}{ccccc}
\toprule 
Problem & Tolerance level  & Y-range & Cont.  & Int., Discr., Cat.\tabularnewline
\midrule 
Butternut Squash & Strict & 0.1\% & 1\% & Exact\tabularnewline
\midrule 
 & Medium  & 0.5\% & 2\% & Exact\tabularnewline
\midrule 
 & Loose & 1\% & 4\% & Exact\tabularnewline
\midrule 
Chemistry & Strict & 1\% & 1\% & Exact\tabularnewline
\midrule 
 & Medium  & 2\% & 2\% & Exact\tabularnewline
\midrule 
 & Loose & 3\% & 3\% & Exact\tabularnewline
\midrule
DUST1 & Strict & 0.5\% & 0.5\% & Exact\tabularnewline
\midrule 
 & Medium  & 2\% & 2\% & Exact\tabularnewline
\midrule 
 & Loose & 3\% & 3\% & Exact\tabularnewline
\midrule 
DUST2 & Strict & 0.5\% & 0.5\% & Exact\tabularnewline
\midrule 
 & Medium  & 1\% & 1\% & Exact\tabularnewline
\midrule 
 & Loose & 5\% & 5\% & Exact\tabularnewline
\midrule 
Actuator & Single level & - & - & Exact\tabularnewline
\bottomrule
\end{tabular}
\par\end{center}

\caption{Tolerance levels used to determine convergence for each benchmark problem. The “Y-range tolerance” column specifies the allowable relative difference between the model’s minimum objective value and the global minimum, expressed as a percentage of the total objective value range. The “X-range tolerance (continuous)” column indicates the maximum allowed deviation of the continuous input parameters from their true optimal values, expressed as a percentage of the input variable range. For all tolerance levels, integer and categorical parameters must exactly match the global minimum for convergence to be declared.}
\label{SI_tab:tolerances}
\end{table}

\subsection{Sum Kernel Investigation}\label{SI:subsection:sum_invesigation}

We consider Gaussian process priors with zero mean and covariance functions defined over a $D$-dimensional input space.  In particular, we study two structured kernels built from one–dimensional base kernels $k_{d}$ with unit diagonal: the product kernel
\[
k_{\prod}(\mathbf{x},\mathbf{x}')
= \prod_{d=1}^{D} k_{d}(x_{d},x'_{d}),
\]
and the sum kernel
\[
k_{\sum}(\mathbf{x},\mathbf{x}')
= \sum_{d=1}^{D} k_{d}(x_{d},x'_{d}),
\]
each equipped with its own signal‐variance (scale) hyperparameter $\sigma_{p}^{2}$ and $\sigma_{s}^{2}$ via
\[
k_{\mathrm{prod}}(\mathbf{x},\mathbf{x}')
= \sigma_{p}^{2}\, k_{\prod}(\mathbf{x},\mathbf{x}'),
\qquad
k_{\mathrm{sum}}(\mathbf{x},\mathbf{x}')
= \sigma_{s}^{2}\, k_{\sum}(\mathbf{x},\mathbf{x}').
\]
With $k_{d}(x_{d},x_{d}) = 1$, the corresponding prior variances are
\[
k_{\mathrm{prod}}(\mathbf{x},\mathbf{x})
= \sigma_{p}^{2},
\qquad
k_{\mathrm{sum}}(\mathbf{x},\mathbf{x})
= D\sigma_{s}^{2}.
\]

\paragraph{Posterior variance and the basic inequality.}
Let $\mathbf{X} = \{\mathbf{x}_{1},\dots,\mathbf{x}_{n}\}$ denote the training inputs, $\mathbf{y}$ the observations, and $\sigma_{0}^{2}$ the observation noise variance. For a test input $\mathbf{x}_{*}$ the posterior variance under a covariance function $k$ is
\begin{equation}
\mathrm{Var}[f_{*}]
= k_{**}
 - \mathbf{k}_{*}^{\top} \bigl( \mathbf{K} + \sigma_{0}^{2}\mathbf{I} \bigr)^{-1} \mathbf{k}_{*},
\label{eq:posterior-var}
\end{equation}
where
\[
k_{**} = k(\mathbf{x}_{*},\mathbf{x}_{*}), \qquad
\mathbf{k}_{*} = \bigl(k(\mathbf{x}_{*},\mathbf{x}_{1}),\dots,k(\mathbf{x}_{*},\mathbf{x}_{n})\bigr)^{\top}, \qquad
\mathbf{K}_{ij} = k(\mathbf{x}_{i},\mathbf{x}_{j}).
\]

Let $\mathbf{K}_{\prod}$ and $\mathbf{K}_{\sum}$ denote the correlation matrices of $k_{\prod}$ and $k_{\sum}$, and $\mathbf{k}_{*,\prod}$, $\mathbf{k}_{*,\sum}$ the corresponding correlation vectors. The full covariance matrices are then $\sigma_{p}^{2}\mathbf{K}_{\prod}$ and $\sigma_{s}^{2}\mathbf{K}_{\sum}$. From \eqref{eq:posterior-var} the posterior variances under the scaled sum and product kernels are
\begin{equation}
\label{eq:posterior-var-sum}
\begin{aligned}
\mathrm{Var}_{\sum}(f_{*})
  &= \sigma_{s}^{2} k_{**}^{(\sum)}
     - \bigl(\sigma_{s}^{2}\mathbf{k}_{*,\sum}\bigr)^{\top}
       \bigl(\sigma_{s}^{2}\mathbf{K}_{\sum} + \sigma_{0}^{2}\mathbf{I}\bigr)^{-1}
       \bigl(\sigma_{s}^{2}\mathbf{k}_{*,\sum}\bigr).
\end{aligned}
\end{equation}

\begin{equation}
\label{eq:posterior-var-prod}
\begin{aligned}
\mathrm{Var}_{\prod}(f_{*})
  &= \sigma_{p}^{2} k_{**}^{(\prod)}
     - \bigl(\sigma_{p}^{2}\mathbf{k}_{*,\prod}\bigr)^{\top}
       \bigl(\sigma_{p}^{2}\mathbf{K}_{\prod} + \sigma_{0}^{2}\mathbf{I}\bigr)^{-1}
       \bigl(\sigma_{p}^{2}\mathbf{k}_{*,\prod}\bigr).
\end{aligned}
\end{equation}
At the maximal prior–variance points on the diagonal we have $k_{**}^{(\prod)} = 1$ and $k_{**}^{(\sum)} = D$. Away from such diagonal locations, the prior variance terms $k_{**}^{(\sum)}$ and $k_{**}^{(\prod)}$ are strictly smaller, meaning that the required threshold inequality is $k_{**}^{(\sum)}D\sigma_{s}^{2} - k_{**}^{(\prod)}\sigma_{p}^{2}$ in most settings. Assuming $\operatorname{diag}\!\left(\sigma_{p}^{2}\mathbf{K}_{\prod}\right)\gg\sigma_{0}^{2}\mathbf{I}$ 
and $\operatorname{diag}\!\left(\sigma_{s}^{2}\mathbf{K}_{\sum}\right)\gg\sigma_{0}^{2}\mathbf{I}$,
we approximate
\[
\bigl(\sigma_{p}^{2}\mathbf{K}_{\prod} + \sigma_{0}^{2}\mathbf{I}\bigr)^{-1}
\approx (\sigma_{p}^{2}\mathbf{K}_{\prod})^{-1},
\qquad
\bigl(\sigma_{s}^{2}\mathbf{K}_{\sum} + \sigma_{0}^{2}\mathbf{I}\bigr)^{-1}
\approx (\sigma_{s}^{2}\mathbf{K}_{\sum})^{-1}.
\]
Applying these approximation to Equations \ref{eq:posterior-var-prod} and \ref{eq:posterior-var-sum} we get:
\[
\mathrm{Var}_{\sum}(f_{*}) \le \mathrm{Var}_{\prod}(f_{*})
\quad\Longleftrightarrow\quad
\underbrace{
  \sigma_{s}^{2}\mathbf{k}_{*,\sum}^{\top}
  (\mathbf{K}_{\sum})^{-1}
  \mathbf{k}_{*,\sum}
  \;-\;
  \sigma_{p}^{2}\mathbf{k}_{*,\prod}^{\top}
  (\mathbf{K}_{\prod})^{-1}
  \mathbf{k}_{*,\prod}
}_{\text{variance reduction difference}}
\;\ge\;
\underbrace{k_{**}^{(\sum)}D\sigma_{s}^{2} -k_{**}^{(\prod)}\sigma_{p}^{2}}_{\text{prior variance difference}}.
\]

Thus, even with separate scale hyperparameters, the sum kernel yields a smaller posterior variance at $\mathbf{x}_{*}$ if and only if its (scaled) reduction
term exceeds its larger prior variance by at least $k_{**}^{(\sum)} D\sigma_{s}^{2} - k_{**}^{(\prod)} \sigma_{p}^{2}$. Supported by numerical experiments, in the next paragraph we show how structural differences in the two posterior formulations cause this inequality to hold in practice.

\begin{figure}[H]
    \centering
    \includegraphics[width=1\linewidth]{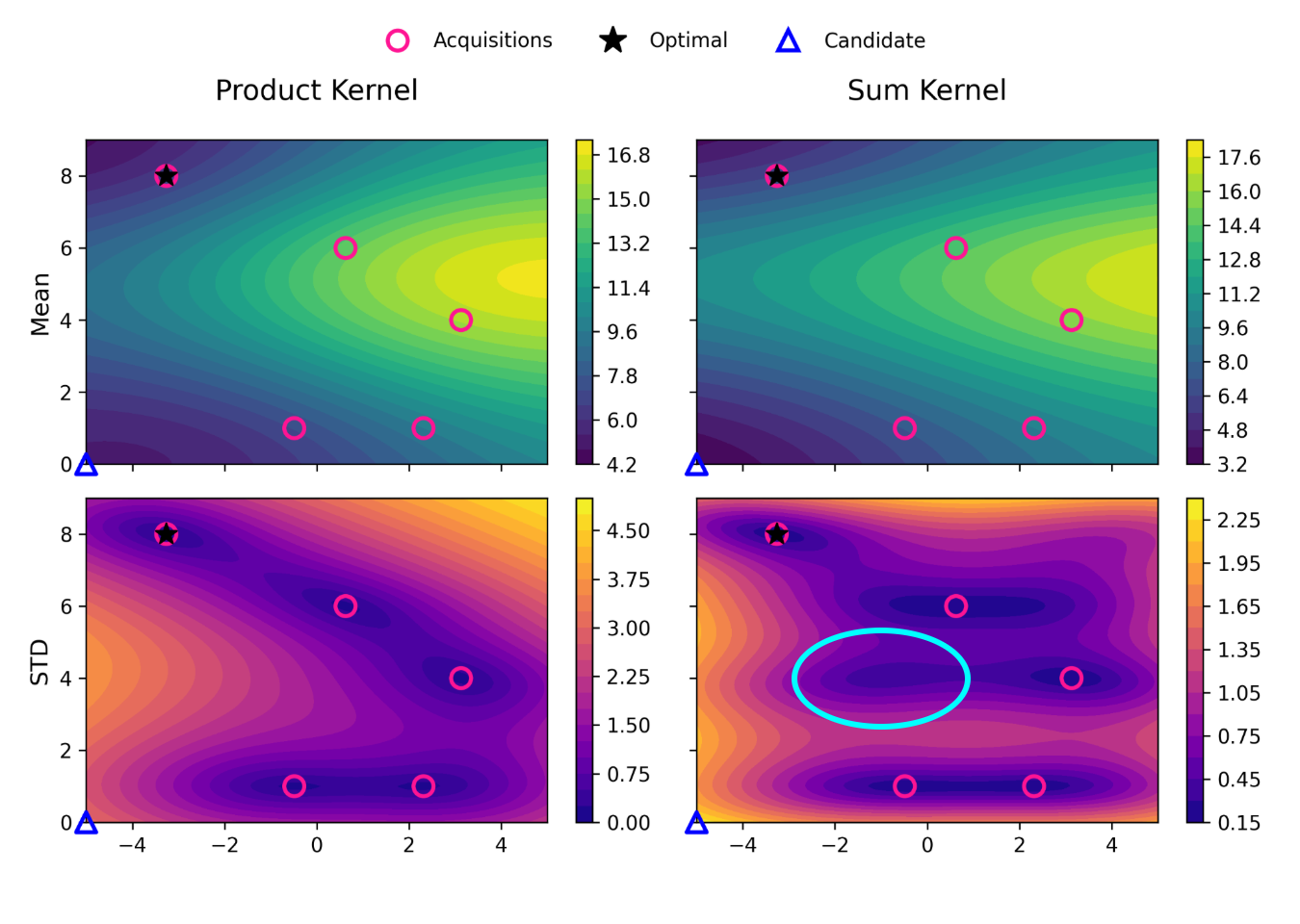}
    \caption{The difference in posterior mean prediction of a product formulation kernel GP model (left) and sum kernel GP model (right) where STD is the standard deviation. Both models were fitted using the same Mat\'ern-5/2 kernel and sobol datapoints sampled from the BS function. Besides the sum kernel having lower overall uncertainty in the design space, the blue circle shows a low uncertainty region within the sum plot that is not present in the product formulation showcasing the sum kernels confidence and stronger beliefs about the objective landscape}
    \label{fig:2D_cont_int_sumvprod}
\end{figure}

\begin{figure}[H]
    \centering
    \includegraphics[width=1\linewidth]{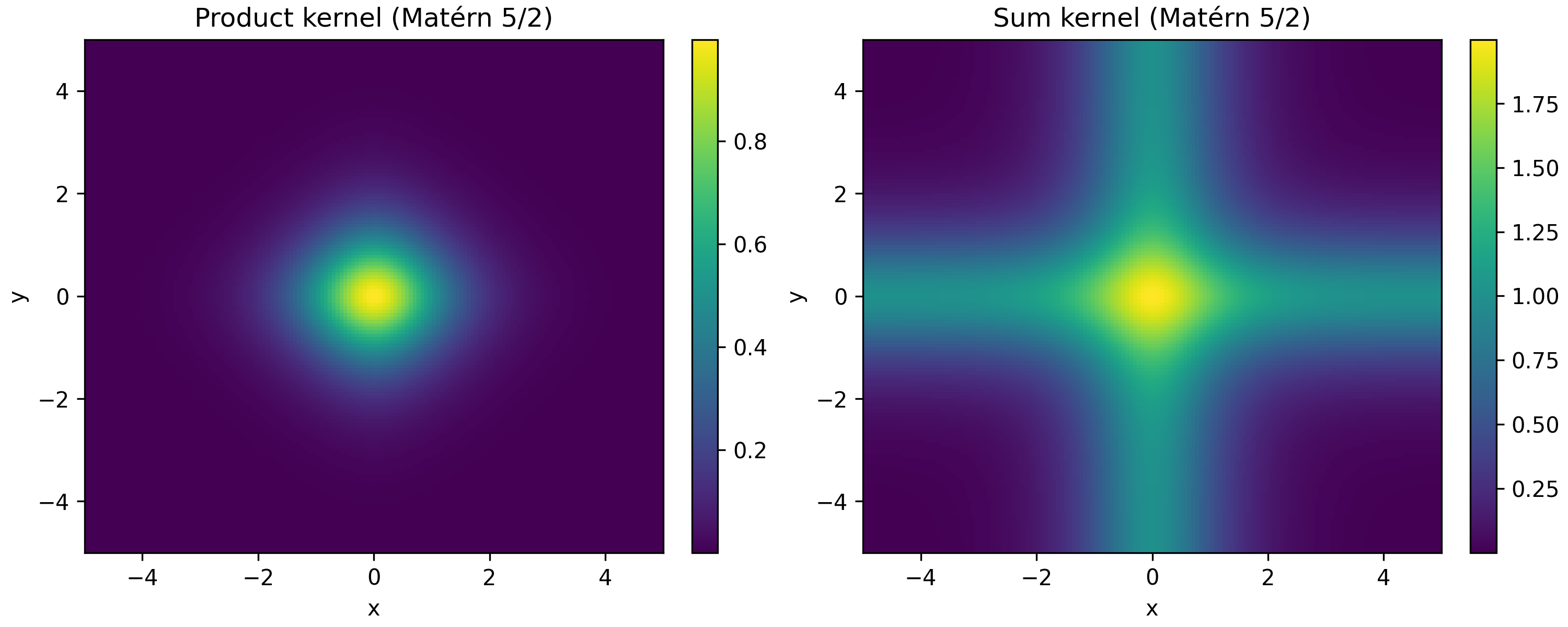}
    \caption{Two dimensional kernel correlation maps for the product and sum kernel formulations, each constructed from one dimensional Mat\'ern 5/2 components with unit prior variance ($\sigma_{s}^{2}$ \& $\sigma_{p}^{2}$ = 1 ) and lengthscale equal to one. The values represent the kernel $K((x,y),(x_{0},y_{0}))$ evaluated over the two dimensional input space with a fixed reference point $(x_{0},y_{0})=(0,0)$. In the product kernel, correlations decay rapidly away from the reference point, forming a localized blob centered at the origin. In contrast, the sum kernel exhibits a broader cross shaped correlation structure, remaining large along the coordinate axes, indicating that correlations persist when either coordinate is similar to the reference point.}

    \label{fig:prod_vs_sum_kernel}
\end{figure}

\paragraph{Interpretation.}
With separate scale hyperparameters, the sum kernel has with prior variance 
of $D\sigma_{s}^{2}$ and induces broad, additive correlations across 
dimensions, whereas the product kernel has with prior variance of 
$\sigma_{p}^{2}$ and produces much more localized correlations that require simultaneous similarity in all coordinates. This contrast is clearly visible in the kernel correlation maps shown in Figure~\ref{fig:prod_vs_sum_kernel}. As a consequence, the broad 
correlations inherent to the sum kernel can lead to very different posterior uncertainty landscapes to the product formulation, as illustrated in 
Figure~\ref{fig:2D_cont_int_sumvprod}.

\begin{figure}[H]
    \centering
    \includegraphics[width=1\linewidth]{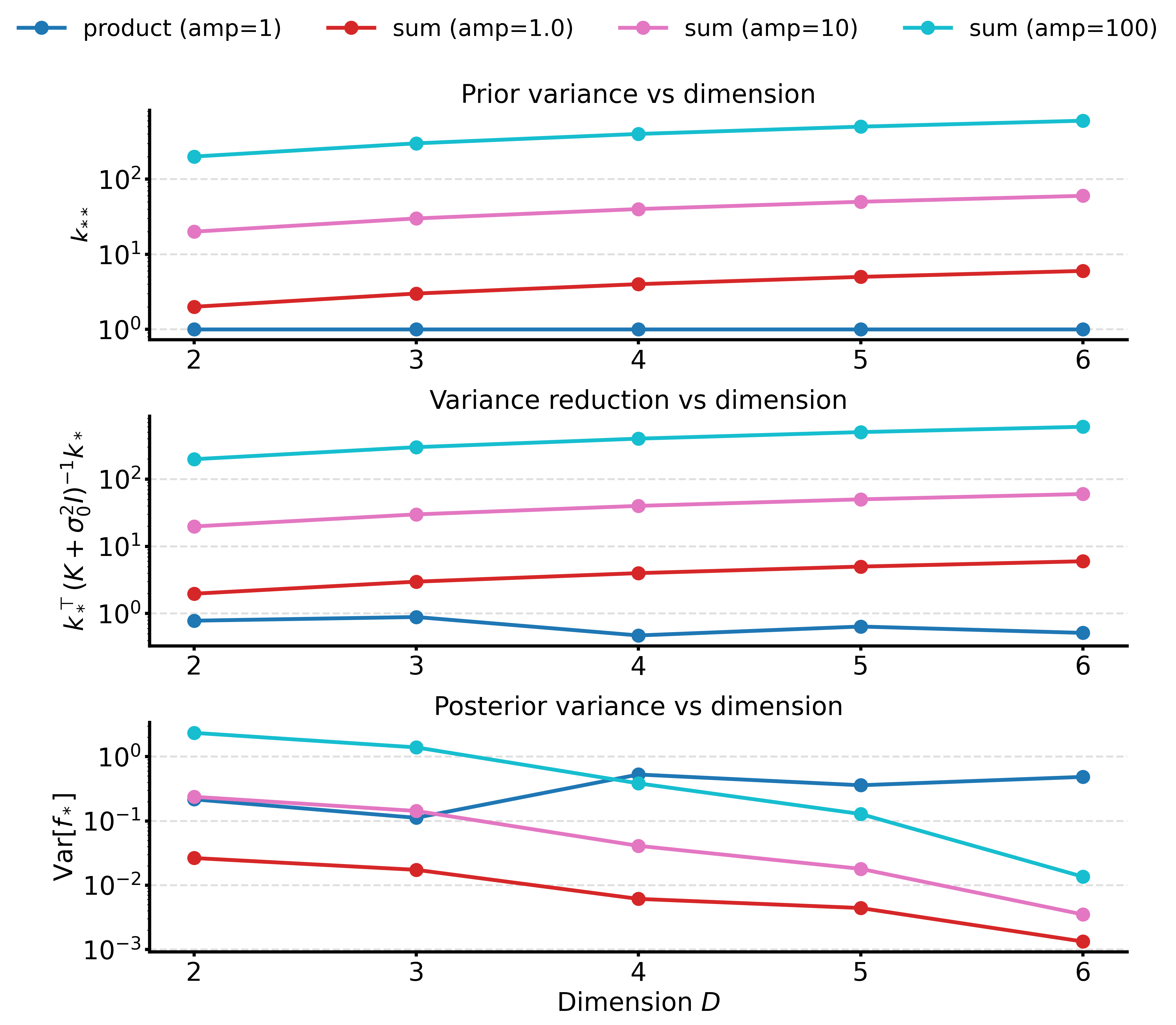}
    \caption{Comparison of prior variance $k_{**}$, variance reduction 
    $k_*^{\top}(K+\sigma_0^2 I)^{-1}k_*$, and posterior variance for product and sum kernels across increasing dimension $D$. Training inputs are sampled uniformly in $[0,1]^D$ with $n_{\text{train}} = 10 \cdot 2^{D-2}$ points (10, 20, 40, 80, 160 for $D=2\ldots6$) with a test point at the zero-origin. The red, pink and cyan plots show the sum kernel formulation with a prior variance values of 1, 10, 100, respectively.}
    \label{fig:sum_prod_decomp}
\end{figure}

\begin{figure}[H]
    \centering
    \includegraphics[width=1\linewidth]{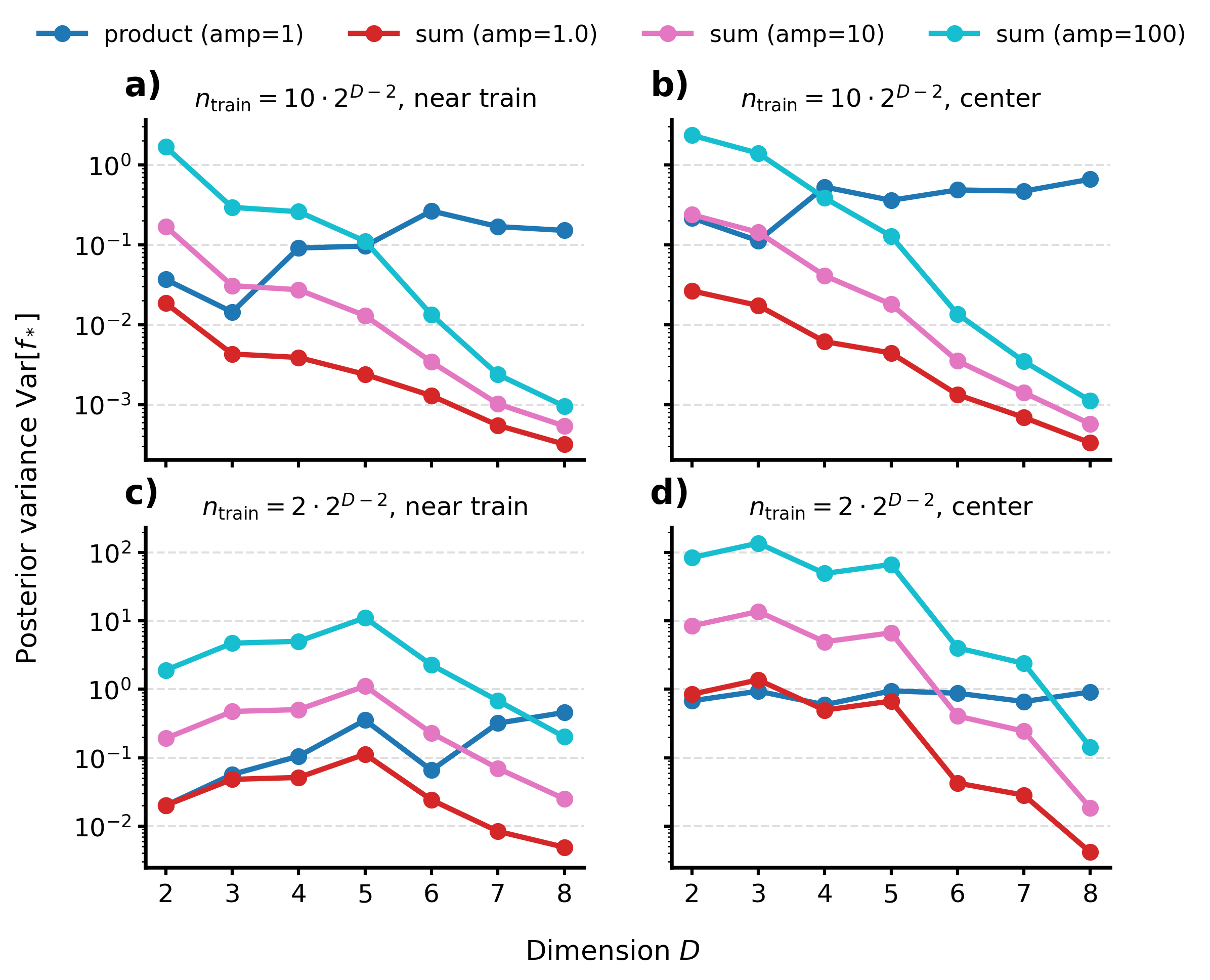}
    \caption{Posterior variances for the sum and product kernel formulations across eight dimensions. The training data consist of $n_{\mathrm{train}} = 10 \cdot 2^{D-2}$ uniformly sampled points in the domain $[0,1]^D$, where $D$ denotes the dimensionality. For the sum kernel, prior variance hyperparameters of $1$, $10$, and $100$ are shown in red, pink, and cyan, respectively. The labels ``near train'' and ``center'' indicate whether the posterior variance was evaluated at a test point located near an existing training sample or at thecenter of the domain.}
    \label{fig:4conditions_sum_prod_posterior_var}
\end{figure}

To examine how these structural differences influence posterior variance, we conducted a set of numerical experiments decomposing the posterior variance into its prior and reduction terms, with results shown in 
Figure~\ref{fig:sum_prod_decomp}. As expected, the product kernel maintains 
a constant prior variance, whereas the sum kernel’s prior variance grows linearly with dimension. For the sampling densities considered, the product kernel’s variance reduction term remains relatively stable across dimensions, while the sum kernel’s reduction term also increases with dimension but at a larger rate. Taken together, this leads to a decreasing posterior variance for the sum kernel as dimension increases. This trend is further demonstrated in 
Figure~\ref{fig:4conditions_sum_prod_posterior_var}, which shows that the 
posterior variance of the sum kernel exceeds that of the product kernel 
only when $\sigma_{s}^{2}$ is significantly larger than $\sigma_{p}^{2}$ and/or when the data are sparse—such as with low training-point counts where much of the space remains uncorrelated. In both sparse and dense sampling regimes, the sum kernel’s posterior variance decreases with dimension, reflecting its accumulation of contributions from all $D$ dimensions, whereas the product kernel becomes increasingly restrictive. It is important to note that the sampled training points are identical for both
kernel formulations, and that the trends described above persist across all random seeds used to generate the training data.

\begin{figure}[H]
    \centering
    \includegraphics[width=0.5\linewidth]{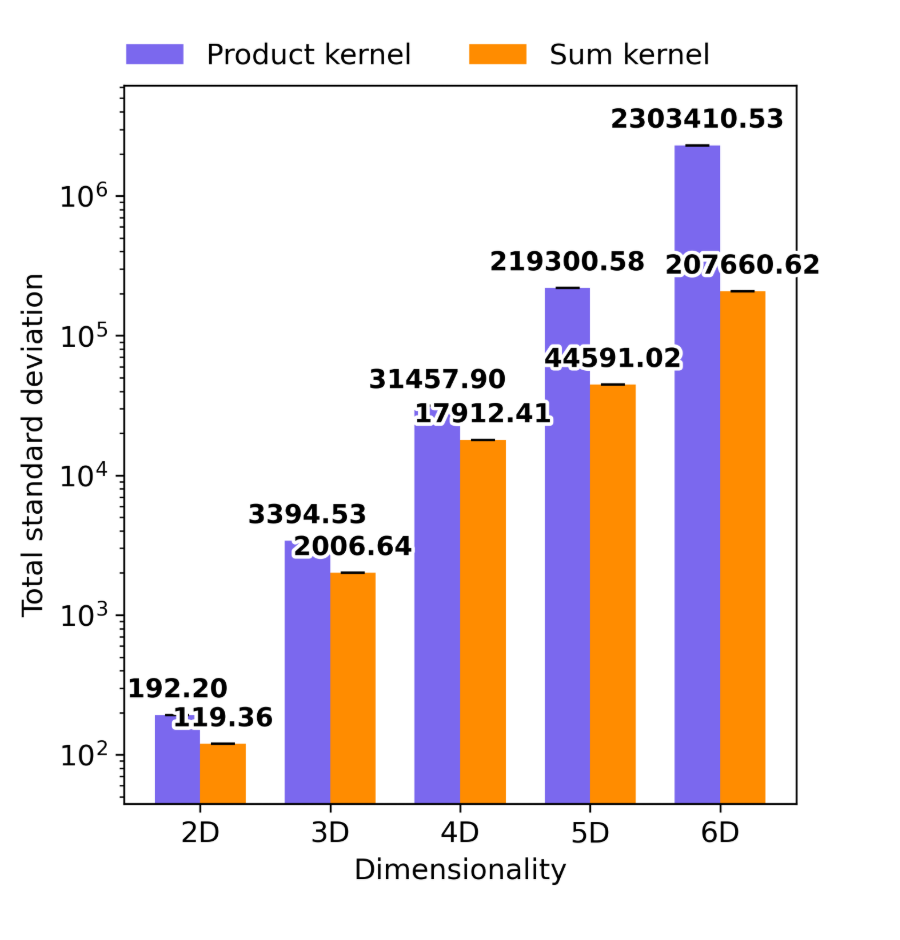}
    \caption{The total summed standard deviation of both the sum and product Mat\'ern-5/2 GP model predictions on fully integer cases of the BS function landscape across dimensionalities. The number of fitted sobol sampled points are [5,10,20,40,60] for each dimension respectively. Both model were fitted on the same exact points across each dimension.}
    \label{fig:SI:std_dim_sum}
\end{figure}

To further quantify the magnitude of the posterior-variance differences 
between the two kernels, we performed an additional numerical experiment in 
which both models were fitted to the same sets of training points $(X, Y)$ 
for a fully integer version of the \textit{BS} test problem across $2$–$6$ dimensions. The resulting total predictive uncertainties, shown in 
Figure~\ref{fig:SI:std_dim_sum}, demonstrate that the sum kernel yields a consistently lower total posterior variance across the deisng space in general when compares to the product kernel when trained on identical data, and that this gap widens with dimensionality. The fitted hyperparameters are listed in 
Table~\ref{tab:SI:sum_test_hyper_table}where it can be seen that the 
sum-kernel prior variance $\sigma_{s}^{2}$ is not fitted to substantially larger values than the product-kernel prior variance $\sigma_{p}^{2}$. These trends therefore are in-line with expectations given our inital numerical tests.

\begin{table*}[ht]
\centering
\scriptsize
\begin{tabular}{c|c|c|c}
\hline
\textbf{Dim} & \textbf{Kernel} & \textbf{Variance / Noise} & \textbf{Lengthscales} \\
\hline

\multirow{2}{*}{2D} 
& Product (Mat52) 
& var = 2.1182, noise = 0.001 
& $\ell = [1.0189,\ 1.1745]$ \\

& Sum (Mat52) 
& var = 1.9897, noise = 0.001 
& $\ell = [1.1065,\ 0.9621]$ \\
\hline

\multirow{2}{*}{3D} 
& Product (Mat52) 
& var = 2.2211, noise = 0.001 
& $\ell = [1.5205,\ 0.9352,\ 0.3170]$ \\

& Sum (Mat52) 
& var = 2.5875, noise = 0.001 
& $\ell = [0.9601,\ 0.7587,\ 0.3574]$ \\
\hline

\multirow{2}{*}{4D} 
& Product (Mat52) 
& var = 2.6357, noise = 0.001 
& $\ell = [0.4082,\ 2.0836,\ 1.2681,\ 0.4527]$ \\

& Sum (Mat52) 
& var = 0.3589, noise = 0.001 
& $\ell = [0.1677,\ 0.0795,\ 0.4745,\ 0.5131]$ \\
\hline

\multirow{2}{*}{5D} 
& Product (Mat52) 
& var = 1.2384, noise = 0.001 
& $\ell = [0.7439,\ 0.2428,\ 0.6447,\ 1.9996,\ 0.8431]$ \\

& Sum (Mat52) 
& var = 3.0141, noise = 0.001 
& $\ell = [0.4724,\ 0.5097,\ 0.5054,\ 0.5087,\ 0.5001]$ \\
\hline

\multirow{2}{*}{6D} 
& Product (Mat52) 
& var = 2.5295, noise = 0.001 
& $\ell = [1.0625,\ 0.5349,\ 1.4238,\ 0.4891,\ 0.8973,\ 0.7935]$ \\

& Sum (Mat52) 
& var = 2.9789, noise = 0.001 
& $\ell = [0.5290,\ 0.5398,\ 0.5240,\ 0.5228,\ 0.5286,\ 0.5323]$ \\
\hline

\end{tabular}
\caption{Hyperparameters of BOSS Mat\'ern-5/2 product and sum kernels for the 2D--6D Yuh–Styblinski–Tang test problems corresponding to Figure \ref{fig:SI:std_dim_sum}.}
\label{tab:SI:sum_test_hyper_table}
\end{table*}

\subsection{Repeated Sampling Complementary Results}

\begin{figure}[H]
    \centering
    \includegraphics[width=1\linewidth]{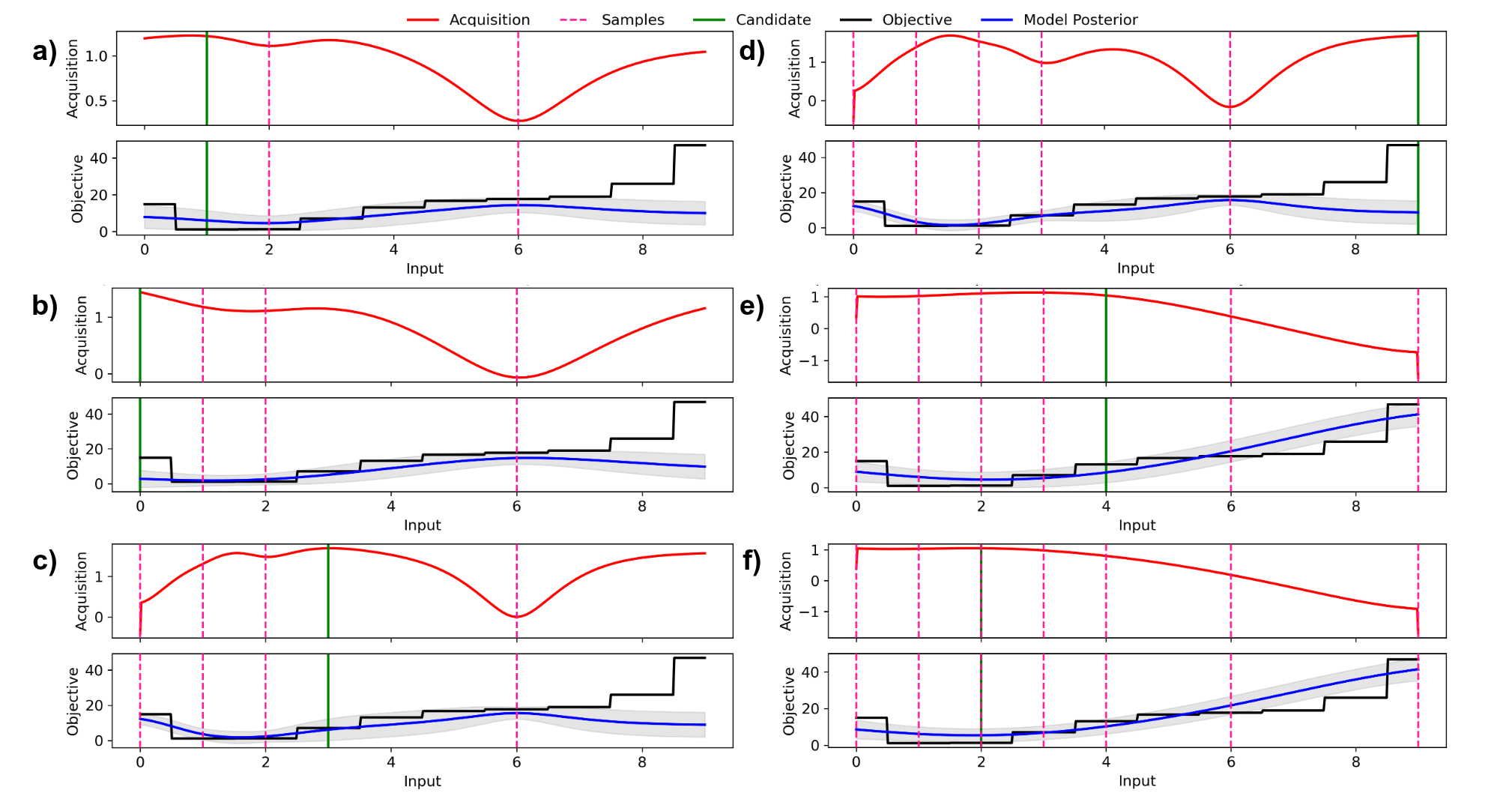}
    \caption{Iterations 1-6 marked by a-f showing subtract method leading to repeated acquisitions in LCB}
    \label{SI_fig:0.2_LCB_subtract_repeat}
\end{figure}

\subsection{Discrete Temperature Parameter (\texorpdfstring{$\tau$}{tau}) Investigation}%
\label{SI:subsection:tau_invesigation}

\begin{figure}[H]
    \centering
    \includegraphics[width=1\linewidth]{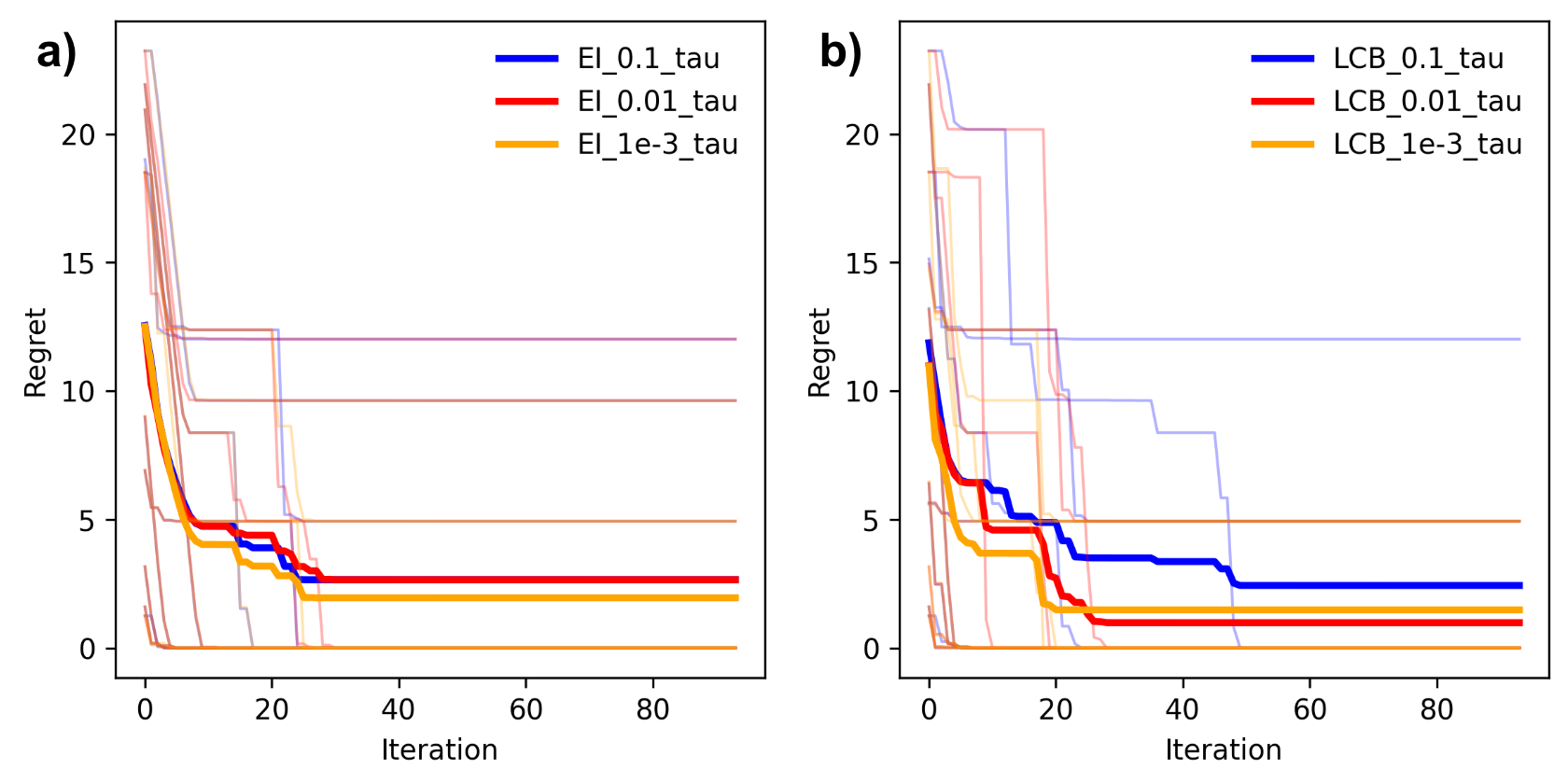}
    \caption{Empirical convergence of the \texttt{BOSS\_on\_gam\_Mat52} model on the DUST1 benchmark function using the acquisition functions: (a) Expected Improvement (EI) and (b) Lower Confidence Bound (LCB).}
    \label{fig:SI:tau_investigation}
\end{figure}

\begin{figure}[H]
    \centering
    \includegraphics[width=1\linewidth]{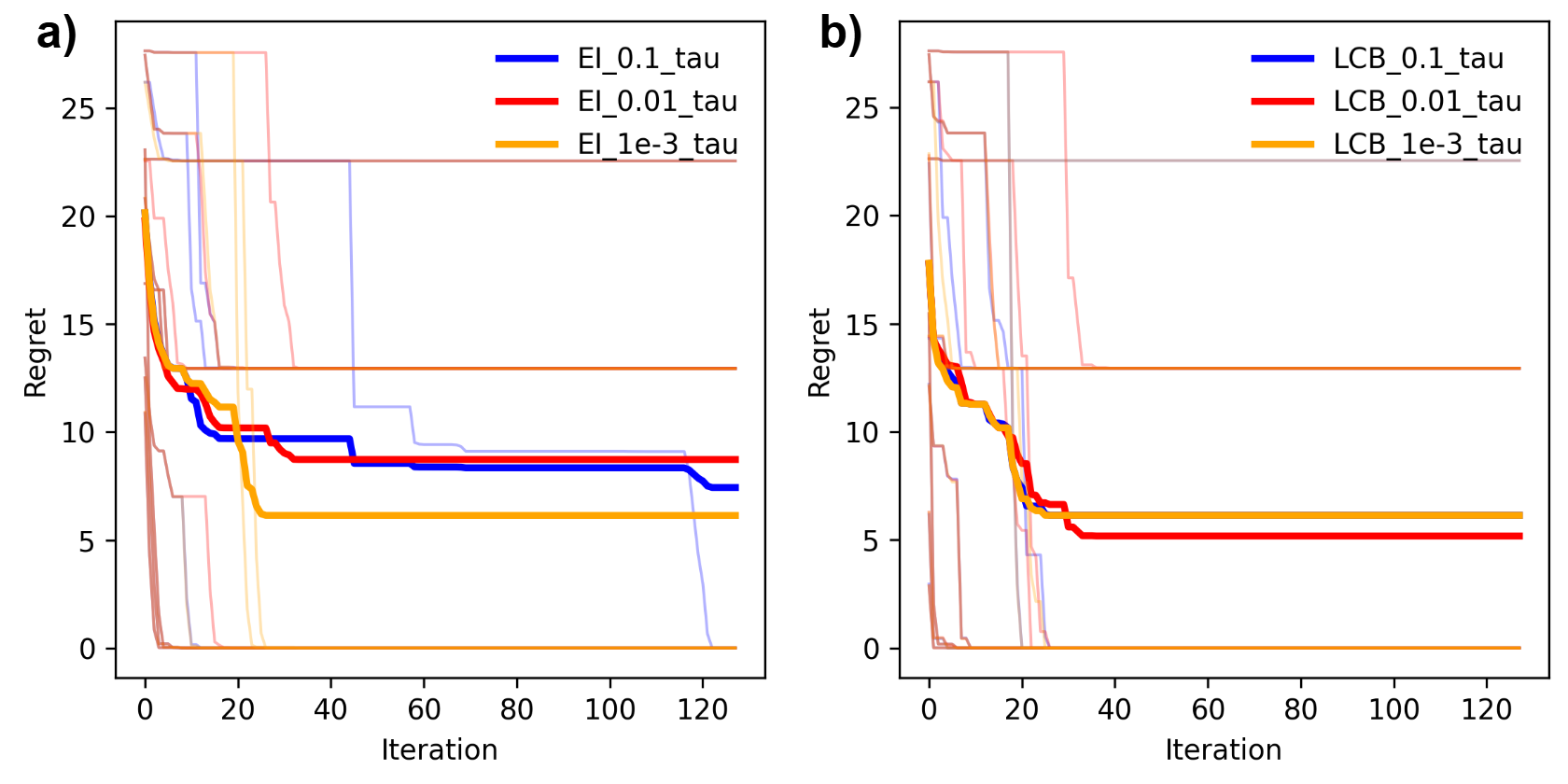}
    \caption{Empirical convergence of the \texttt{BOSS\_on\_gam\_Mat52} model on the DUST2 benchmark function using the acquisition functions: (a) Expected Improvement (EI) and (b) Lower Confidence Bound (LCB).}
    \label{fig:SI:tau_investigation_Dust2}
\end{figure}

\subsection{Butternut Squash Benchmark Complementary Results}\label{SI:subsubection_BS_complementary_results}
\subsubsection{Loose Tolerance}

\begin{figure}[H]
    \centering
    \includegraphics[width=1\linewidth]{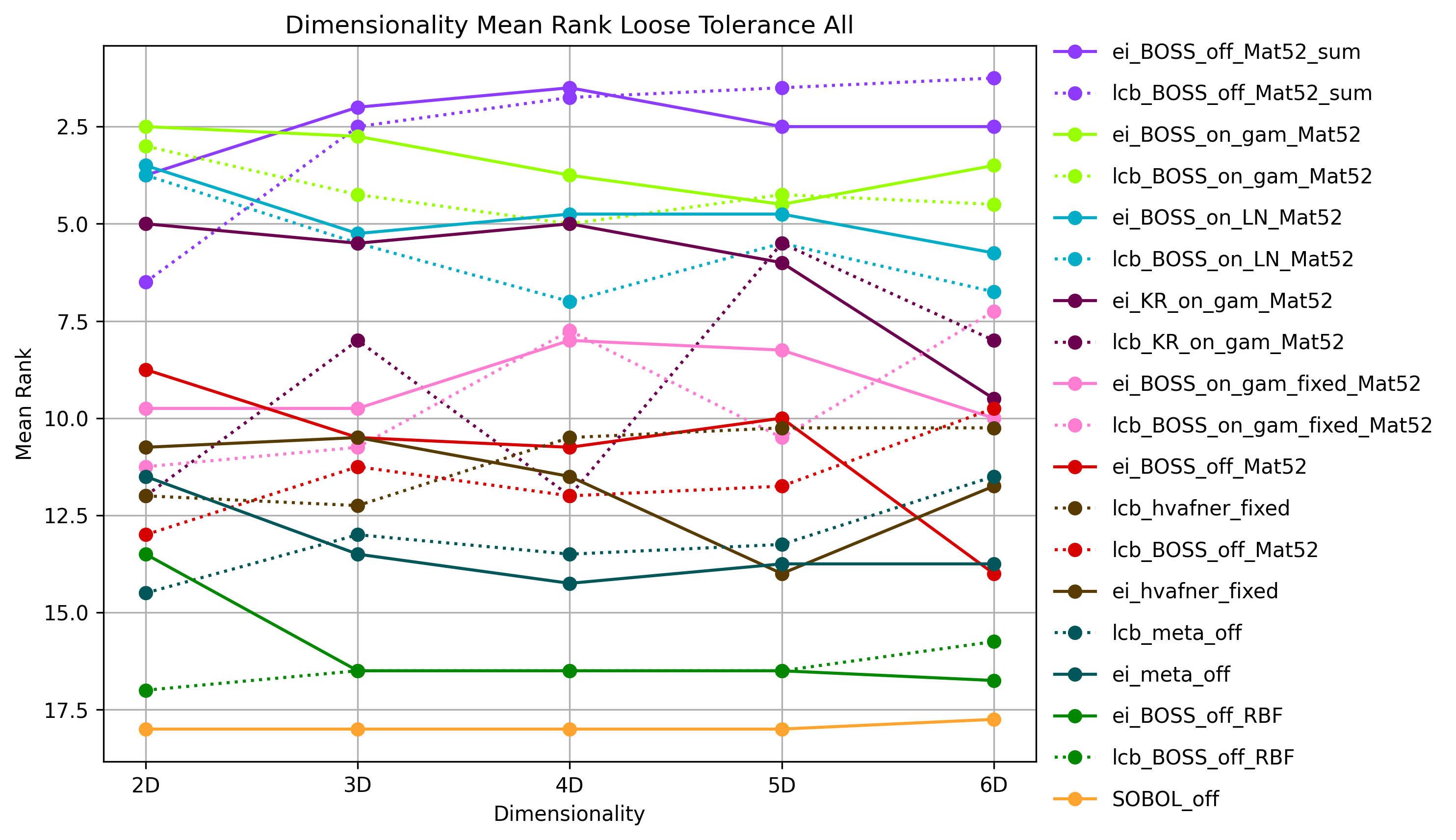}
    \caption{Plot showing the mean composite score ranks of all models across each dimension of the Butternut Squash benchmark function variants with continuous+integer domains under the loose tolerance}
    \label{fig:SI:Dimension_all_los}
\end{figure}

\begin{figure}[H]
    \centering
    \includegraphics[width=1\linewidth]{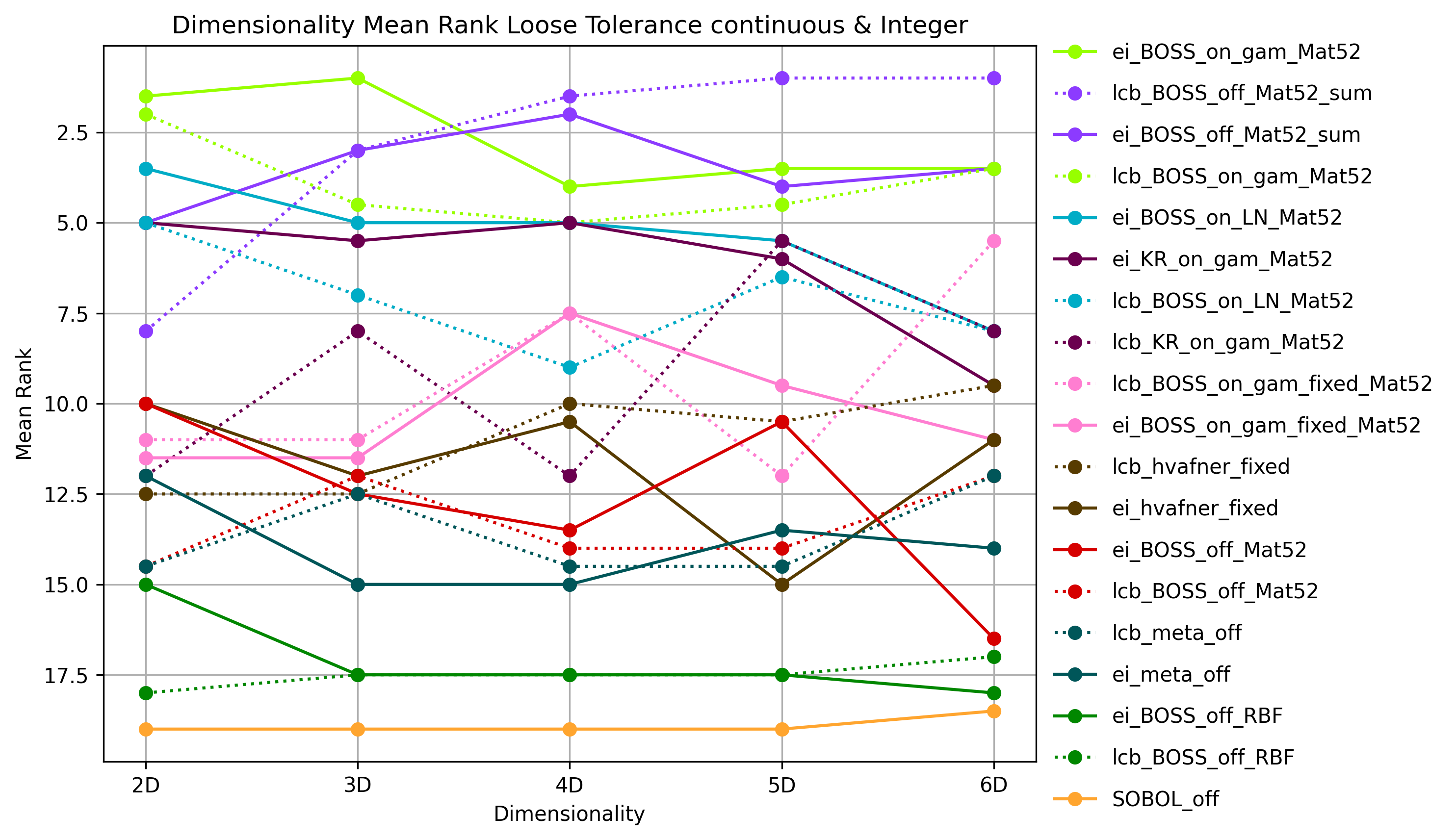}
    \caption{Plot showing the mean composite score ranks of all models across each dimension of the Butternut Squash benchmark function variants with continuous+integer domains under the loose tolerance.}
    \label{fig:SI:Dimension_cont_int_los}
\end{figure}

\begin{figure}[H]
    \centering
    \includegraphics[width=1\linewidth]{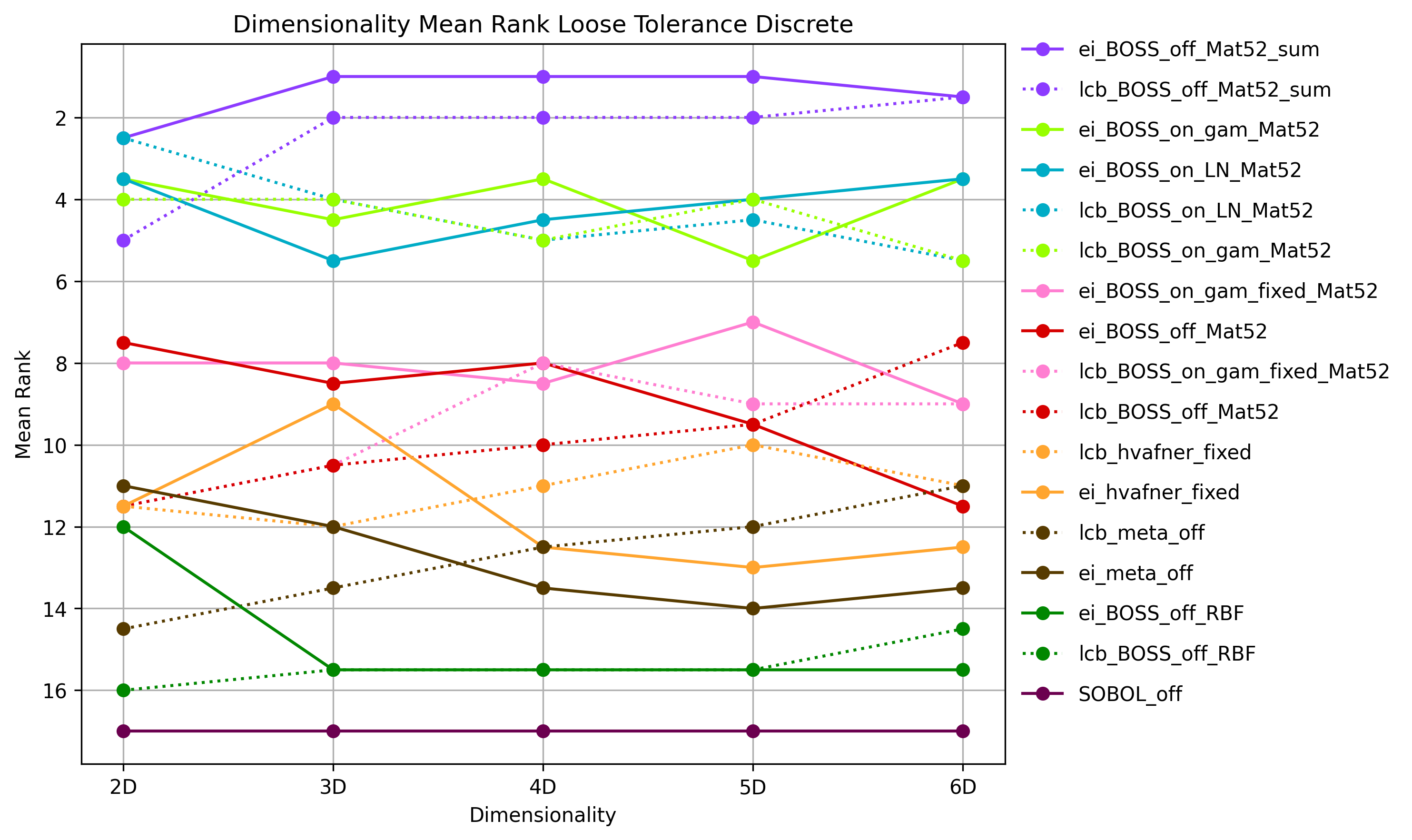}
    \caption{Plot showing the mean composite score ranks of all models across each dimension of the Butternut Squash benchmark function variants with discrete domains under loose tolerance.}
    \label{fig:SI:Dimension_discrete_los}
\end{figure}

In the tables below, R is abbreviated as Rank.

\begin{table}[H]
\centering
\begin{tabular}{lrrrrr}
\toprule
\textbf{Model Settings} & \textbf{Num Ranks} & \textbf{Mean R.} & \textbf{Median R.} & \textbf{Min R.} & \textbf{Max R.} \\
\midrule
ei\_BOSS\_off\_Mat52\_sum & 20 & 2.35 & 2.00 & 1 & 6 \\
lcb\_BOSS\_off\_Mat52\_sum & 20 & 2.85 & 2.00 & 1 & 9 \\
ei\_BOSS\_on\_gam\_Mat52 & 20 & 3.40 & 3.50 & 1 & 6 \\
lcb\_BOSS\_on\_gam\_Mat52 & 20 & 4.25 & 4.00 & 1 & 7 \\
ei\_BOSS\_on\_LN\_Mat52 & 20 & 5.00 & 5.00 & 1 & 10 \\
lcb\_BOSS\_on\_LN\_Mat52 & 20 & 5.55 & 5.00 & 1 & 12 \\
ei\_KR\_on\_gam\_Mat52 & 10 & 6.90 & 7.00 & 2 & 12 \\
lcb\_KR\_on\_gam\_Mat52 & 10 & 8.90 & 9.50 & 2 & 16 \\
ei\_BOSS\_on\_gam\_fixed\_Mat52 & 20 & 8.95 & 8.00 & 7 & 16 \\
lcb\_BOSS\_on\_gam\_fixed\_Mat52 & 20 & 9.85 & 10.00 & 5 & 14 \\
lcb\_hvafner\_fixed & 20 & 10.90 & 11.00 & 8 & 15 \\
ei\_BOSS\_off\_Mat52 & 20 & 10.50 & 10.50 & 7 & 17 \\
lcb\_BOSS\_off\_Mat52 & 20 & 11.20 & 11.00 & 7 & 16 \\
ei\_hvafner\_fixed & 20 & 11.85 & 12.00 & 6 & 16 \\
lcb\_meta\_off & 20 & 13.25 & 13.00 & 10 & 17 \\
ei\_meta\_off & 20 & 13.65 & 13.50 & 10 & 16 \\
ei\_BOSS\_off\_RBF & 20 & 16.15 & 16.00 & 11 & 19 \\
lcb\_BOSS\_off\_RBF & 20 & 16.25 & 16.00 & 14 & 18 \\
SOBOL\_off & 20 & 17.50 & 17.50 & 17 & 19 \\
\bottomrule
\end{tabular}
\caption{Summary of model composite score ranking statistics across all 20 variants of the BS function under the loose tolerance. Note that since the \textbf{KR} model only applies to the continuous+integer domains, it has only 10 computed ranks corresponding to the 10 integer+continuous domain variants of the BS functions.}
\label{tab:SI:all_loose}
\end{table}

\begin{table}[H]
\centering
\begin{tabular}{lrrrrr}
\toprule
\textbf{Model Settings} & \textbf{Num Ranks} & \textbf{Mean R.} & \textbf{Median R.} & \textbf{Min R.} & \textbf{Max R.} \\
\midrule
ei\_BOSS\_on\_gam\_Mat52 & 10 & 2.70 & 3.00 & 1 & 4 \\
lcb\_BOSS\_off\_Mat52\_sum & 10 & 2.90 & 1.50 & 1 & 9 \\
ei\_BOSS\_off\_Mat52\_sum & 10 & 3.50 & 3.00 & 1 & 6 \\
lcb\_BOSS\_on\_gam\_Mat52 & 10 & 3.90 & 3.50 & 1 & 7 \\
ei\_BOSS\_on\_LN\_Mat52 & 10 & 5.40 & 5.00 & 3 & 10 \\
ei\_KR\_on\_gam\_Mat52 & 10 & 6.20 & 7.00 & 2 & 12 \\
lcb\_BOSS\_on\_LN\_Mat52 & 10 & 7.10 & 6.50 & 5 & 12 \\
lcb\_KR\_on\_gam\_Mat52 & 10 & 9.10 & 9.50 & 2 & 16 \\
lcb\_BOSS\_on\_gam\_fixed\_Mat52 & 10 & 9.40 & 10.00 & 5 & 14 \\
ei\_BOSS\_on\_gam\_fixed\_Mat52 & 10 & 10.20 & 8.50 & 7 & 16 \\
lcb\_hvafner\_fixed & 10 & 11.00 & 11.00 & 8 & 15 \\
ei\_hvafner\_fixed & 10 & 11.70 & 11.50 & 6 & 16 \\
ei\_BOSS\_off\_Mat52 & 10 & 12.60 & 12.50 & 9 & 17 \\
lcb\_BOSS\_off\_Mat52 & 10 & 13.30 & 14.50 & 9 & 16 \\
lcb\_meta\_off & 10 & 13.60 & 13.00 & 10 & 17 \\
ei\_meta\_off & 10 & 13.90 & 14.50 & 11 & 16 \\
ei\_BOSS\_off\_RBF & 10 & 17.10 & 17.00 & 13 & 19 \\
lcb\_BOSS\_off\_RBF & 10 & 17.50 & 18.00 & 16 & 18 \\
SOBOL\_off & 10 & 18.90 & 19.00 & 18 & 19 \\
\bottomrule
\end{tabular}
\caption{Summary of all model composite score ranking statistics across only the continuous+integer variable domains (10 variants) of the BS function under the loose tolerance.}
\label{tab:SI:cont_int_loose}
\end{table}

\begin{table}[H]
\centering
\begin{tabular}{lrrrrr}
\toprule
\textbf{Model Settings} & \textbf{Num Ranks} & \textbf{Mean R.} & \textbf{Median R.} & \textbf{Min R.} & \textbf{Max R.} \\
\midrule
ei\_BOSS\_off\_Mat52\_sum & 10 & 1.40 & 1.00 & 1 & 3 \\
lcb\_BOSS\_off\_Mat52\_sum & 10 & 2.50 & 2.00 & 1 & 6 \\
ei\_BOSS\_on\_gam\_Mat52 & 10 & 4.10 & 4.00 & 2 & 6 \\
ei\_BOSS\_on\_LN\_Mat52 & 10 & 4.20 & 4.50 & 1 & 6 \\
lcb\_BOSS\_on\_LN\_Mat52 & 10 & 4.30 & 5.00 & 1 & 6 \\
lcb\_BOSS\_on\_gam\_Mat52 & 10 & 4.50 & 4.00 & 3 & 6 \\
ei\_BOSS\_on\_gam\_fixed\_Mat52 & 10 & 8.10 & 8.00 & 7 & 10 \\
ei\_BOSS\_off\_Mat52 & 10 & 9.00 & 8.00 & 7 & 16 \\
lcb\_BOSS\_on\_gam\_fixed\_Mat52 & 10 & 9.60 & 9.50 & 7 & 14 \\
lcb\_BOSS\_off\_Mat52 & 10 & 9.80 & 10.00 & 7 & 13 \\
lcb\_hvafner\_fixed & 10 & 11.10 & 11.00 & 9 & 14 \\
ei\_hvafner\_fixed & 10 & 11.70 & 12.50 & 8 & 15 \\
lcb\_meta\_off & 10 & 12.70 & 12.50 & 10 & 15 \\
ei\_meta\_off & 10 & 12.80 & 13.00 & 10 & 14 \\
ei\_BOSS\_off\_RBF & 10 & 14.80 & 15.00 & 11 & 16 \\
lcb\_BOSS\_off\_RBF & 10 & 15.40 & 15.50 & 14 & 16 \\
SOBOL\_off & 10 & 17.00 & 17.00 & 17 & 17 \\
\bottomrule
\end{tabular}
\caption{Summary of all model composite score ranking statistics across only the discrete variable domains (10 variants) of the BS function under the \textbf{loose tolerance} level.}
\label{tab:SI:disc_loose}
\end{table}

\subsubsection{Medium Tolerance}

\begin{figure}[H]
    \centering
    \includegraphics[width=1\linewidth]{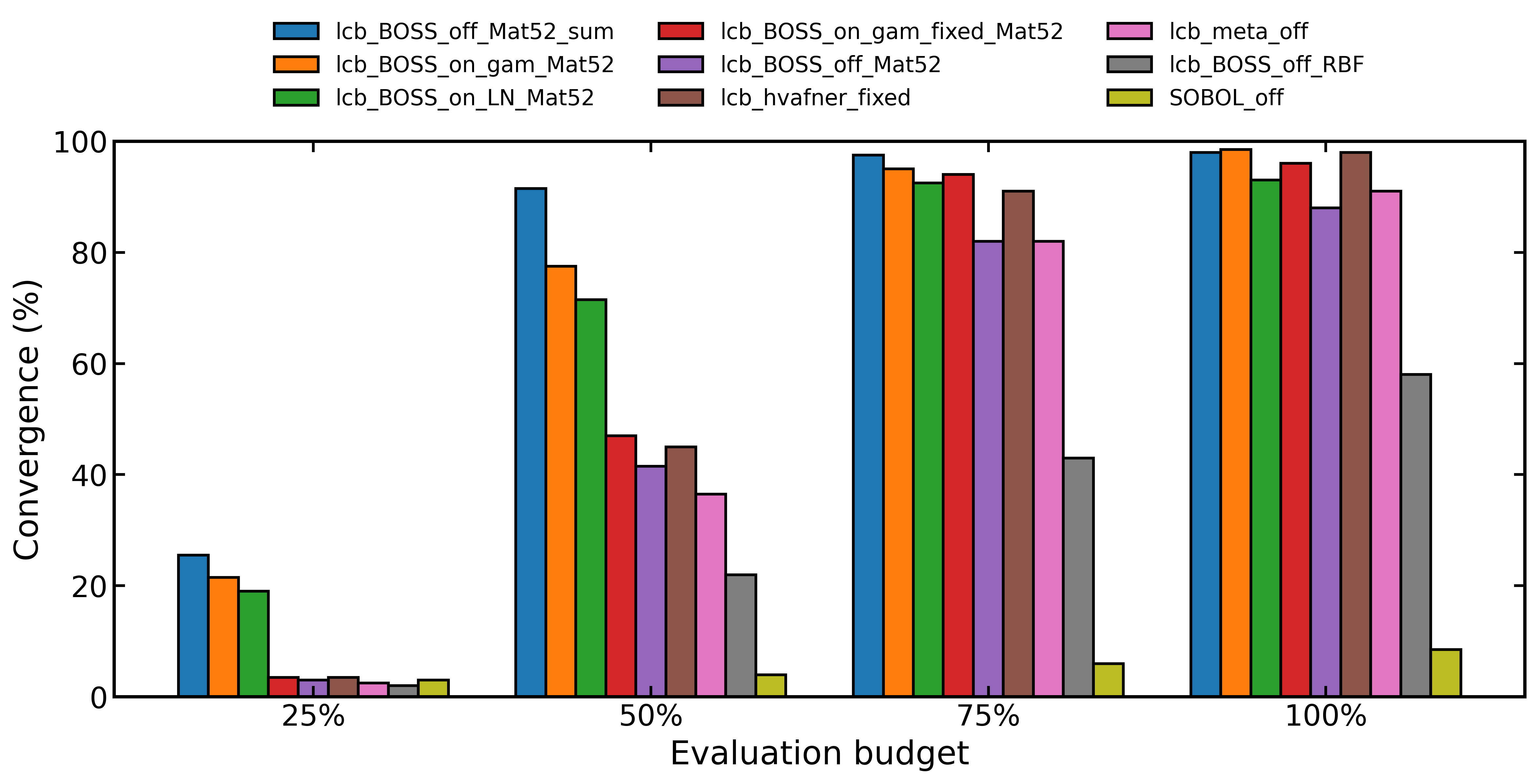}
    \caption{Histogram illustrating absolute convergence performance, measured as the percentage of converged runs aggregated over all problem instances (10 runs per model per instance, each with different Sobol initializations), for all models (excluding KR) using the LCB AF on the \textit{BS} benchmark function. Convergence defined under our medium tolerance criterion counts are pooled across all 18 instances, and evaluated under relative budget constraints of 25\%, 50\%, 75\%, and 100\%, where each percentage corresponds to the respective fraction of the maximum iterations for each individual instance.}
    \label{fig:all_lcb_bins_medium}
\end{figure}

\begin{figure}[H]
    \centering
    \includegraphics[width=1\linewidth]{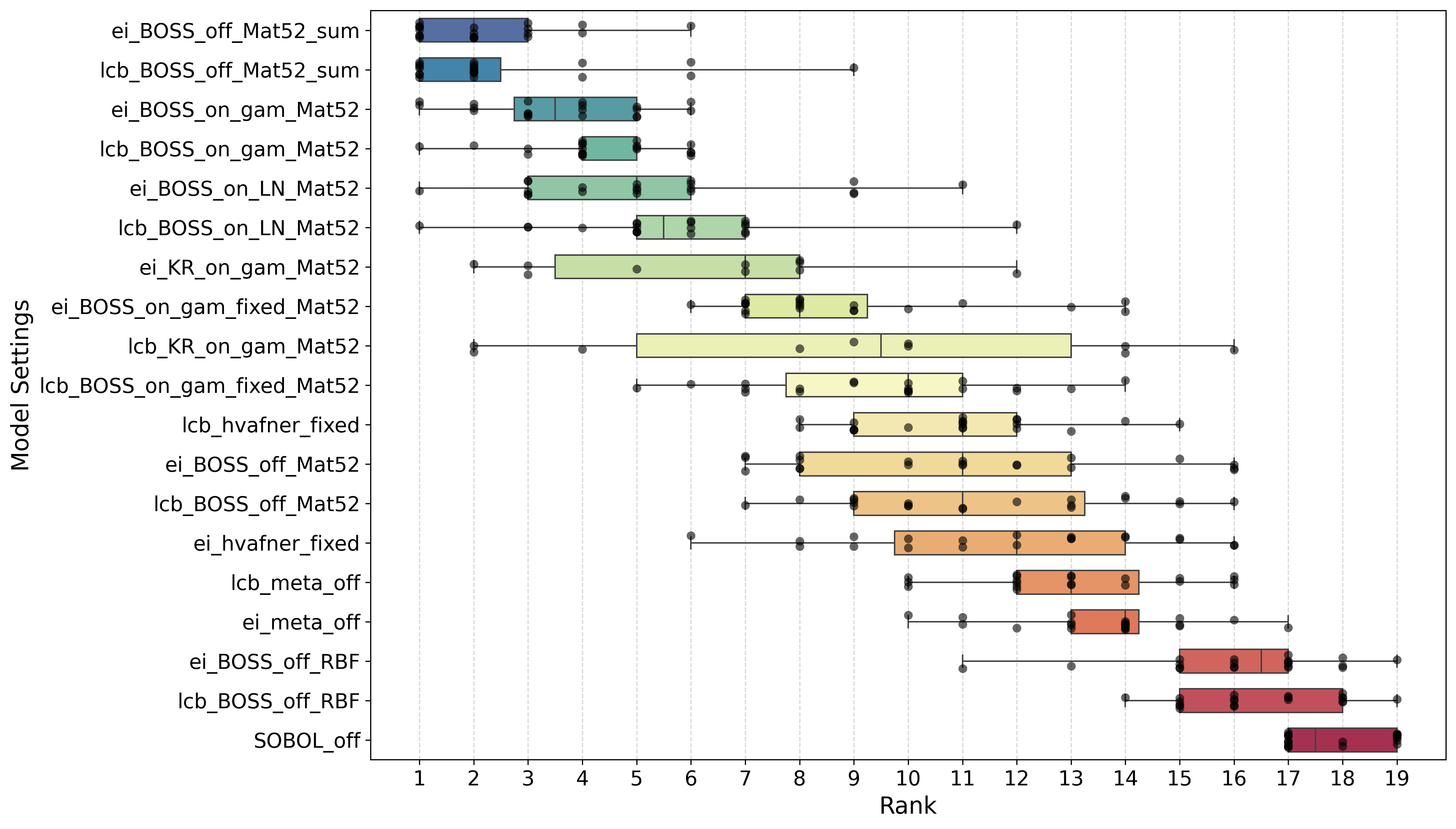}
    \caption{Box plot showing the model rank statistics in terms of composite score (as defined in Eq.\ref{eq:Composite score}) for all 20 variants of the \textit{BS} benchmark function. Note that the \texttt{KR\_on\_gam} models are limited to 10 variants as it only applies to the continuous + integer cases.}
    \label{fig:all_rank_medium}
\end{figure}

\begin{figure}[H]
    \centering
    \includegraphics[width=1\linewidth]{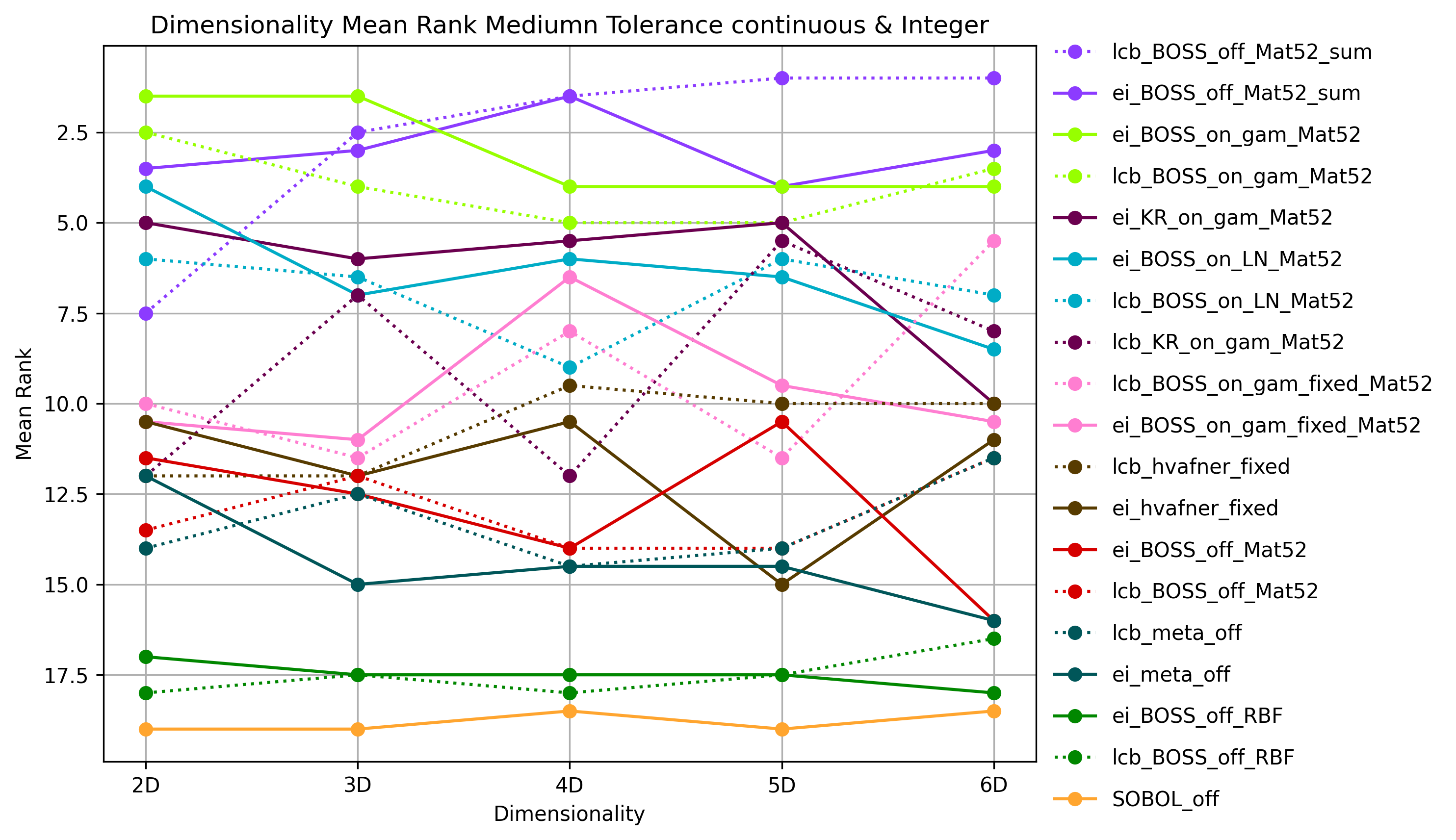}
    \caption{Plot showing the mean composite score ranks of all models across each dimension of the Butternut Squash benchmark function variants with continuous+integer domains under the medium tolerance}
    \label{fig:SI:Dimension_cont_int_med}
\end{figure}

\begin{figure}[H]
    \centering
    \includegraphics[width=1\linewidth]{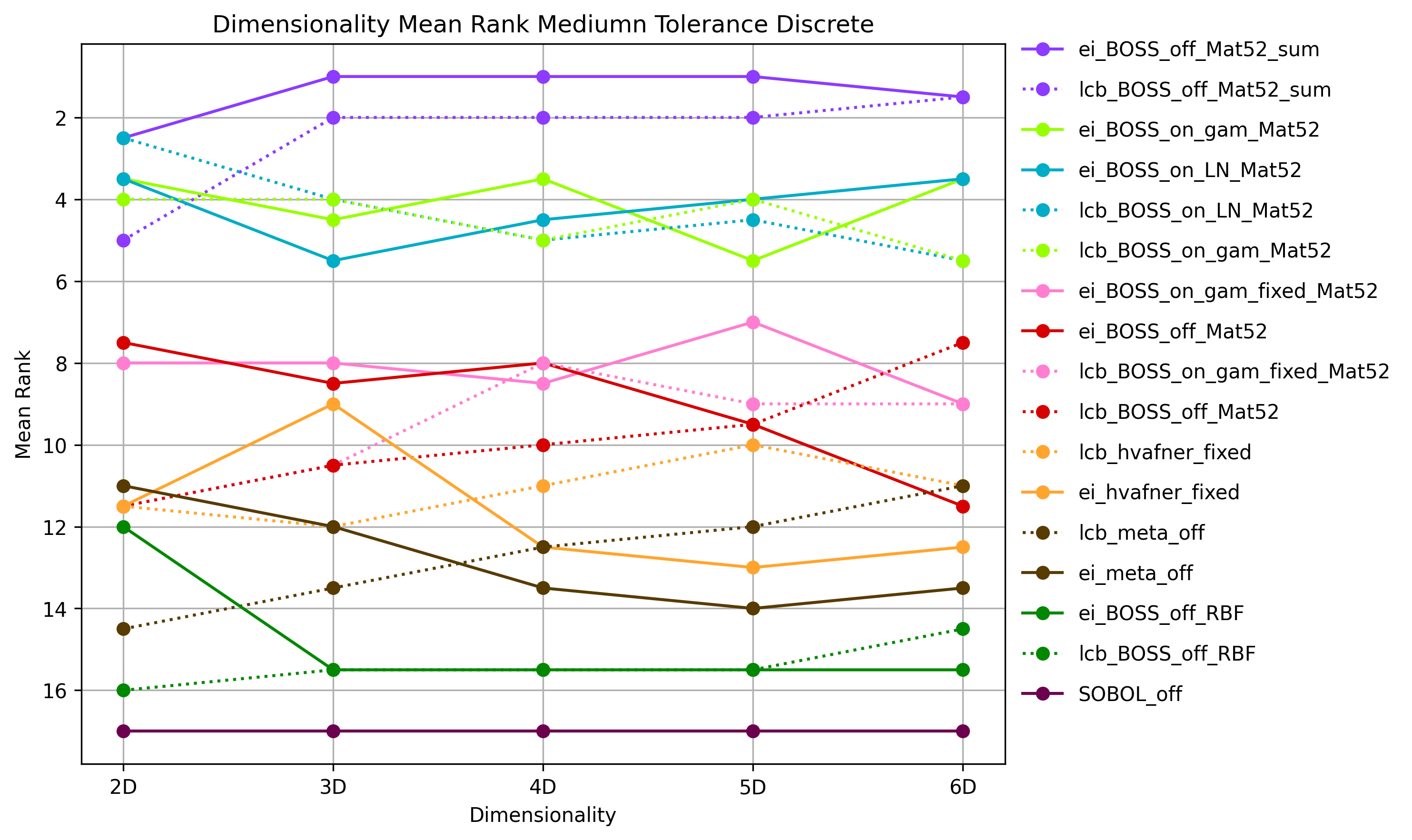}
    \caption{Plot showing the mean composite score ranks of all models across each dimension of the Butternut Squash benchmark function variants with discrete domains under the medium tolerance.}
    \label{fig:SI:Dimension_discrete_med}
\end{figure}

In the tables below, R is abbreviated as Rank.

\begin{table}[H]
\centering
\begin{tabular}{lrrrrr}
\toprule
\textbf{Model Settings} & \textbf{Num Ranks} & \textbf{Mean R.} & \textbf{Median R.} & \textbf{Min R.} & \textbf{Max R.} \\
\midrule
ei\_BOSS\_off\_Mat52\_sum & 20 & 2.20 & 2.00 & 1 & 6 \\
lcb\_BOSS\_off\_Mat52\_sum & 20 & 2.60 & 2.00 & 1 & 9 \\
ei\_BOSS\_on\_gam\_Mat52 & 20 & 3.55 & 3.50 & 1 & 6 \\
lcb\_BOSS\_on\_gam\_Mat52 & 20 & 4.25 & 4.00 & 1 & 6 \\
ei\_BOSS\_on\_LN\_Mat52 & 20 & 5.30 & 5.00 & 1 & 11 \\
lcb\_BOSS\_on\_LN\_Mat52 & 20 & 5.60 & 5.50 & 1 & 12 \\
ei\_KR\_on\_gam\_Mat52 & 10 & 6.30 & 7.00 & 2 & 12 \\
ei\_BOSS\_on\_gam\_fixed\_Mat52 & 20 & 8.85 & 8.00 & 6 & 14 \\
lcb\_KR\_on\_gam\_Mat52 & 10 & 8.90 & 9.50 & 2 & 16 \\
lcb\_BOSS\_on\_gam\_fixed\_Mat52 & 20 & 9.45 & 10.00 & 5 & 14 \\
lcb\_hvafner\_fixed & 20 & 10.90 & 11.00 & 8 & 15 \\
ei\_BOSS\_off\_Mat52 & 20 & 10.95 & 11.00 & 7 & 16 \\
lcb\_BOSS\_off\_Mat52 & 20 & 11.40 & 11.00 & 7 & 16 \\
ei\_hvafner\_fixed & 20 & 11.75 & 12.00 & 6 & 16 \\
lcb\_meta\_off & 20 & 13.00 & 13.00 & 10 & 16 \\
ei\_meta\_off & 20 & 13.60 & 14.00 & 10 & 17 \\
ei\_BOSS\_off\_RBF & 20 & 16.15 & 16.50 & 11 & 19 \\
lcb\_BOSS\_off\_RBF & 20 & 16.45 & 16.00 & 14 & 19 \\
SOBOL\_off & 20 & 17.90 & 17.50 & 17 & 19 \\
\bottomrule
\end{tabular}
\caption{Summary of model composite score ranking statistics across all 20 variants of the BS function under the medium tolerance. Note that since the \textbf{KR} model only applies to the continuous+integer domains, it has only 10 computed ranks corresponding to the 10 integer+continuous domain variants of the BS functions.}
\label{tab:SI:all_medium}
\end{table}

\begin{table}[H]
\centering
\begin{tabular}{lrrrrr}
\toprule
\textbf{Model Settings} & \textbf{Num Ranks} & \textbf{Mean R.} & \textbf{Median R.} & \textbf{Min R.} & \textbf{Max R.} \\
\midrule
lcb\_BOSS\_off\_Mat52\_sum & 10 & 2.70 & 1.00 & 1 & 9 \\
ei\_BOSS\_on\_gam\_Mat52 & 10 & 3.00 & 3.00 & 1 & 5 \\
ei\_BOSS\_off\_Mat52\_sum & 10 & 3.00 & 3.00 & 1 & 6 \\
lcb\_BOSS\_on\_gam\_Mat52 & 10 & 4.00 & 4.00 & 1 & 6 \\
ei\_KR\_on\_gam\_Mat52 & 10 & 6.30 & 7.00 & 2 & 12 \\
ei\_BOSS\_on\_LN\_Mat52 & 10 & 6.40 & 5.50 & 3 & 11 \\
lcb\_BOSS\_on\_LN\_Mat52 & 10 & 6.90 & 7.00 & 5 & 12 \\
lcb\_KR\_on\_gam\_Mat52 & 10 & 8.90 & 9.50 & 2 & 16 \\
lcb\_BOSS\_on\_gam\_fixed\_Mat52 & 10 & 9.30 & 10.00 & 5 & 13 \\
ei\_BOSS\_on\_gam\_fixed\_Mat52 & 10 & 9.60 & 8.00 & 6 & 14 \\
lcb\_hvafner\_fixed & 10 & 10.70 & 11.00 & 8 & 15 \\
ei\_hvafner\_fixed & 10 & 11.80 & 11.50 & 6 & 16 \\
ei\_BOSS\_off\_Mat52 & 10 & 12.90 & 12.50 & 10 & 16 \\
lcb\_BOSS\_off\_Mat52 & 10 & 13.00 & 13.50 & 9 & 16 \\
lcb\_meta\_off & 10 & 13.30 & 13.00 & 10 & 16 \\
ei\_meta\_off & 10 & 14.40 & 14.50 & 11 & 17 \\
ei\_BOSS\_off\_RBF & 10 & 17.50 & 17.00 & 17 & 19 \\
lcb\_BOSS\_off\_RBF & 10 & 17.50 & 18.00 & 15 & 19 \\
SOBOL\_off & 10 & 18.80 & 19.00 & 18 & 19 \\
\bottomrule
\end{tabular}
\caption{Summary of all model composite score ranking statistics across only the continuous+integer variable domains (10 variants) of the BS function under the medium tolerance level.}
\label{tab:SI:cont_int_medium}
\end{table}

\begin{table}[H]
\centering
\begin{tabular}{lrrrrr}
\toprule
\textbf{Model Settings} & \textbf{Num Ranks} & \textbf{Mean R.} & \textbf{Median R.} & \textbf{Min R.} & \textbf{Max R.} \\
\midrule
ei\_BOSS\_off\_Mat52\_sum & 10 & 1.40 & 1.00 & 1 & 3 \\
lcb\_BOSS\_off\_Mat52\_sum & 10 & 2.50 & 2.00 & 1 & 6 \\
ei\_BOSS\_on\_gam\_Mat52 & 10 & 4.10 & 4.00 & 2 & 6 \\
ei\_BOSS\_on\_LN\_Mat52 & 10 & 4.20 & 4.50 & 1 & 6 \\
lcb\_BOSS\_on\_LN\_Mat52 & 10 & 4.30 & 5.00 & 1 & 6 \\
lcb\_BOSS\_on\_gam\_Mat52 & 10 & 4.50 & 4.00 & 3 & 6 \\
ei\_BOSS\_on\_gam\_fixed\_Mat52 & 10 & 8.10 & 8.00 & 7 & 10 \\
ei\_BOSS\_off\_Mat52 & 10 & 9.00 & 8.00 & 7 & 16 \\
lcb\_BOSS\_on\_gam\_fixed\_Mat52 & 10 & 9.60 & 9.50 & 7 & 14 \\
lcb\_BOSS\_off\_Mat52 & 10 & 9.80 & 10.00 & 7 & 13 \\
lcb\_hvafner\_fixed & 10 & 11.10 & 11.00 & 9 & 14 \\
ei\_hvafner\_fixed & 10 & 11.70 & 12.50 & 8 & 15 \\
lcb\_meta\_off & 10 & 12.70 & 12.50 & 10 & 15 \\
ei\_meta\_off & 10 & 12.80 & 13.00 & 10 & 14 \\
ei\_BOSS\_off\_RBF & 10 & 14.80 & 15.00 & 11 & 16 \\
lcb\_BOSS\_off\_RBF & 10 & 15.40 & 15.50 & 14 & 16 \\
SOBOL\_off & 10 & 17.00 & 17.00 & 17 & 17 \\
\bottomrule
\end{tabular}
\caption{Summary of all model composite score ranking statistics across only the discrete variable domains (10 variants) of the BS function under the medium tolerance level.}
\label{tab:SI:discrete_medium}
\end{table}

\subsubsection{Strict Tolerance}

\begin{figure}[H]
    \centering
    \includegraphics[width=1\linewidth]{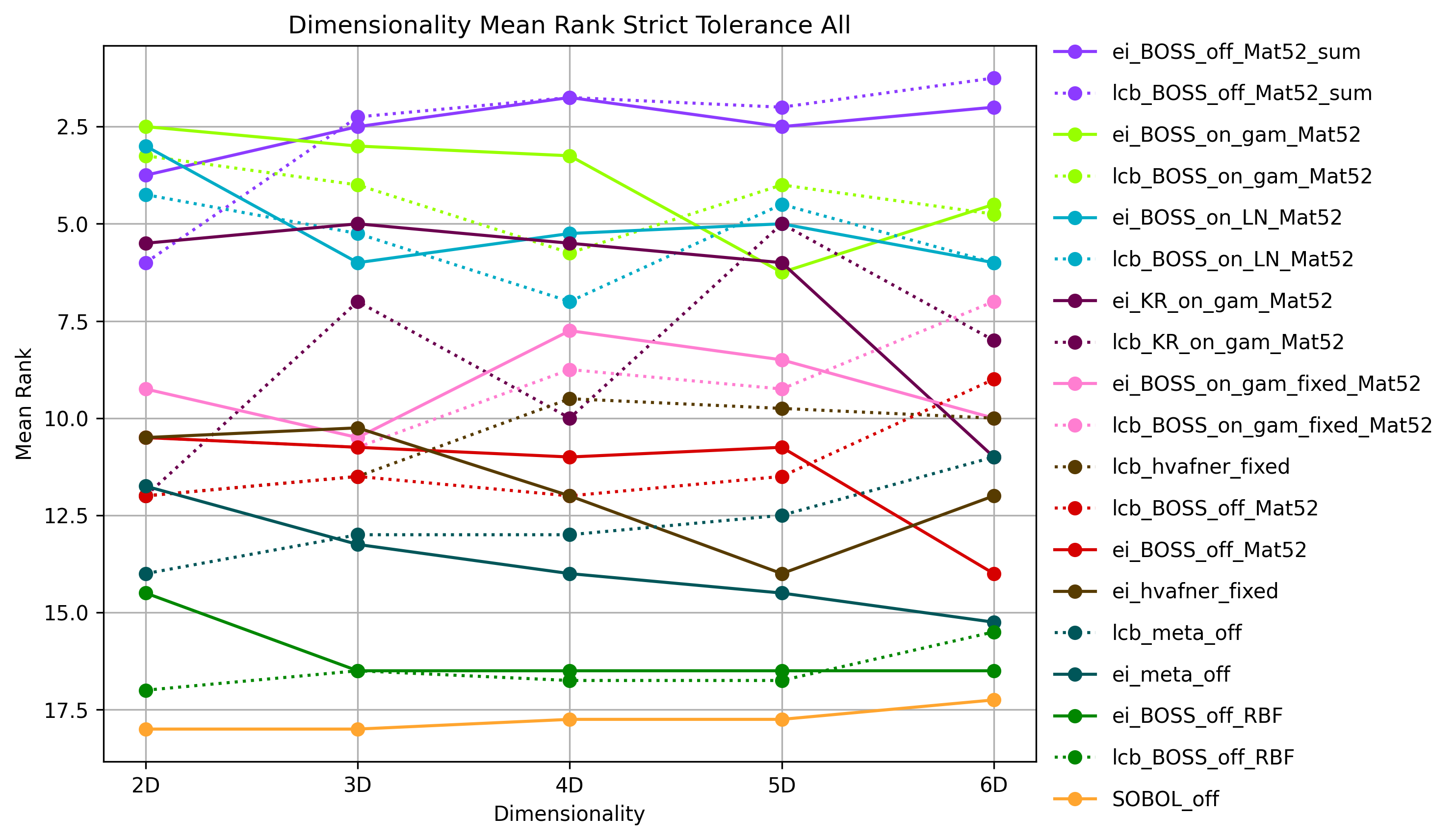}
    \caption{Plot showing the mean composite score ranks of all models across each dimension of the Butternut Squash benchmark function variants with continuous+integer domains under the strict tolerance.}
    \label{fig:SI:Dimension_all_str}
\end{figure}

\begin{figure}[H]
    \centering
    \includegraphics[width=1\linewidth]{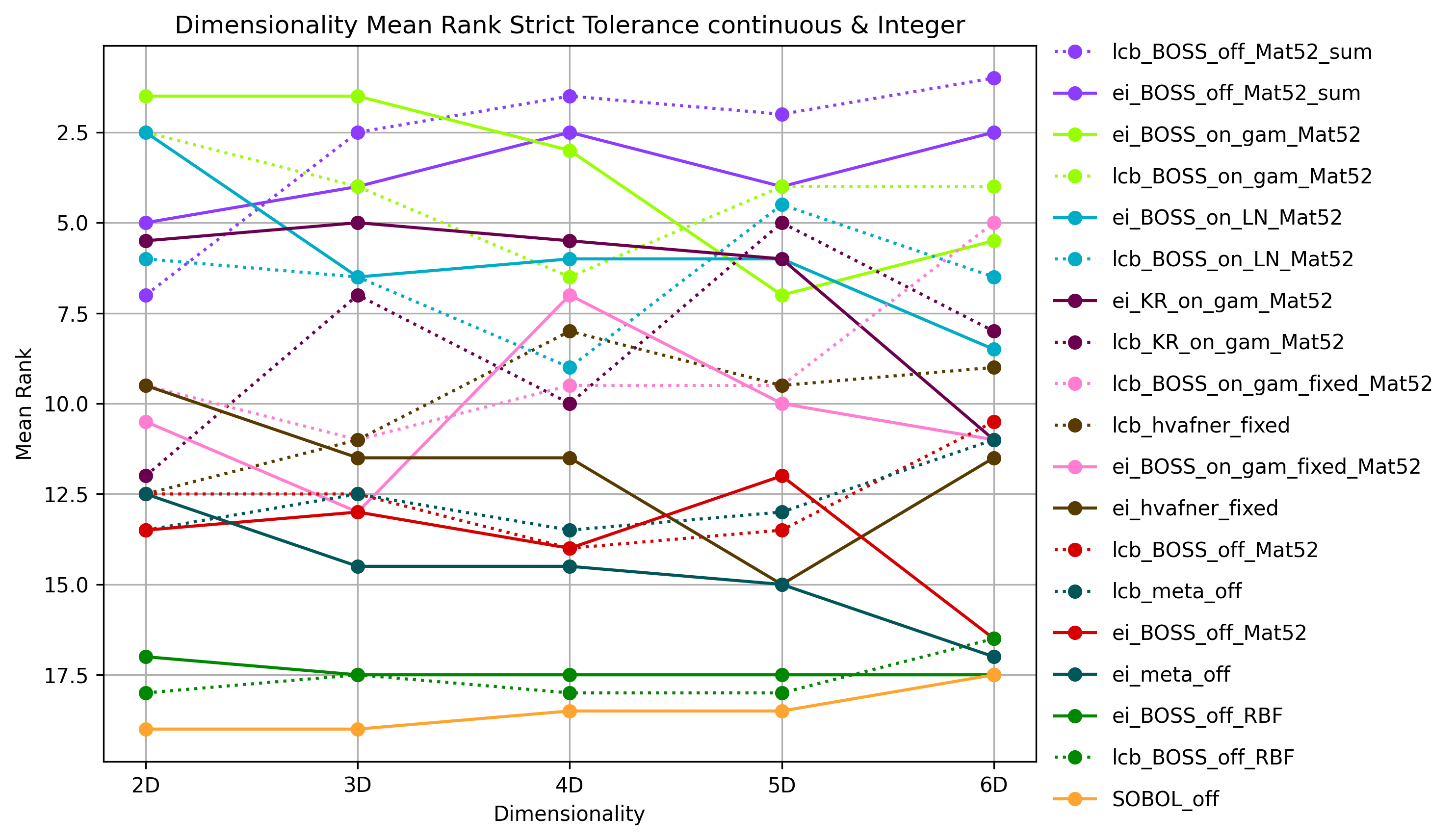}
    \caption{Plot showing the mean composite score ranks of all models across each dimension of the Butternut Squash benchmark function variants with continuous+integer domains under the strict tolerance.}
    \label{fig:SI:Dimension_cont_int_str}
\end{figure}

\begin{figure}[H]
    \centering
    \includegraphics[width=1\linewidth]{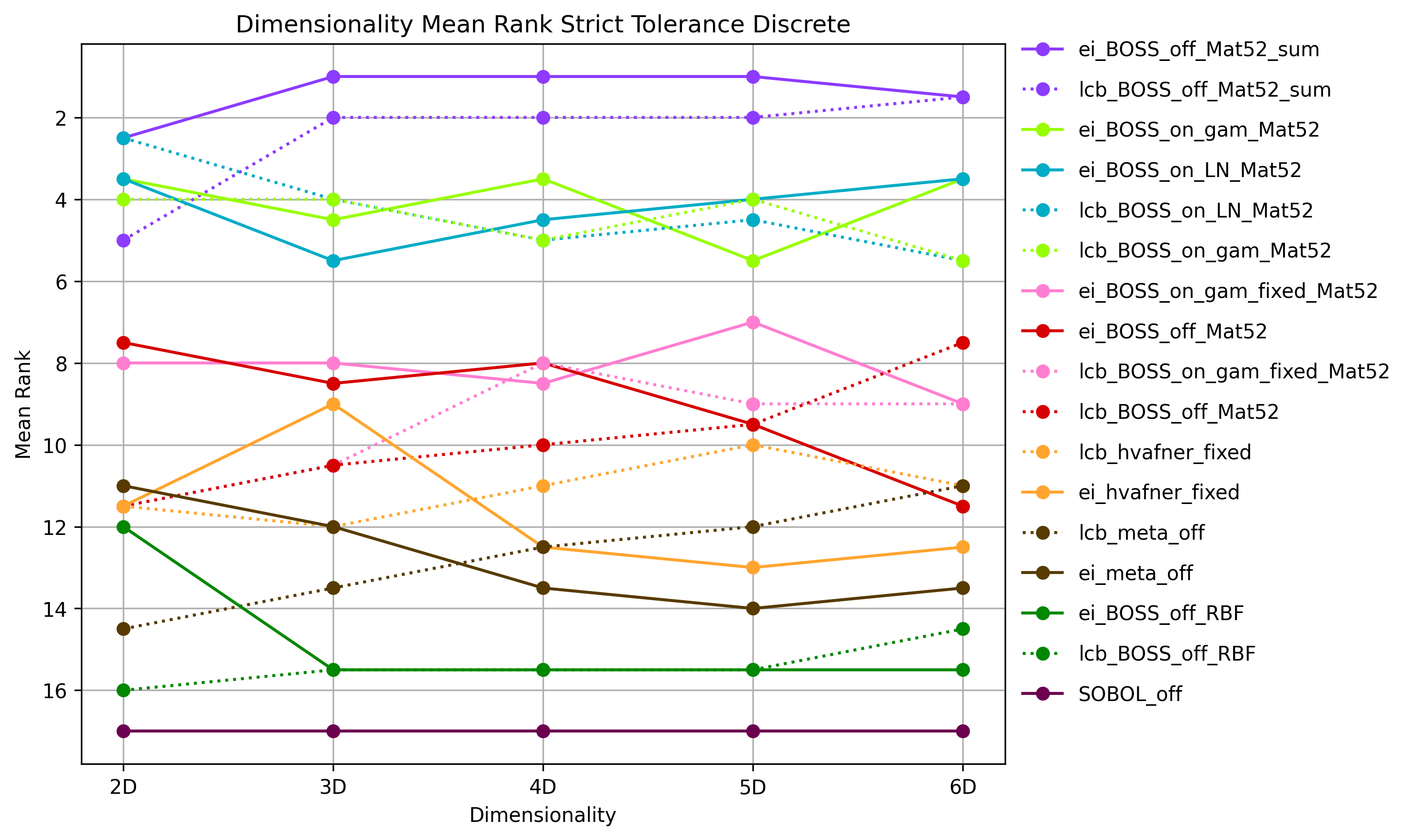}
    \caption{Plot showing the mean composite score ranks of all models across each dimension of the Butternut Squash benchmark function variants with discrete domains under the strict tolerance.}
    \label{fig:SI:Dimension_discrete_str}
\end{figure}

In the tables below, R is abbreviated as Rank.

\begin{table}[H]
\centering
\begin{tabular}{lrrrrr}
\toprule
\textbf{Model Settings} & \textbf{Num Ranks} & \textbf{Mean R.} & \textbf{Median R.} & \textbf{Min R.} & \textbf{Max R.} \\
\midrule
ei\_BOSS\_off\_Mat52\_sum & 20 & 2.50 & 2.00 & 1 & 6 \\
lcb\_BOSS\_off\_Mat52\_sum & 20 & 2.65 & 2.00 & 1 & 9 \\
ei\_BOSS\_on\_gam\_Mat52 & 20 & 3.90 & 3.00 & 1 & 11 \\
lcb\_BOSS\_on\_gam\_Mat52 & 20 & 4.35 & 4.00 & 1 & 8 \\
ei\_BOSS\_on\_LN\_Mat52 & 20 & 5.05 & 5.00 & 1 & 11 \\
lcb\_BOSS\_on\_LN\_Mat52 & 20 & 5.40 & 5.00 & 1 & 12 \\
ei\_KR\_on\_gam\_Mat52 & 10 & 6.60 & 7.00 & 3 & 12 \\
lcb\_KR\_on\_gam\_Mat52 & 10 & 8.40 & 8.50 & 1 & 16 \\
ei\_BOSS\_on\_gam\_fixed\_Mat52 & 20 & 9.20 & 8.00 & 7 & 14 \\
lcb\_BOSS\_on\_gam\_fixed\_Mat52 & 20 & 9.25 & 9.50 & 5 & 14 \\
lcb\_hvafner\_fixed & 20 & 10.55 & 11.00 & 5 & 15 \\
lcb\_BOSS\_off\_Mat52 & 20 & 11.20 & 11.00 & 7 & 16 \\
ei\_BOSS\_off\_Mat52 & 20 & 11.40 & 11.00 & 7 & 17 \\
ei\_hvafner\_fixed & 20 & 11.75 & 13.00 & 6 & 16 \\
lcb\_meta\_off & 20 & 12.70 & 12.50 & 9 & 16 \\
ei\_meta\_off & 20 & 13.75 & 14.00 & 10 & 19 \\
ei\_BOSS\_off\_RBF & 20 & 16.10 & 16.50 & 11 & 18 \\
lcb\_BOSS\_off\_RBF & 20 & 16.50 & 16.00 & 14 & 19 \\
SOBOL\_off & 20 & 17.75 & 17.00 & 16 & 19 \\
\bottomrule
\end{tabular}
\caption{Summary of model composite score ranking statistics across all 20 variants of the BS function under the strict tolerance level. Note that since the \textbf{KR} model only applies to the continuous+integer domains, it has only 10 computed ranks corresponding to the 10 integer+continuous domain variants of the BS functions.}
\label{tab:SI:all_strict}
\end{table}

\begin{table}[H]
\centering
\begin{tabular}{lrrrrr}
\toprule
\textbf{Model Settings} & \textbf{Num Ranks} & \textbf{Mean R.} & \textbf{Median R.} & \textbf{Min R.} & \textbf{Max R.} \\
\midrule
lcb\_BOSS\_off\_Mat52\_sum & 10 & 2.80 & 1.50 & 1 & 9 \\
ei\_BOSS\_off\_Mat52\_sum & 10 & 3.60 & 3.50 & 1 & 6 \\
ei\_BOSS\_on\_gam\_Mat52 & 10 & 3.70 & 2.50 & 1 & 11 \\
lcb\_BOSS\_on\_gam\_Mat52 & 10 & 4.20 & 4.00 & 1 & 8 \\
ei\_BOSS\_on\_LN\_Mat52 & 10 & 5.90 & 5.50 & 2 & 11 \\
lcb\_BOSS\_on\_LN\_Mat52 & 10 & 6.50 & 6.00 & 4 & 12 \\
ei\_KR\_on\_gam\_Mat52 & 10 & 6.60 & 7.00 & 3 & 12 \\
lcb\_KR\_on\_gam\_Mat52 & 10 & 8.40 & 8.50 & 1 & 16 \\
lcb\_BOSS\_on\_gam\_fixed\_Mat52 & 10 & 8.90 & 9.50 & 5 & 11 \\
lcb\_hvafner\_fixed & 10 & 10.00 & 10.50 & 5 & 15 \\
ei\_BOSS\_on\_gam\_fixed\_Mat52 & 10 & 10.30 & 10.00 & 7 & 14 \\
ei\_hvafner\_fixed & 10 & 11.80 & 13.00 & 6 & 16 \\
lcb\_BOSS\_off\_Mat52 & 10 & 12.60 & 12.50 & 9 & 16 \\
lcb\_meta\_off & 10 & 12.70 & 12.50 & 9 & 16 \\
ei\_BOSS\_off\_Mat52 & 10 & 13.80 & 13.50 & 11 & 17 \\
ei\_meta\_off & 10 & 14.70 & 15.00 & 12 & 19 \\
ei\_BOSS\_off\_RBF & 10 & 17.40 & 17.00 & 17 & 18 \\
lcb\_BOSS\_off\_RBF & 10 & 17.60 & 18.00 & 15 & 19 \\
SOBOL\_off & 10 & 18.50 & 19.00 & 16 & 19 \\
\bottomrule
\end{tabular}
\caption{Summary of all model composite score ranking statistics across only the continuous+integer variable domains (10 variants) of the BS function under the strict tolerance level.}
\label{tab:SI:cont_int_strict}
\end{table}

\begin{table}[H]
\centering
\begin{tabular}{lrrrrr}
\toprule
\textbf{Model Settings} & \textbf{Num Ranks} & \textbf{Mean R.} & \textbf{Median R.} & \textbf{Min R.} & \textbf{Max R.} \\
\midrule
ei\_BOSS\_off\_Mat52\_sum & 10 & 1.40 & 1.00 & 1 & 3 \\
lcb\_BOSS\_off\_Mat52\_sum & 10 & 2.50 & 2.00 & 1 & 6 \\
ei\_BOSS\_on\_gam\_Mat52 & 10 & 4.10 & 4.00 & 2 & 6 \\
ei\_BOSS\_on\_LN\_Mat52 & 10 & 4.20 & 4.50 & 1 & 6 \\
lcb\_BOSS\_on\_LN\_Mat52 & 10 & 4.30 & 5.00 & 1 & 6 \\
lcb\_BOSS\_on\_gam\_Mat52 & 10 & 4.50 & 4.00 & 3 & 6 \\
ei\_BOSS\_on\_gam\_fixed\_Mat52 & 10 & 8.10 & 8.00 & 7 & 10 \\
ei\_BOSS\_off\_Mat52 & 10 & 9.00 & 8.00 & 7 & 16 \\
lcb\_BOSS\_on\_gam\_fixed\_Mat52 & 10 & 9.60 & 9.50 & 7 & 14 \\
lcb\_BOSS\_off\_Mat52 & 10 & 9.80 & 10.00 & 7 & 13 \\
lcb\_hvafner\_fixed & 10 & 11.10 & 11.00 & 9 & 14 \\
ei\_hvafner\_fixed & 10 & 11.70 & 12.50 & 8 & 15 \\
lcb\_meta\_off & 10 & 12.70 & 12.50 & 10 & 15 \\
ei\_meta\_off & 10 & 12.80 & 13.00 & 10 & 14 \\
ei\_BOSS\_off\_RBF & 10 & 14.80 & 15.00 & 11 & 16 \\
lcb\_BOSS\_off\_RBF & 10 & 15.40 & 15.50 & 14 & 16 \\
SOBOL\_off & 10 & 17.00 & 17.00 & 17 & 17 \\
\bottomrule
\end{tabular}
\caption{Summary of all model composite score ranking statistics across only the discrete variable domains (10 variants) of the BS function under the strict tolerance level.}
\label{tab:SI:disc_strict}
\end{table}

\subsubsection{BS Convergence Plots}

Below we provide the convergence plots for all 20 variants of the BS function. Within the plots the letters \textit{c,i,d} represent continuous, integer and discrete variables respectively, such that for example \textit{ccii} means that the first 2 dimensions of the function take continuous values and latter two integer values.

\begin{figure}[H]
    \centering
    \includegraphics[width=1\linewidth]{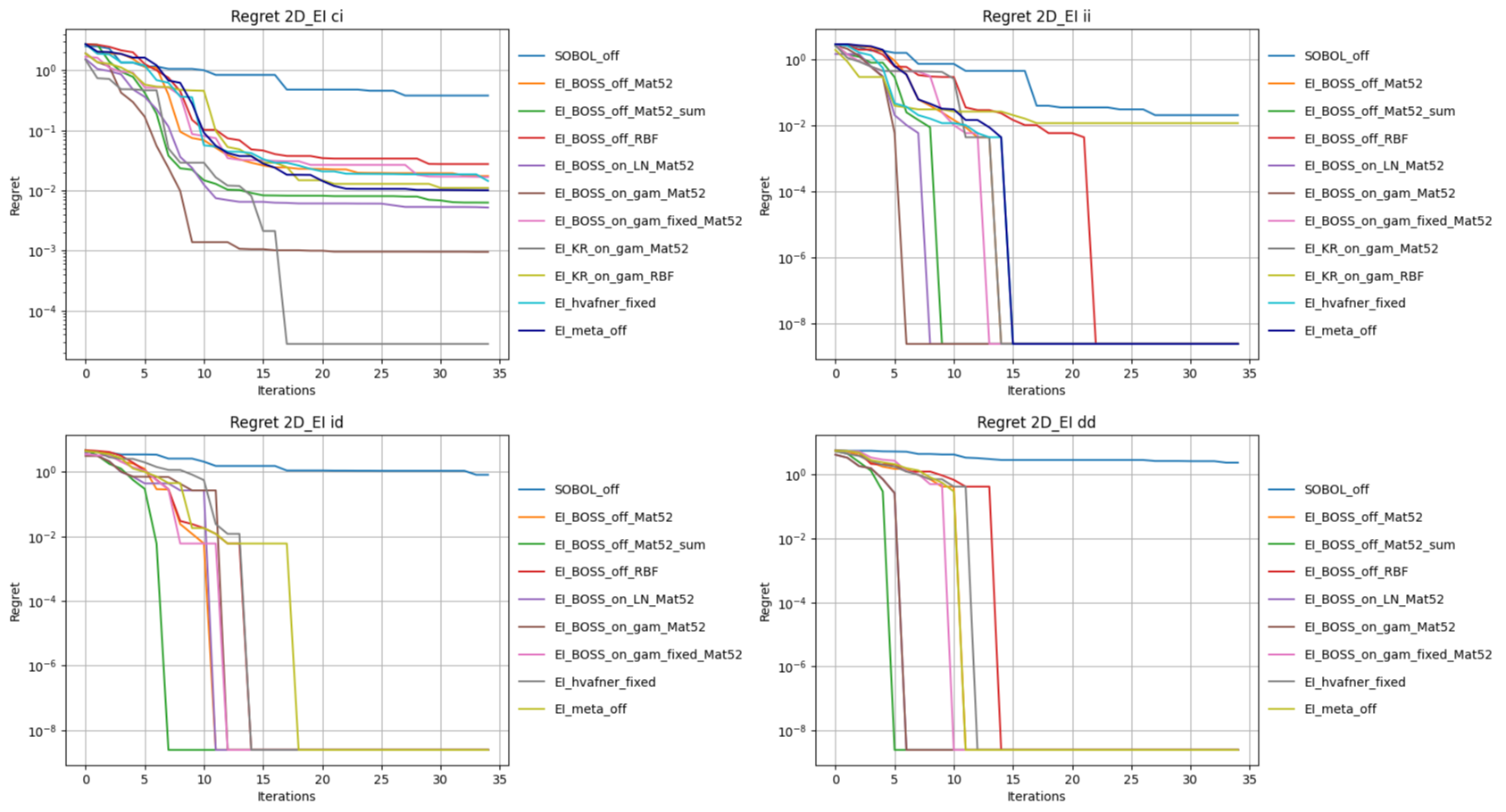}
    \caption{Mean convergence plots over 10 runs for all benchmarked models using the EI acquisition function on the 2D BS function.}
    \label{fig:SI:BS2DEI}
\end{figure}

\begin{figure}[H]
    \centering
    \includegraphics[width=1\linewidth]{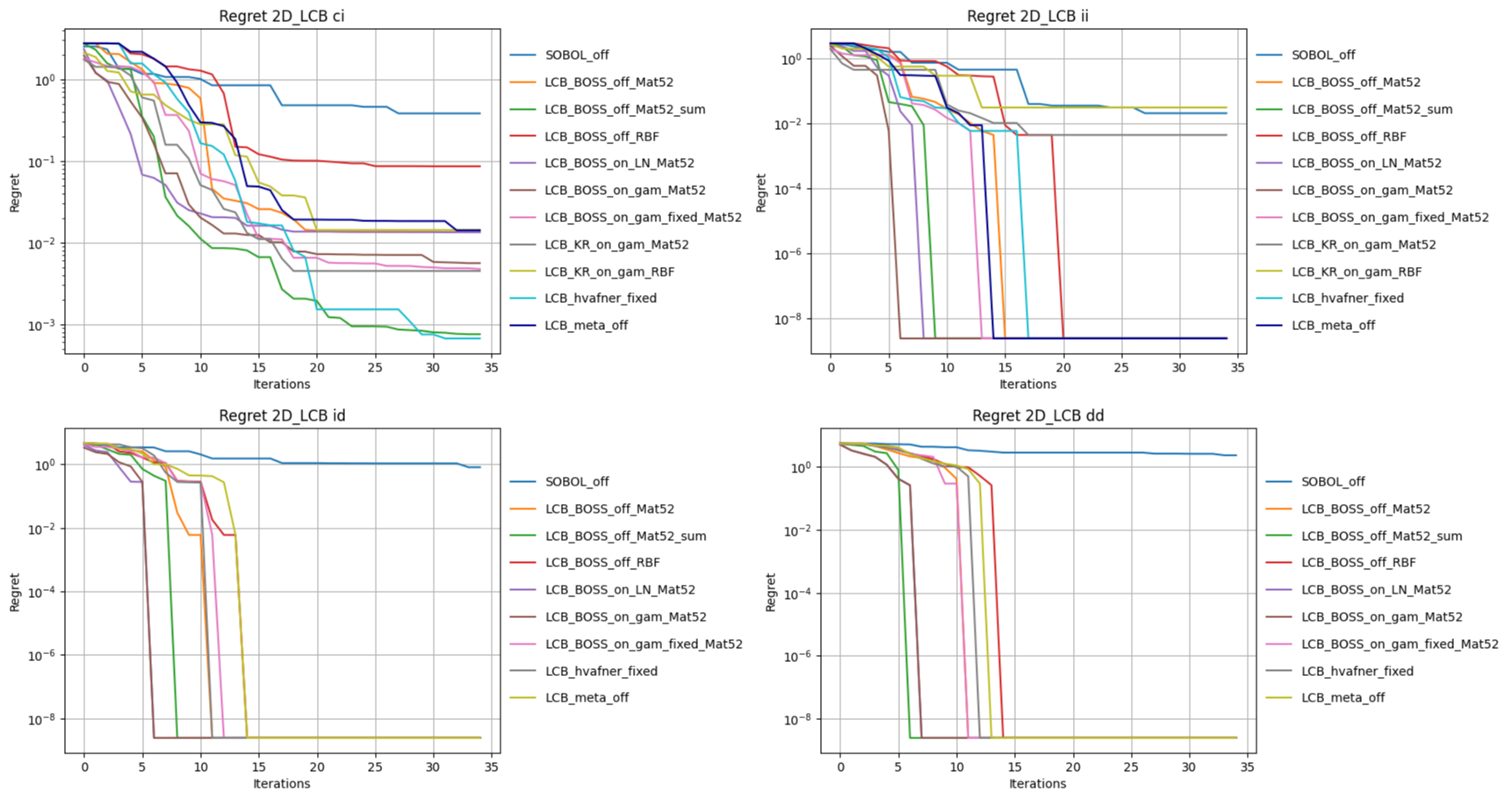}
    \caption{Mean convergence plots over 10 runs for all benchmarked models using the LCB acquisition function on the 2D BS function.}
    \label{fig:SI:BS2DLCB}
\end{figure}

\begin{figure}[H]
    \centering
    \includegraphics[width=1\linewidth]{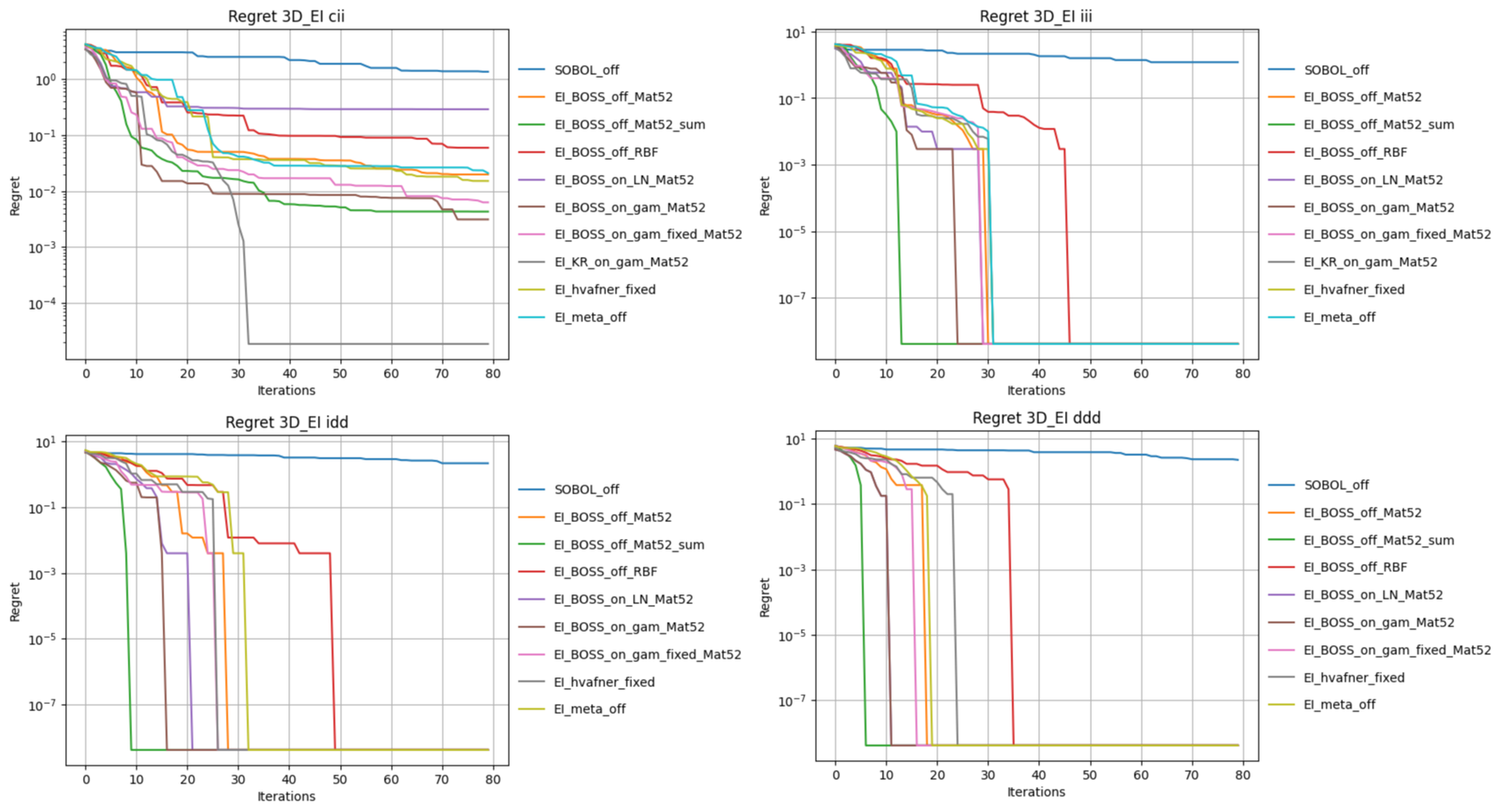}
    \caption{Mean convergence plots over 10 runs for all benchmarked models using the EI acquisition function on the 3D BS function.}
    \label{fig:SI:BS3DEI}
\end{figure}

\begin{figure}[H]
    \centering
    \includegraphics[width=1\linewidth]{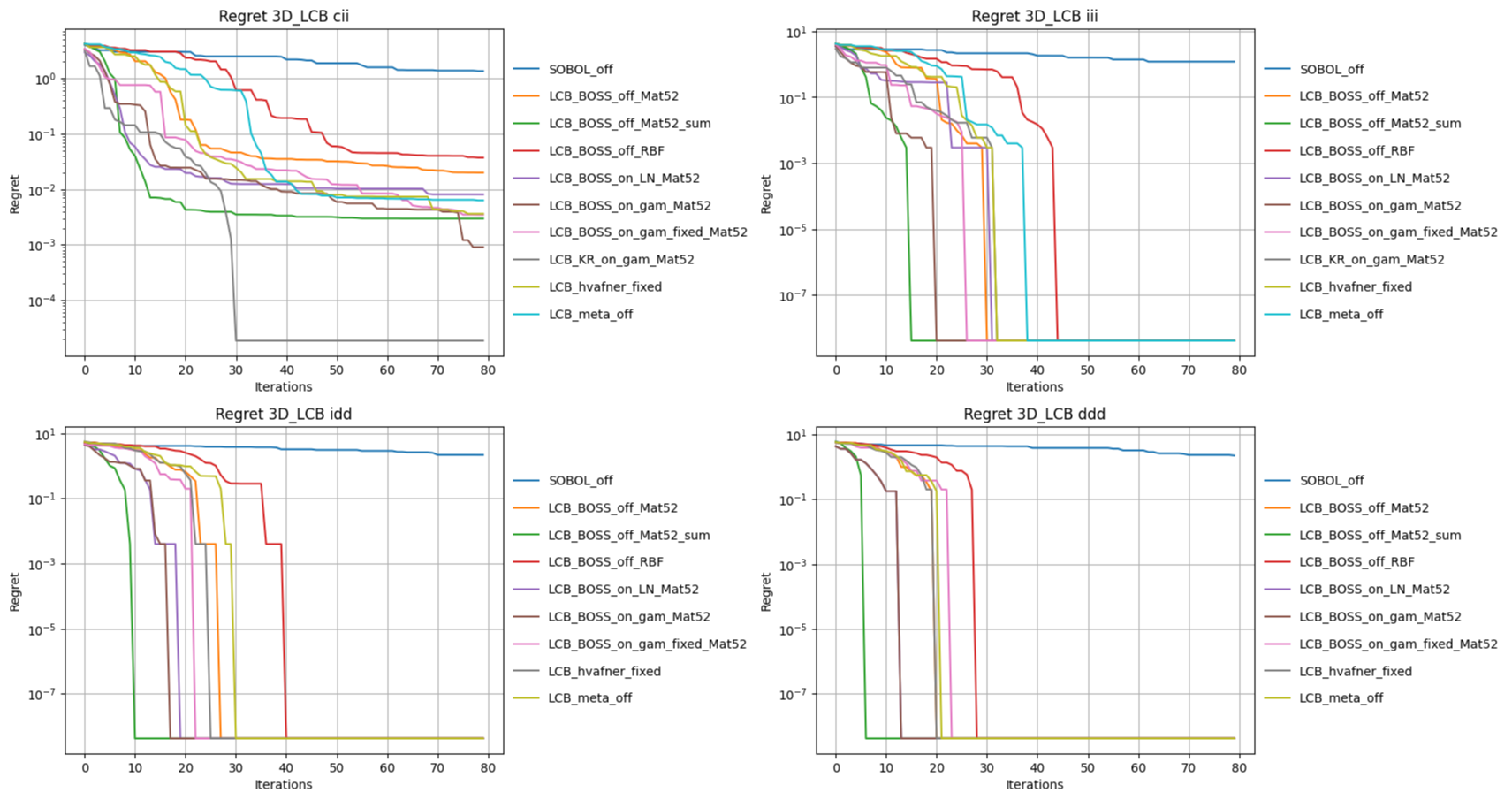}
    \caption{Mean convergence plots over 10 runs for all benchmarked models using the LCB acquisition function on the 3D BS function.}
    \label{fig:SI:BS3DLCB}
\end{figure}

\begin{figure}[H]
    \centering
    \includegraphics[width=1\linewidth]{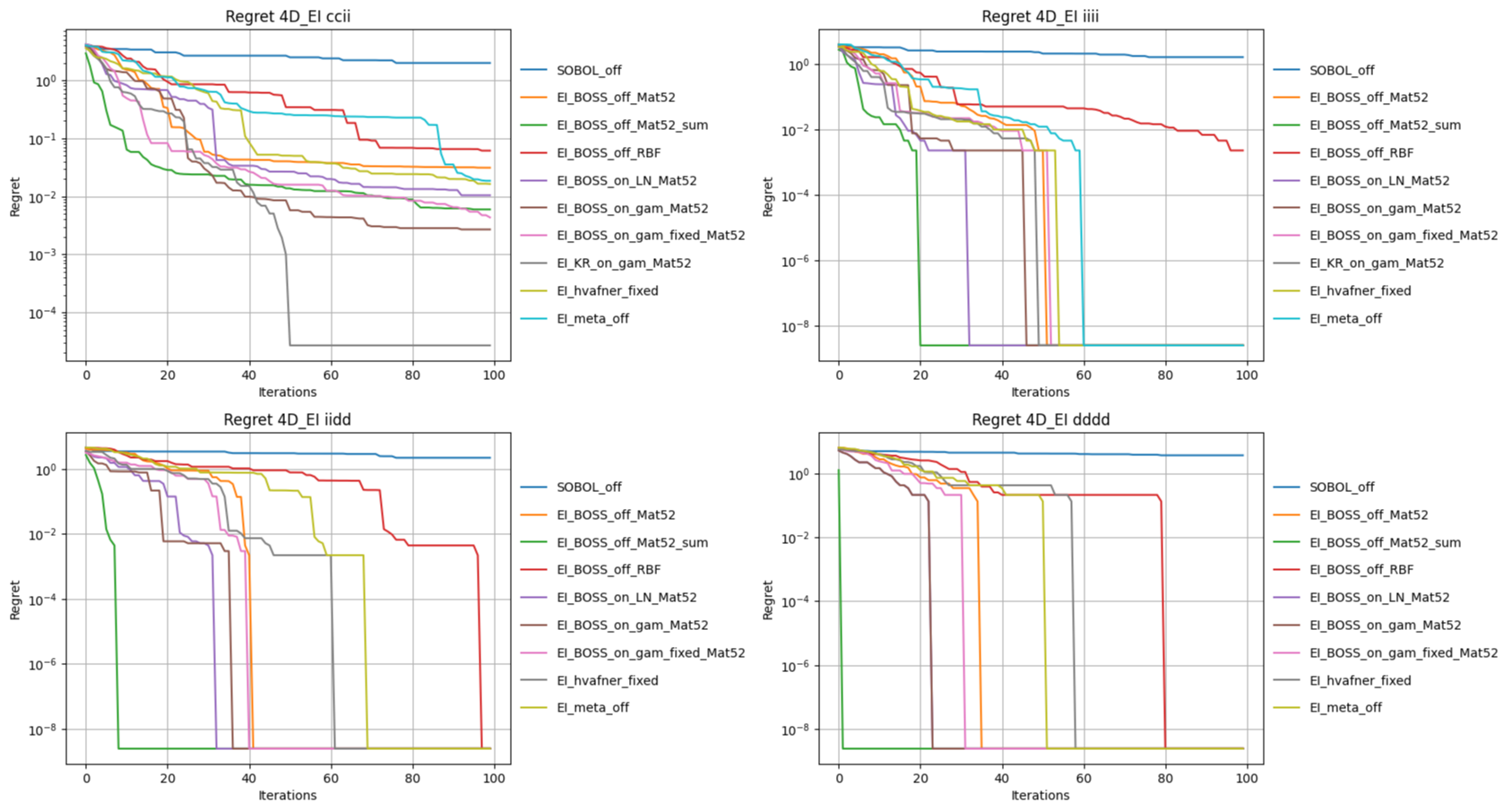}
    \caption{Mean convergence plots over 10 runs for all benchmarked models using the EI acquisition function on the 4D BS function.}
    \label{fig:SI:BS4DEI}
\end{figure}

\begin{figure}[H]
    \centering
    \includegraphics[width=1\linewidth]{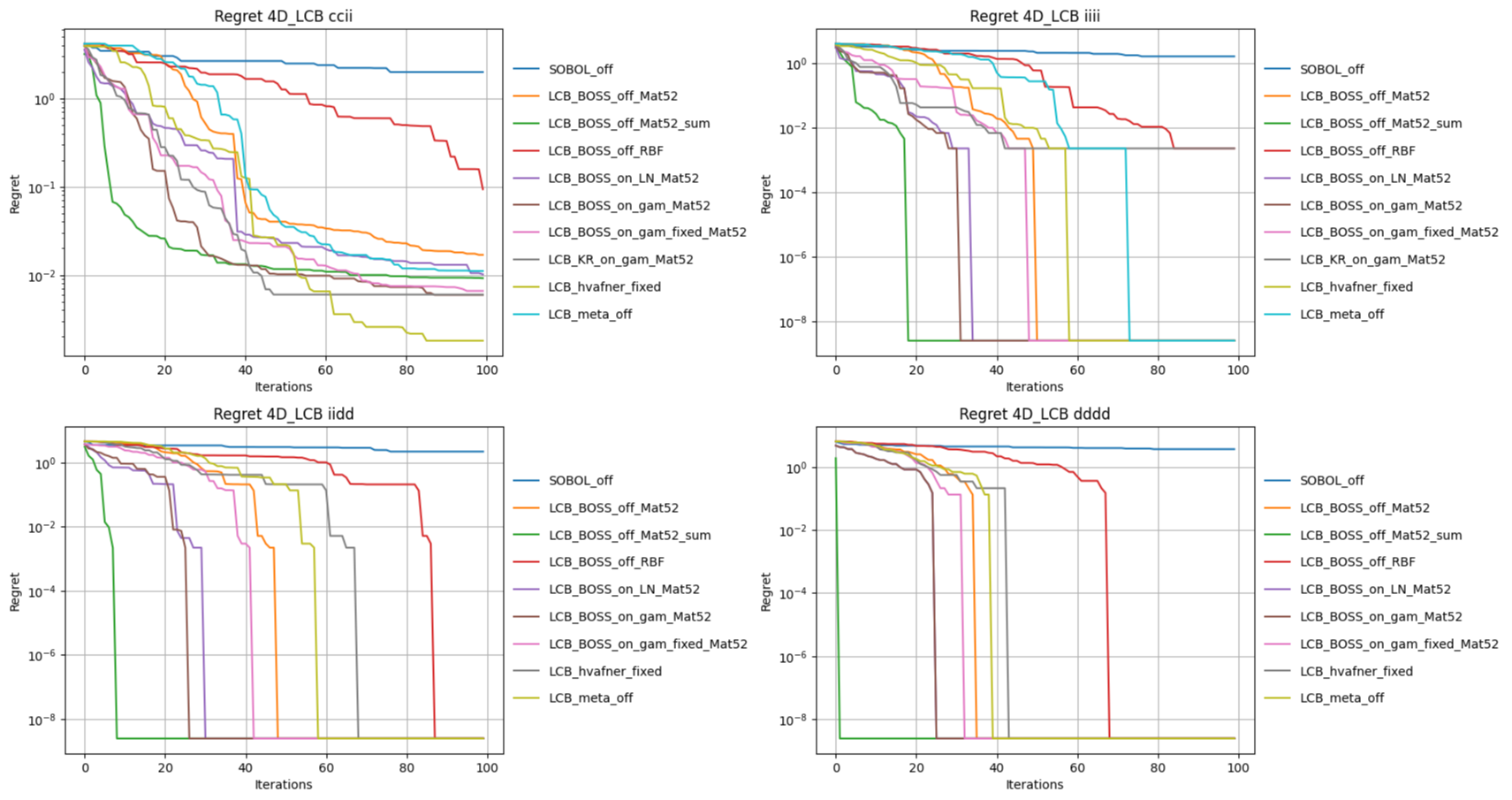}
    \caption{Mean convergence plots over 10 runs for all benchmarked models using the LCB acquisition function on the 4D BS function.}
    \label{fig:SI:BS4DLCB}
\end{figure}

\begin{figure}[H]
    \centering
    \includegraphics[width=1\linewidth]{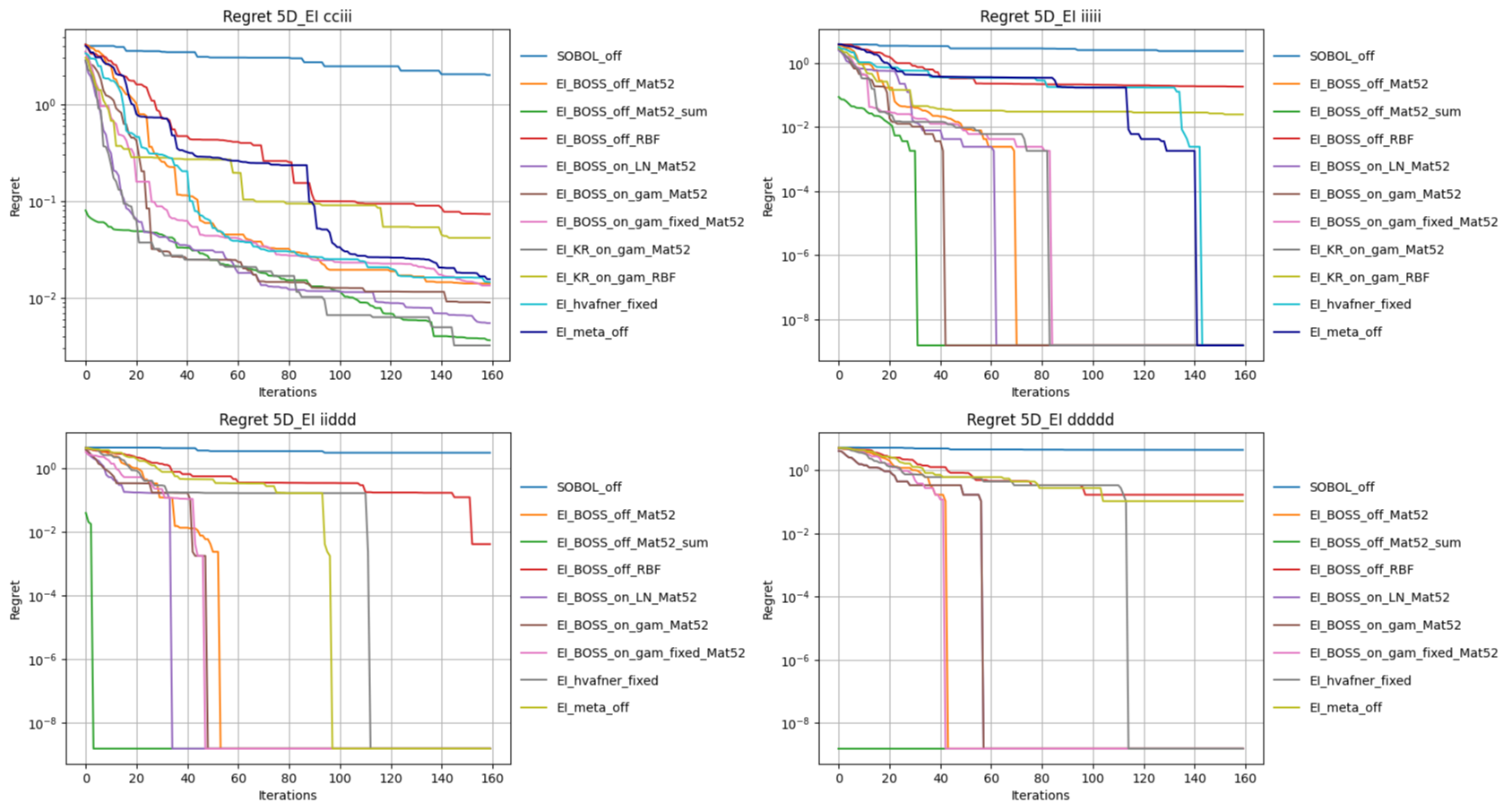}
    \caption{Mean convergence plots over 10 runs for all benchmarked models using the EI acquisition function on the 5D BS function.}
    \label{fig:SI:BS5DEI}
\end{figure}

\begin{figure}[H]
    \centering
    \includegraphics[width=1\linewidth]{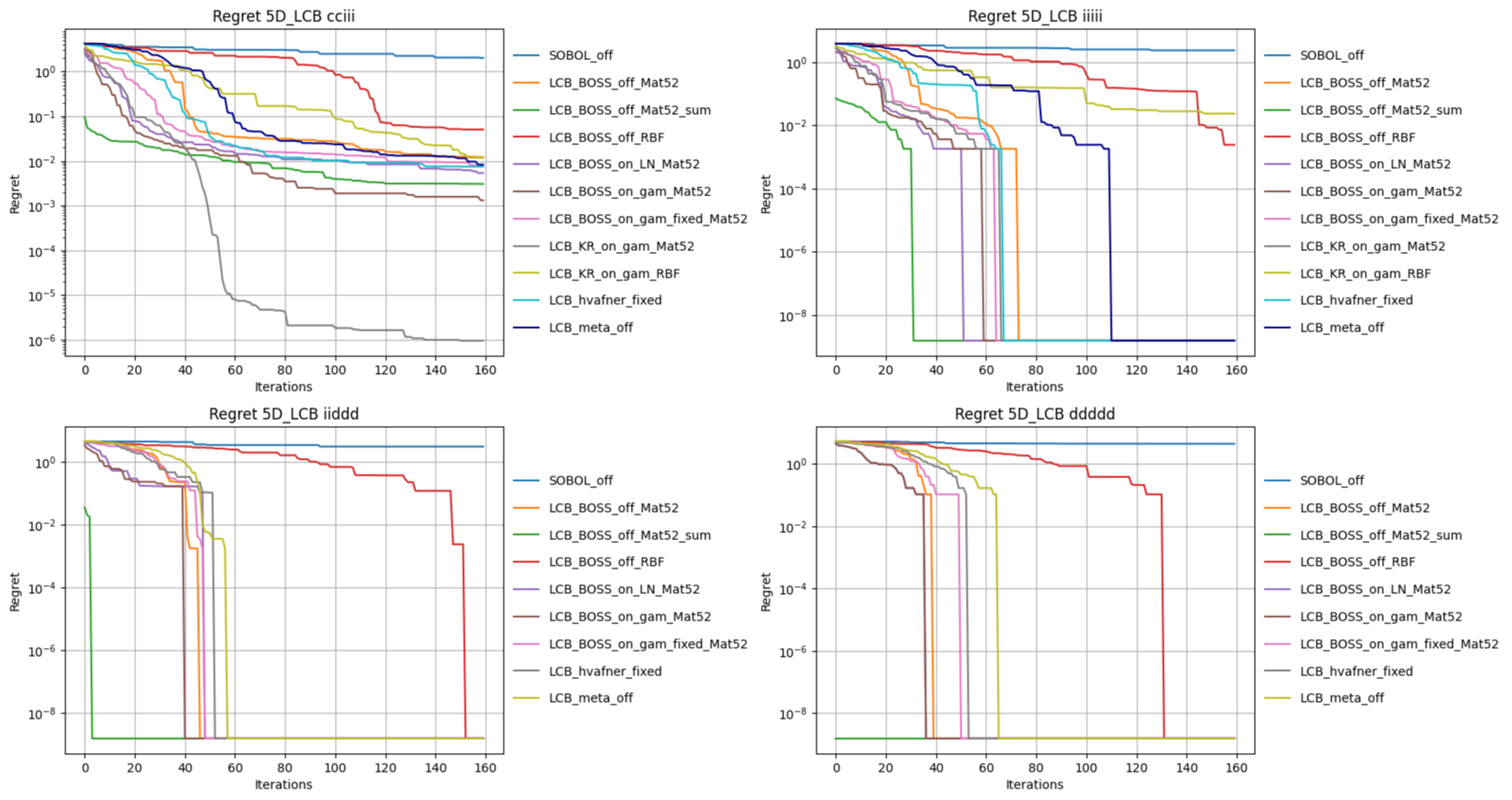}
    \caption{Mean convergence plots over 10 runs for all benchmarked models using the LCB acquisition function on the 5D BS function.}
    \label{fig:SI:B52DLCB}
\end{figure}

\begin{figure}[H]
    \centering
    \includegraphics[width=1\linewidth]{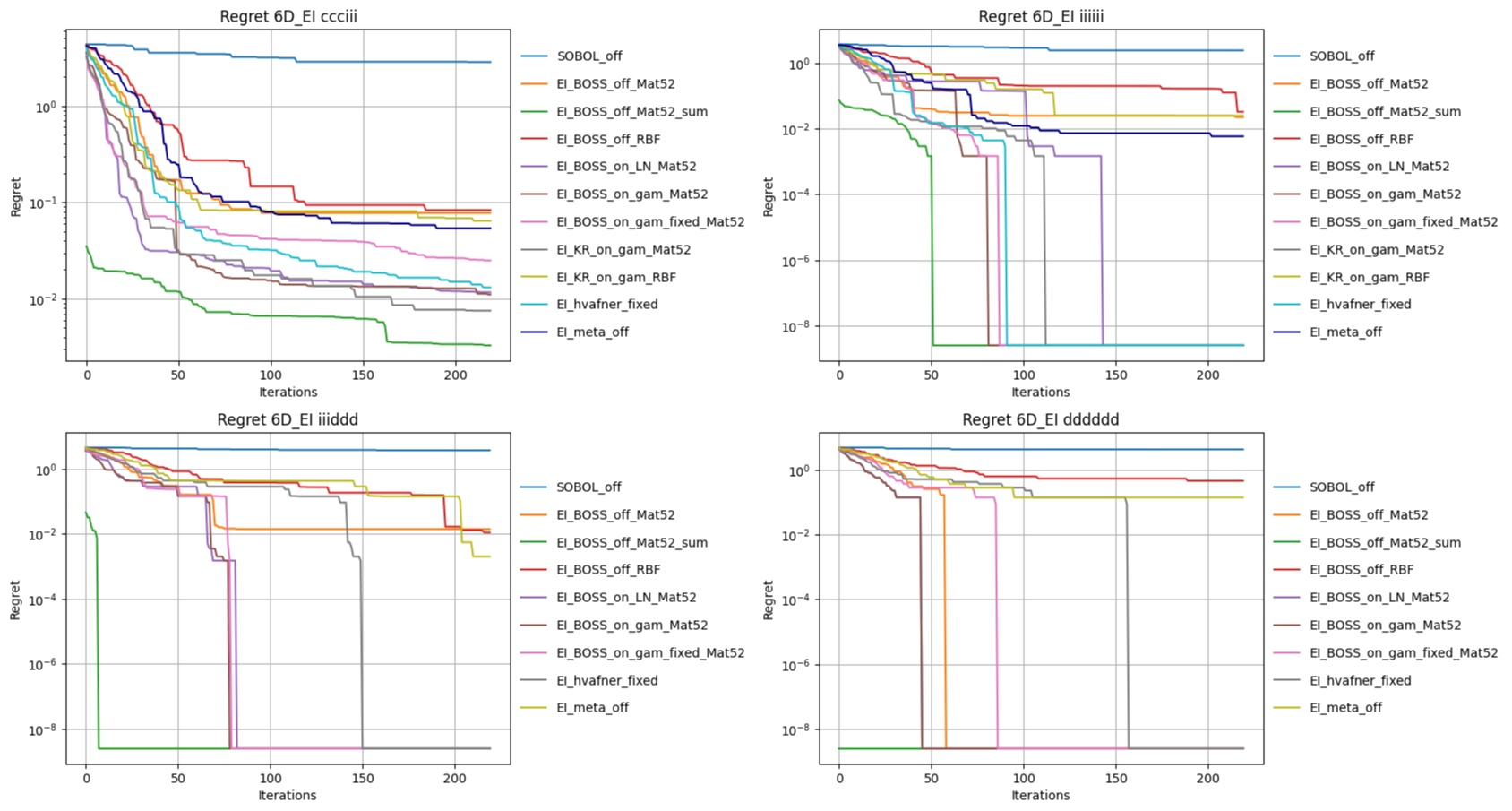}
    \caption{Mean convergence plots over 10 runs for all benchmarked models using the EI acquisition function on the 6D BS function.}
    \label{fig:SI:BS6DEI}
\end{figure}

\begin{figure}[H]
    \centering
    \includegraphics[width=1\linewidth]{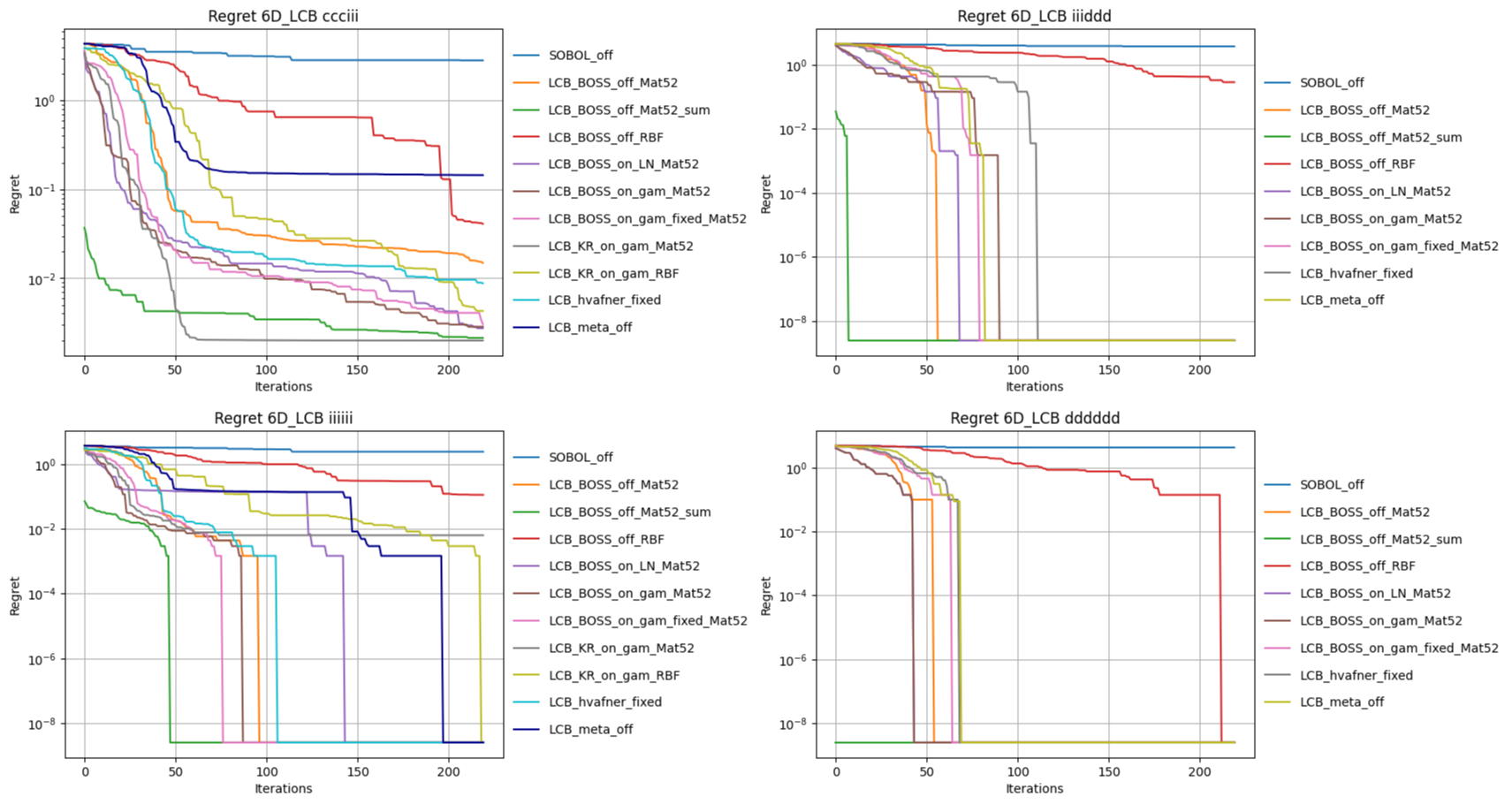}
    \caption{Mean convergence plots over 10 runs for all benchmarked models using the LCB acquisition function on the 6D BS function.}
    \label{fig:SI:B62DLCB}
\end{figure}

\subsection{Chemistry Benchmark Complementary Results}\label{SI:subsec:chem}

\begin{table}[H]
\centering
\begin{tabular}{lrrr}
\toprule
\textbf{Model Settings} & \textbf{Converged Runs} & \textbf{Mean Iteration} & \textbf{Composite Score} \\
\midrule
ei\_meta\_off & 6 & 37.67 & 0.015929 \\
ei\_BOSS\_on\_gam\_Mat52 & 5 & 41.40 & 0.012077 \\
lcb\_meta\_off & 4 & 41.00 & 0.009756 \\
lcb\_BOSS\_on\_gam\_Mat52 & 3 & 57.33 & 0.005233 \\
ei\_BOSS\_off\_Mat52\_sum & 0 & --- & 0.000000 \\
ei\_hvafner\_fixed & 0 & --- & 0.000000 \\
lcb\_BOSS\_off\_Mat52\_sum & 0 & --- & 0.000000 \\
lcb\_hvafner\_fixed & 0 & --- & 0.000000 \\
SOBOL\_off\_chem & 0 & --- & 0.000000 \\
\bottomrule
\end{tabular}
\caption{Composite score ranking summary of all benchmarked models on the Chemistry function under the strict tolerance level. Mean iteration is reported for successful runs only.}
\label{tab:SI:composite_score_chem_hightol}
\end{table}

\begin{table}[H]
\centering
\begin{tabular}{lrrr}
\toprule
\textbf{Model Settings} & \textbf{Converged Runs} & \textbf{Mean Iteration} & \textbf{Composite Score} \\
\midrule
ei\_BOSS\_on\_gam\_Mat52 & 6 & 35.83 & 0.016744 \\
ei\_meta\_off & 6 & 36.33 & 0.016514 \\
lcb\_meta\_off & 6 & 42.33 & 0.014173 \\
lcb\_BOSS\_on\_gam\_Mat52 & 6 & 59.50 & 0.010084 \\
ei\_BOSS\_off\_Mat52\_sum & 0 & --- & 0.000000 \\
ei\_hvafner\_fixed & 0 & --- & 0.000000 \\
lcb\_BOSS\_off\_Mat52\_sum & 0 & --- & 0.000000 \\
lcb\_hvafner\_fixed & 0 & --- & 0.000000 \\
SOBOL\_off\_chem & 0 & --- & 0.000000 \\
\bottomrule
\end{tabular}
\caption{Composite score ranking summary of all benchmarked models on the Chemistry function under the medium tolerance level. Mean iteration is reported for successful runs only.}
\label{tab:SI:composite_score_chem_medtol}
\end{table}

\begin{table}[H]
\centering
\begin{tabular}{lrrr}
\toprule
\textbf{Model Settings} & \textbf{Converged Runs} & \textbf{Mean Iteration} & \textbf{Composite Score} \\
\midrule
ei\_BOSS\_on\_gam\_Mat52 & 7 & 35.57 & 0.019679 \\
ei\_meta\_off & 7 & 40.43 & 0.017314 \\
lcb\_BOSS\_on\_gam\_Mat52 & 8 & 53.13 & 0.015059 \\
lcb\_meta\_off & 6 & 42.33 & 0.014173 \\
ei\_BOSS\_off\_Mat52\_sum & 1 & 56.00 & 0.001786 \\
ei\_hvafner\_fixed & 0 & --- & 0.000000 \\
lcb\_BOSS\_off\_Mat52\_sum & 0 & --- & 0.000000 \\
lcb\_hvafner\_fixed & 0 & --- & 0.000000 \\
SOBOL\_off\_chem & 0 & --- & 0.000000 \\
\bottomrule
\end{tabular}
\caption{Composite score ranking summary of all benchmarked models on the Chemistry function under the loose tolerance level. Mean iteration is reported for successful runs only.}
\label{tab:SI:composite_score_chem_lowtol}
\end{table}

\begin{figure}[H]
    \centering
    \includegraphics[width=1\linewidth]{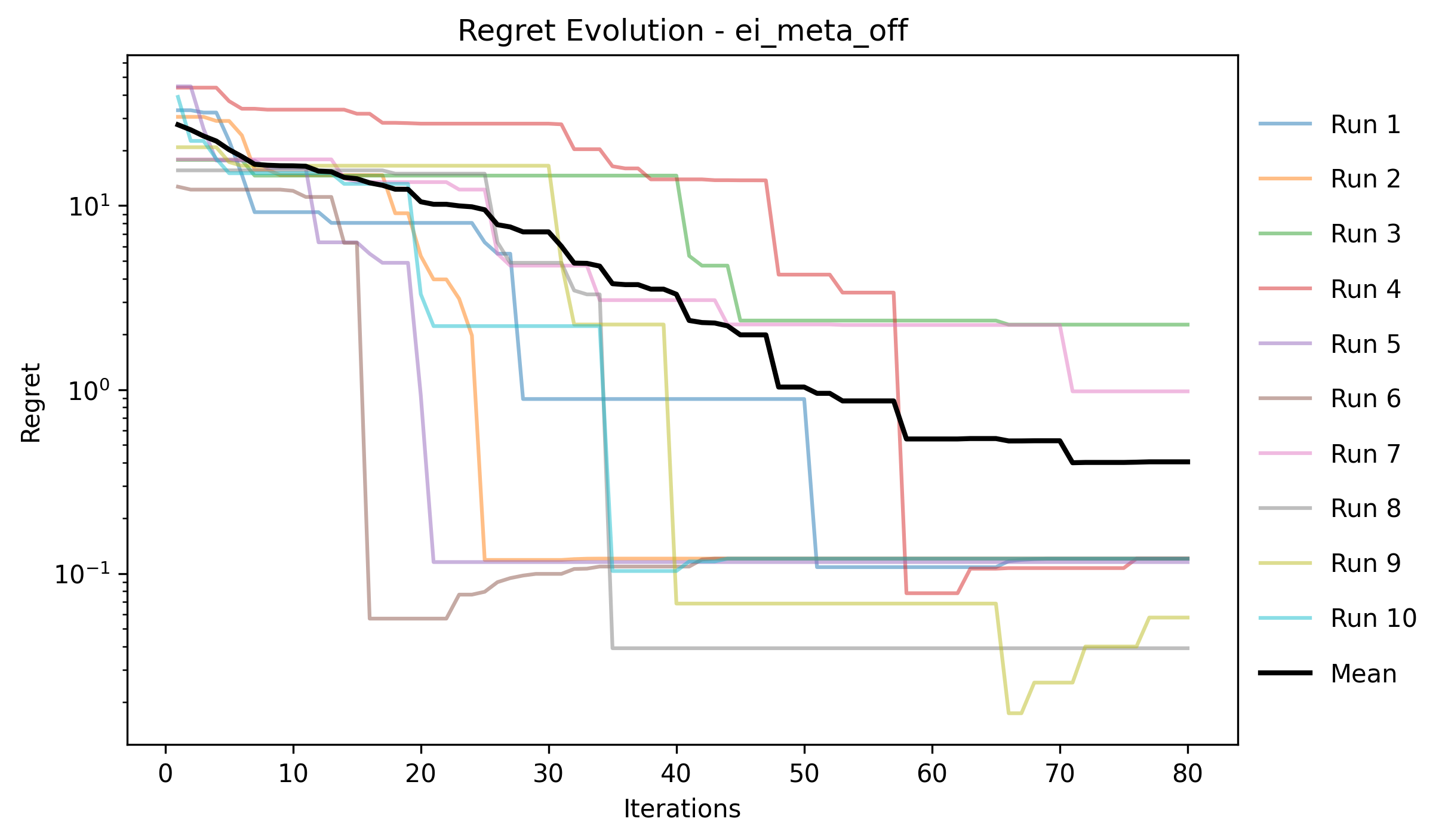}
    \caption{Convergence plots of the \textbf{EI meta} model on the Chemistry benchamrk function. The bold curve is the mean convergences from its 10 different sobol point initiated runs.}
    \label{fig:SI:meta_chem_runs}
\end{figure}

\begin{figure}[H]
    \centering
    \includegraphics[width=1\linewidth]{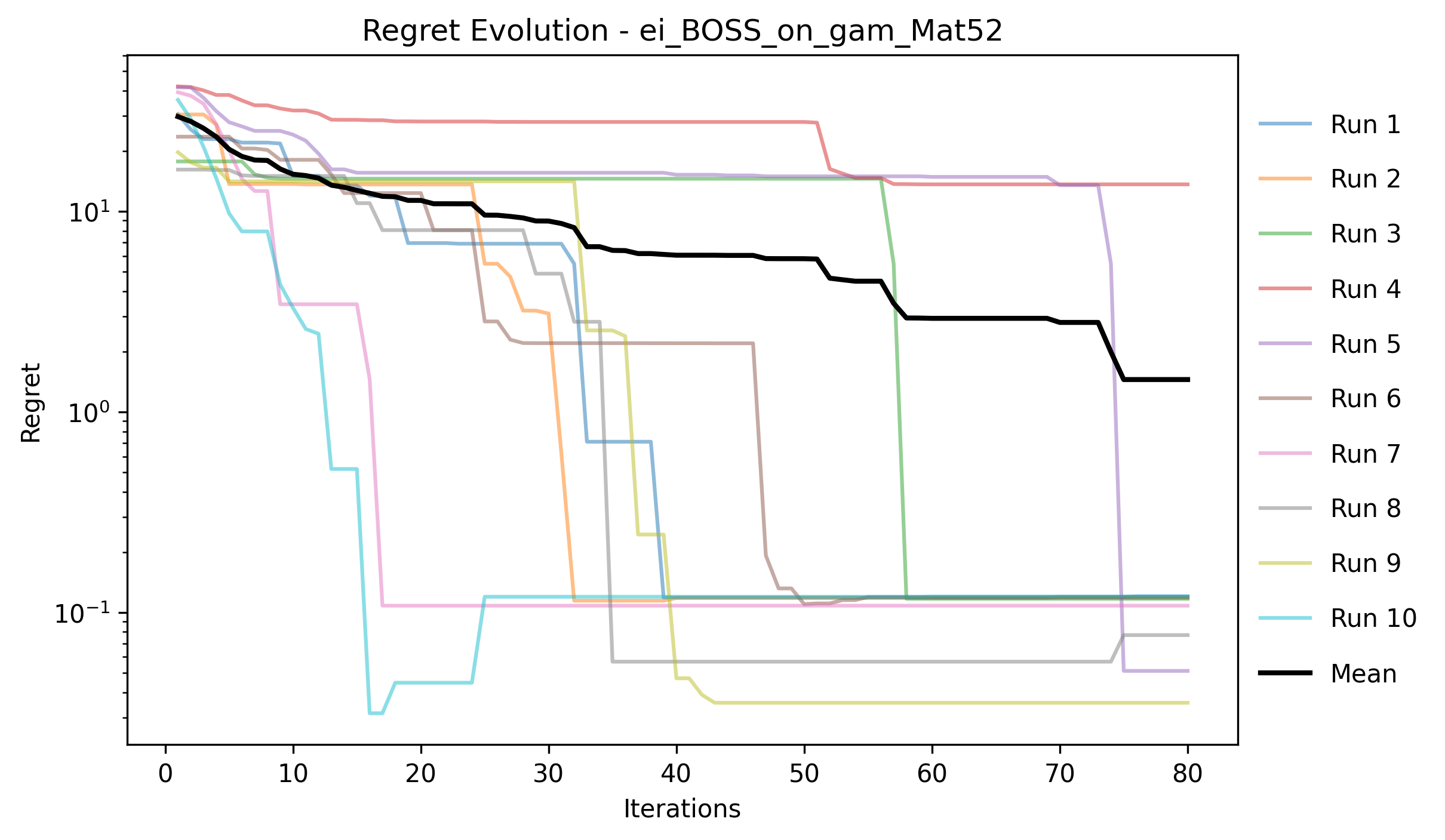}
    \caption{Convergence plots of the \textbf{EI PMG} model on the Chemistry benchamrk function. The bold curve is the mean convergences from its 10 different sobol point initiated runs.}
    \label{fig:SI:mat_chem_runs}
\end{figure}

\subsection{DUST Benchmark Complementary Results}\label{SI:subsec:DUST1}

\begin{figure}[H]
    \centering
    \includegraphics[width=1\linewidth]{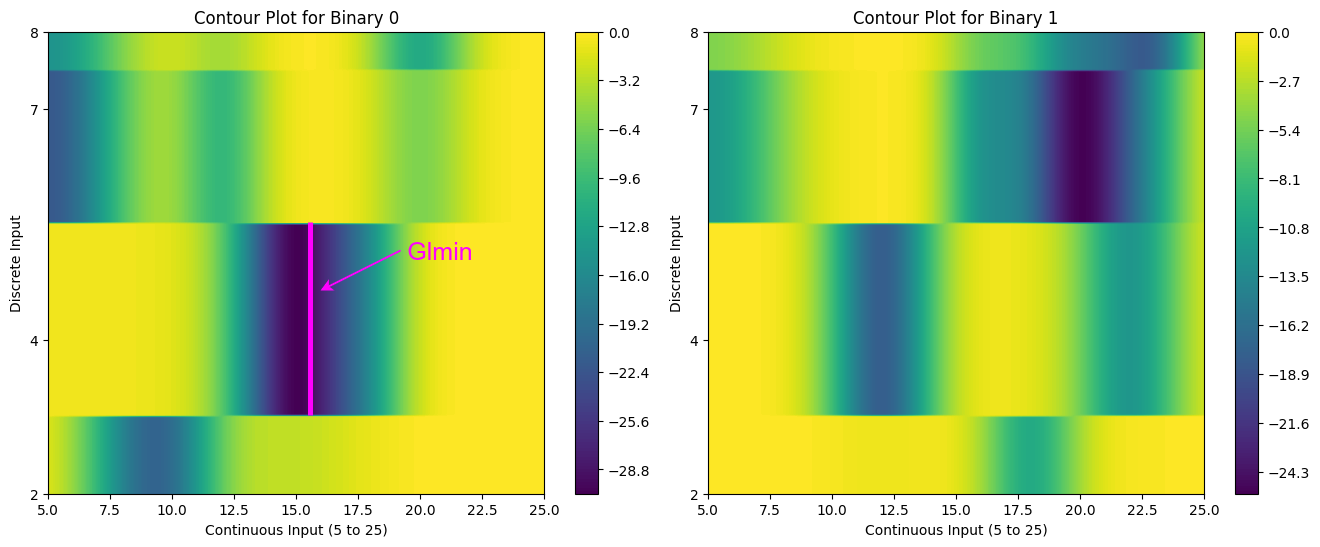}
    \caption{Objective Landscape of the DUST function with the global minima (Glmin) indicated in Pink.}
    \label{fig:SI:DUST_landscape}
\end{figure}

\begin{figure}[H]
    \centering
    \includegraphics[width=1\linewidth]{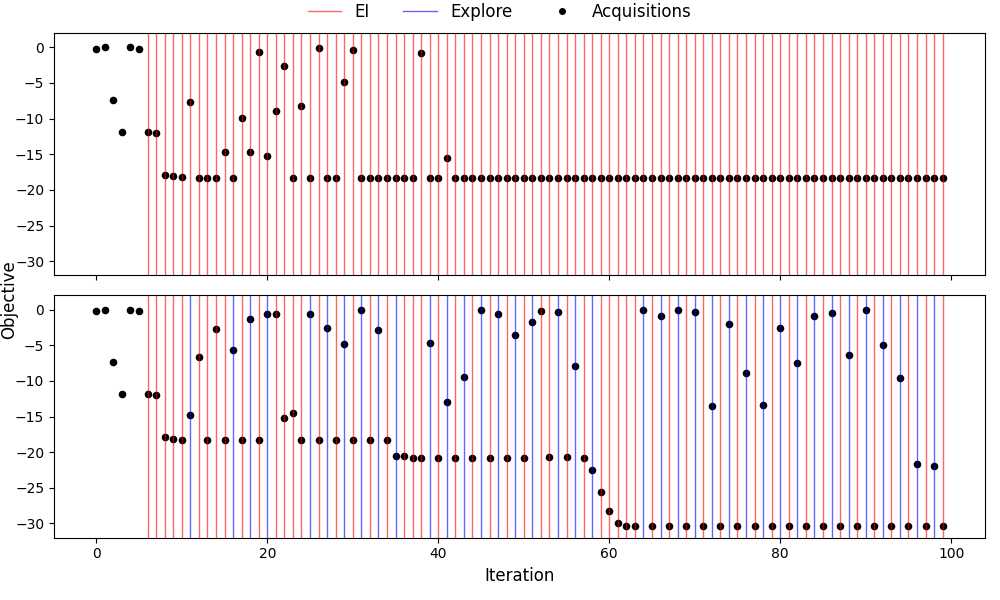}
    \caption{All acquisition points in terms of objective value from run 9 of the EI+MGP+V\_BO approach (up) versus the EI+MGP+EA\_BO approach (down). The acquisition function used at each iteration is shown using red and blue vertical lines corresponding to the EI and pure exploration AF respectively. The global minima objective value is -30}
    \label{fig:SI:explore_seed9}
\end{figure}

\begin{table}[H]
\centering
\begin{tabular}{lrrr}
\toprule
\textbf{Model Settings} & \textbf{Converged Runs} & \textbf{Mean Iteration} & \textbf{Composite Score} \\
\midrule
lcb\_penalty & 6 & 25.33 & 0.023684 \\
lcb\_penalty\_HITL\_explore & 10 & 44.10 & 0.022676 \\
ei\_penalty & 5 & 24.60 & 0.020325 \\
ei\_penalty\_HITL\_explore & 7 & 42.86 & 0.016333 \\
ei\_RF & 4 & 28.00 & 0.014286 \\
SOBOL\_off & 1 & 8.00 & 0.012500 \\
\bottomrule
\end{tabular}
\caption{Composite score ranking summary of all benchmarked models on the DUST1 function under the strict tolerance level. Mean iteration is reported for successful runs only.}
\label{tab:SI:composite_score_DUST1_strict}
\end{table}

\begin{table}[H]
\centering
\begin{tabular}{lrrr}
\toprule
\textbf{Model Settings} & \textbf{Converged Runs} & \textbf{Mean Iteration} & \textbf{Composite Score} \\
\midrule
ei\_penalty & 7 & 14.71 & 0.047573 \\
ei\_penalty\_HITL\_explore & 10 & 29.10 & 0.034364 \\
lcb\_penalty\_HITL\_explore & 10 & 31.00 & 0.032258 \\
lcb\_penalty & 6 & 21.00 & 0.028571 \\
SOBOL\_off & 3 & 23.00 & 0.013043 \\
ei\_RF & 5 & 40.80 & 0.012255 \\
\bottomrule
\end{tabular}
\caption{Composite score ranking summary of all benchmarked models on the DUST1 function under the medium tolerance level. Mean iteration is reported for successful runs only.}
\label{tab:SI:composite_score_DUST1_medium}
\end{table}

\begin{table}[H]
\centering
\begin{tabular}{lrrr}
\toprule
\textbf{Model Settings} & \textbf{Converged Runs} & \textbf{Mean Iteration} & \textbf{Composite Score} \\
\midrule
ei\_penalty & 7 & 13.71 & 0.051042 \\
lcb\_penalty\_HITL\_explore & 10 & 27.50 & 0.036364 \\
ei\_penalty\_HITL\_explore & 10 & 27.90 & 0.035842 \\
lcb\_penalty & 6 & 18.83 & 0.031858 \\
SOBOL\_off & 10 & 35.60 & 0.028090 \\
ei\_RF & 5 & 24.40 & 0.020492 \\
\bottomrule
\end{tabular}
\caption{Composite score ranking summary of all benchmarked models on the DUST1 function under the loose tolerance level. Mean iteration is reported for successful runs only.}
\label{tab:SI:composite_score_DUST1_loose}
\end{table}

\subsection{DUST2 Benchmark Complementary Results}\label{SI:subsec:DUST2}

\begin{figure}[H]
    \centering
    \includegraphics[width=1\linewidth]{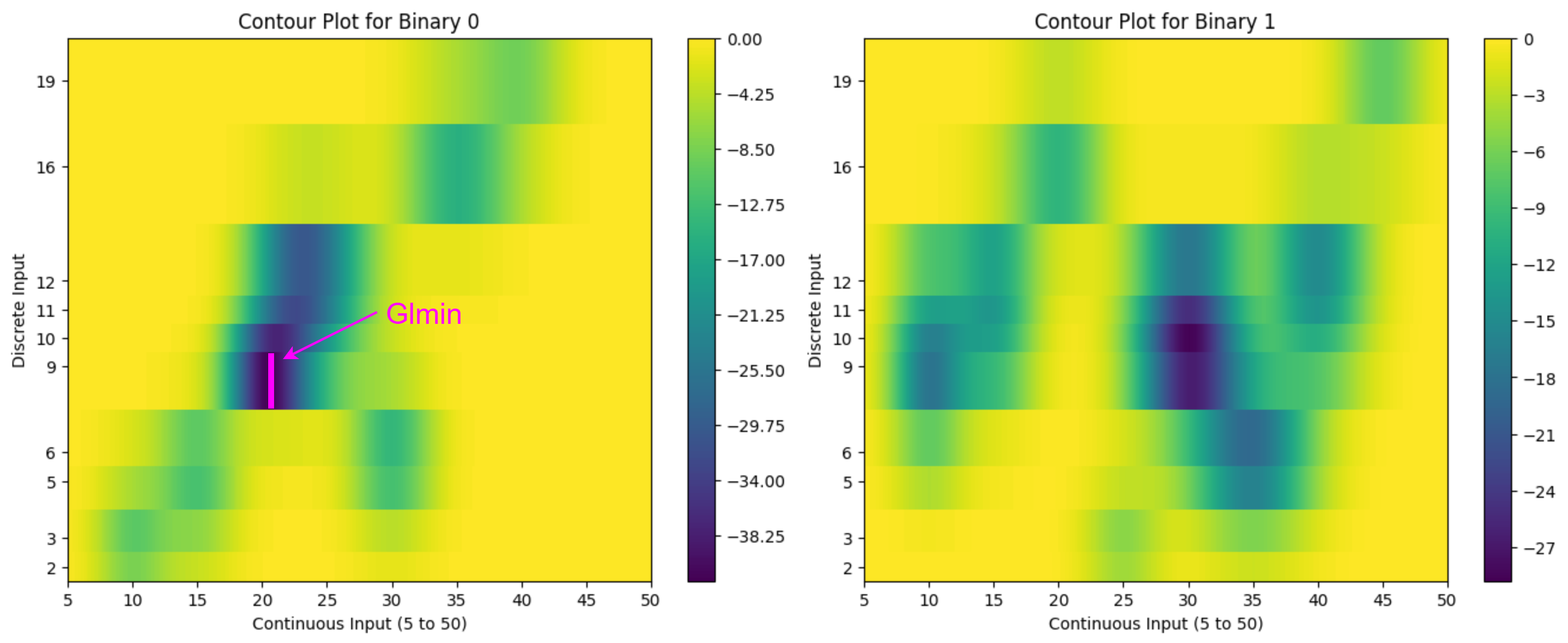}
    \caption{Objective Landscape of the DUST2 function with the global minima (Glmin) indicated in Pink.}
    \label{fig:SI:DUST2_landscape}
\end{figure}

\begin{table}[H]
\centering
\begin{tabular}{lrrr}
\toprule
\textbf{Model Settings} & \textbf{Converged Runs} & \textbf{Mean Iteration} & \textbf{Composite Score} \\
\midrule
lcb\_penalty & 6 & 23.00 & 0.026087 \\
ei\_penalty\_HITL\_explore & 7 & 37.29 & 0.018774 \\
lcb\_penalty\_HITL\_explore & 8 & 51.63 & 0.015496 \\
ei\_penalty & 5 & 42.20 & 0.011848 \\
ei\_RF & 4 & 62.75 & 0.006375 \\
SOBOL\_off & 0 & --- & 0.000000 \\
\bottomrule
\end{tabular}
\caption{Composite score ranking summary of all benchmarked models on the DUST2 function under the strict tolerance level. Mean iteration is reported for successful runs only.}
\label{tab:SI:composite_score_DUST2_strict}
\end{table}

\begin{table}[H]
\centering
\begin{tabular}{lrrr}
\toprule
\textbf{Model Settings} & \textbf{Converged Runs} & \textbf{Mean Iteration} & \textbf{Composite Score} \\
\midrule
lcb\_penalty & 6 & 23.00 & 0.026087 \\
ei\_penalty\_HITL\_explore & 7 & 34.71 & 0.020165 \\
lcb\_penalty\_HITL\_explore & 8 & 44.63 & 0.017927 \\
ei\_penalty & 5 & 41.40 & 0.012077 \\
ei\_RF & 4 & 62.75 & 0.006375 \\
SOBOL\_off & 1 & 68.00 & 0.001471 \\
\bottomrule
\end{tabular}
\caption{Composite score ranking summary of all benchmarked models on the DUST2 function under the medium tolerance level. Mean iteration is reported for successful runs only.}
\label{tab:SI:composite_score_DUST2_medium}
\end{table}

\begin{table}[H]
\centering
\begin{tabular}{lrrr}
\toprule
\textbf{Model Settings} & \textbf{Converged Runs} & \textbf{Mean Iteration} & \textbf{Composite Score} \\
\midrule
lcb\_penalty & 6 & 21.33 & 0.028125 \\
ei\_penalty\_HITL\_explore & 7 & 33.71 & 0.020763 \\
lcb\_penalty\_HITL\_explore & 9 & 53.78 & 0.016736 \\
ei\_penalty & 5 & 40.20 & 0.012438 \\
ei\_RF & 4 & 33.75 & 0.011852 \\
SOBOL\_off & 2 & 80.00 & 0.002500 \\
\bottomrule
\end{tabular}
\caption{Composite score ranking summary of all benchmarked models on the DUST2 function under the loose tolerance level. Mean iteration is reported for successful runs only.}
\label{tab:SI:composite_score_DUST2_loose}
\end{table}


\end{document}